\documentclass[letterpaper]{article} 
\usepackage{aaai23}  
\usepackage{times}  
\usepackage{helvet}  
\usepackage{courier}  
\usepackage[hyphens]{url}  
\usepackage{graphicx} 
\usepackage{natbib}  
\usepackage{caption} 
\usepackage{algorithm}
\usepackage{algorithmic}
\usepackage{todonotes}
\usepackage{afterpage}

\usepackage{newfloat}
\usepackage{listings}
\DeclareCaptionStyle{ruled}{labelfont=normalfont,labelsep=colon,strut=off} 
\floatstyle{ruled}
\newfloat{listing}{tb}{lst}{}
\floatname{listing}{Listing}
\title{Continuous Mixtures of Tractable Probabilistic Models}
\author{
    Alvaro H.C. Correia\textsuperscript{\rm 1,}\equalcontrib, Gennaro Gala\textsuperscript{\rm 1,}\equalcontrib,\\
    Erik Quaeghebeur\textsuperscript{\rm 1},
    Cassio de Campos\textsuperscript{\rm 1},
    Robert Peharz\textsuperscript{\rm 1,}\textsuperscript{\rm 2}
}
\affiliations{
    \textsuperscript{\rm 1} Eindhoven University of Technology\\
    \textsuperscript{\rm 2} Graz University of Technology\\
    \{a.h.chaim.correia, g.gala, e.quaeghebeur, c.decampos, r.peharz\}@tue.nl
}

\usepackage{bibentry}

\usepackage{subcaption}
\usepackage{mathtools}
\usepackage{bm}
\usepackage{amsthm}
\usepackage{siunitx}
\usepackage{booktabs}       
\usepackage{amsfonts}       
\usepackage{nicefrac}       
\usepackage{microtype}      
\usepackage{xcolor}         

\usepackage{amsmath,amsfonts,bm}

\newcommand{\cbar}{\,|\,}

\DeclareMathOperator*{\cm}{cm}
\DeclareMathOperator*{\dm}{dm}
\DeclareMathOperator{\struct}{\mathcal{S}}
\DeclareMathOperator{\param}{\phi}

\DeclareMathOperator{\structf}{\struct_{\mathsf{F}}}
\DeclareMathOperator{\structclt}{\struct_{\mathsf{CLT}}}
\DeclareMathOperator{\cmf}{\cm(\structf)}
\DeclareMathOperator{\cmclt}{\cm(\structclt)}
\DeclareMathOperator{\lo}{\mathsf{LO}}

\DeclareMathOperator{\locmclt}{\lo(\cmclt)}
\DeclareMathOperator{\cmvae}{\cm(\struct_{\mathsf{F}})_{\mathsf{VAE}}}

\def\eqref#1{equation~\ref{#1}}

\def\1{\bm{1}}

\def\rvu{{\mathbf{i}}}

\def\rvu{{\mathbf{u}}}

\def\rvx{{\mathbf{x}}}

\def\rvz{{\mathbf{z}}}

\def\rmX{{\mathbf{X}}}

\def\rmZ{{\mathbf{Z}}}

\DeclareMathAlphabet{\mathsfit}{\encodingdefault}{\sfdefault}{m}{sl}
\SetMathAlphabet{\mathsfit}{bold}{\encodingdefault}{\sfdefault}{bx}{n}

\def\gN{{\mathcal{N}}}

\def\gU{{\mathcal{U}}}

\newcommand{\E}{\mathbb{E}}

\DeclareMathOperator*{\argmax}{arg\,max}

\newcommand{\binarydatasetstablecomplete}{
\begin{table*}[!ht]
    \centering
    \resizebox{\textwidth}{!}{
    \begin{tabular}{l r r r r | l r r r r}
        \toprule
        Dataset                 & BestPC            & $\cmf$                   & $\cmclt$               & $\locmclt$           & 
        Dataset                 & BestPC            & $\cmf$                   & $\cmclt$               & $\locmclt$           \\
        \midrule
        \texttt{accidents}      & \textbf{-26.74}   & -33.27$\pm$0.03           & -28.69$\pm$0.01           & -28.81$\pm$0.02           &   
        \texttt{jester}         & -52.46            & \textbf{-51.93}$\pm$0.02  & -51.94$\pm$0.01           & -51.94$\pm$0.02           \\  
        \texttt{ad}             & -16.07            & -18.71$\pm$0.15           & -14.76$\pm$0.10           & \textbf{-14.42}$\pm$0.09  &   
        \texttt{kdd}            & \textbf{-2.12}    & -2.13$\pm$0.00            & \textbf{-2.12}$\pm$0.00   & \textbf{-2.12}$\pm$0.00   \\  
        \texttt{baudio}         & -39.77            & \textbf{-39.02}$\pm$0.02  & \textbf{-39.02}$\pm$0.02  & -39.04$\pm$0.02           &   
        \texttt{kosarek}        & -10.60            & -10.71$\pm$0.01           & -10.56$\pm$0.01           & \textbf{-10.55}$\pm$0.01  \\  
        \texttt{bbc}            & -248.33           & \textbf{-240.19}$\pm$0.29 & -242.83$\pm$0.55          & -242.79$\pm$0.58          &   
        \texttt{msnbc}          & \textbf{-6.03}    & -6.14$\pm$0.01            & -6.05$\pm$0.00            & -6.05$\pm$0.00            \\  
        \texttt{bnetflix}       & -56.27            & -55.49$\pm$0.02           & \textbf{-55.31}$\pm$0.02  & -55.36$\pm$0.02           &   
        \texttt{msweb}          & -9.73             & -9.68$\pm$0.00            & -9.62$\pm$0.01            & \textbf{-9.60}$\pm$0.01   \\  
        \texttt{book}           & -33.83            & -33.67$\pm$0.04           & -33.75$\pm$0.03           & \textbf{-33.55}$\pm$0.02  &   
        \texttt{nltcs}          & \textbf{-5.99}    & \textbf{-5.99}$\pm$0.00   & \textbf{-5.99}$\pm$0.01   & \textbf{-5.99}$\pm$0.00   \\  
        \texttt{c20ng}          & -151.47           & -148.24$\pm$0.10          & \textbf{-148.17}$\pm$0.09 & -148.28$\pm$0.11          &   
        \texttt{plants}         & -12.54            & -12.45$\pm$0.02           & \textbf{-12.26}$\pm$0.10  & -12.27$\pm$0.01           \\  
        \texttt{cr52}           & -83.35            & -81.52$\pm$0.08           & \textbf{-81.17}$\pm$0.11  & -81.31$\pm$0.15           &   
        \texttt{pumbs}          & \textbf{-22.40}   & -27.67$\pm$0.03           & -23.71$\pm$0.03           &  -23.70$\pm$0.02          \\  
        \texttt{cwebkb}         & -151.84           & -150.21$\pm$0.22          & -147.77$\pm$0.26          & \textbf{-147.75}$\pm$0.26 &   
        \texttt{tmovie}         & -50.81            & \textbf{-48.69}$\pm$0.09  & -49.23$\pm$0.10           & -49.29$\pm$0.12           \\  
        \texttt{dna}            & \textbf{-79.05}   & -95.64$\pm$0.37           & -84.91$\pm$0.09           & -84.58$\pm$0.10           &   
        \texttt{tretail}        & -10.84            & 10.85$\pm$0.00            & -10.82$\pm$0.01           & \textbf{-10.81}$\pm$0.01  \\  
        \midrule
        Avg. rank   & 2.85              & 2.65                      & 1.85                      & \textbf{1.75}  & & & & & \\
        \bottomrule
    \end{tabular}
    }
    \caption{Average test log-likelihoods and standard deviation across 5 random seeds on 20 density estimation benchmarks. BestPC refers to the best performing PC among Einets \citep{peharz2020einsum}, LearnSPN \citep{gens2013learning}, ID-SPN \citep{rooshenas2014learning}, RAT-SPN \citep{peharz2020random} and HCLT \citep{liu2021tractable}. For $\cmf$ and $\cmclt$, we train with $2^{10}$ points and test with a PC compiled with $2^{13}$ points. LO is run over $2^{10}$ points. Higher is better.
    }
    \label{tab:20datasetscomplete}
\end{table*}}

\newcommand{\binarydatasetstable}{
\begin{table*}[!ht]
    \centering
    {
        \fontsize{9}{9}\selectfont
        \begin{tabular}{l r r r r | l r r r r}
            \toprule
            Dataset                 & BestPC            & $\cmf$                   & $\cmclt$               & $\locmclt$           & 
            Dataset                 & BestPC            & $\cmf$                   & $\cmclt$               & $\locmclt$           \\
            \midrule
            \texttt{accid.}         & \textbf{-26.74}   & -33.27           & -28.69           & -28.81           &   
            \texttt{jester}         & -52.46            & \textbf{-51.93}  & -51.94           & -51.94           \\  
            \texttt{ad}             & -16.07            & -18.71           & -14.76           & \textbf{-14.42}  &   
            \texttt{kdd}            & \textbf{-2.12}    & -2.13            & \textbf{-2.12}   & \textbf{-2.12}   \\  
            \texttt{baudio}         & -39.77            & \textbf{-39.02}  & \textbf{-39.02}  & -39.04           &   
            \texttt{kosarek}        & -10.60            & -10.71           & -10.56           & \textbf{-10.55}  \\  
            \texttt{bbc}            & -248.33           & \textbf{-240.19} & -242.83          & -242.79          &   
            \texttt{msnbc}          & \textbf{-6.03}    & -6.14            & -6.05            & -6.05            \\  
            \texttt{bnetflix}       & -56.27            & -55.49           & \textbf{-55.31}  & -55.36           &   
            \texttt{msweb}          & -9.73             & -9.68            & -9.62            & \textbf{-9.60}   \\  
            \texttt{book}           & -33.83            & -33.67           & -33.75           & \textbf{-33.55}  &   
            \texttt{nltcs}          & \textbf{-5.99}    & \textbf{-5.99}   & \textbf{-5.99}   & \textbf{-5.99}   \\  
            \texttt{c20ng}          & -151.47           & -148.24          & \textbf{-148.17} & -148.28          &   
            \texttt{plants}         & -12.54            & -12.45           & \textbf{-12.26}  & -12.27           \\  
            \texttt{cr52}           & -83.35            & -81.52           & \textbf{-81.17}  & -81.31           &   
            \texttt{pumbs}          & \textbf{-22.40}   & -27.67           & -23.71           &  -23.70          \\  
            \texttt{cwebkb}         & -151.84           & -150.21          & -147.77          & \textbf{-147.75} &   
            \texttt{tmovie}         & -50.81            & \textbf{-48.69}  & -49.23           & -49.29           \\  
            \texttt{dna}            & \textbf{-79.05}   & -95.64           & -84.91           & -84.58           &   
            \texttt{tretail}        & -10.84            & 10.85            & -10.82           & \textbf{-10.81}  \\  
            \midrule
            Avg. rank   & 2.85              & 2.65                      & 1.85                      & \textbf{1.75}  & & & & & \\
            \bottomrule
        \end{tabular}
    }
    \caption{Average test log-likelihoods on 20 density estimation benchmarks. We compare $\cmf$, $\cmclt$ and $\locmclt$ with the best performing PC (BestPC) among 5 PC methods: Einets \citep{peharz2020einsum}, LearnSPN \citep{gens2013learning}, ID-SPN \citep{rooshenas2014learning}, RAT-SPN \citep{peharz2020random} and HCLT \citep{liu2021tractable}. For $\cmf$ and $\cmclt$, we train with $2^{10}$ integration points and test with a PC compiled with $2^{13}$ points. LO is run over $2^{10}$ integration points. See Tables~10 and 11 in Appendix E for more results including standard deviation over 5 random seeds. Higher is better.}
    \label{tab:20datasets}
\end{table*}}

\newcommand{\dmmclttable}{
\begin{table*}[!ht]
    \centering
    \resizebox{\textwidth}{!}{
     \begin{tabular}{l r r r r r r r}
        \toprule
        Dataset     & Mix-$2^{7}$       & Mix-$2^{8}$       & Mix-$2^{9}$       & Mix-$2^{10}$      & Mix-$2^{11}$      & Mix-$2^{12}$      & Mix-$2^{13}$      \\
        \midrule
        accidents   & -31.00$\pm$0.15   & -30.06$\pm$0.09   & -29.28$\pm$0.05   & -28.93$\pm$0.02   & -28.79$\pm$0.01   & -28.74$\pm$0.01   & -28.69$\pm$0.01   \\
        ad          & -16.67$\pm$0.39   & -15.89$\pm$0.25   & -15.27$\pm$0.08   & -14.97$\pm$0.11   & -14.85$\pm$0.11   & -14.81$\pm$0.10   & -14.76$\pm$0.10   \\
        baudio      & -39.88$\pm$0.07   & -39.51$\pm$0.08   & -39.21$\pm$0.02   & -39.08$\pm$0.02   & -39.04$\pm$0.02   & -39.03$\pm$0.02   & -39.02$\pm$0.02   \\
        bbc         & -249.01$\pm$0.94  & -246.56$\pm$0.73  & -244.75$\pm$0.51  & -243.75$\pm$0.56  & -243.27$\pm$0.54  & -243.07$\pm$0.52  & -242.83$\pm$0.55  \\
        bnetflix    & -56.54$\pm$0.06   & -56.01$\pm$0.07   & -55.60$\pm$0.03   & -55.42$\pm$0.02   & -55.35$\pm$0.02   & -55.34$\pm$0.02   & -55.31$\pm$0.02   \\  
        book        & -34.29$\pm$0.10   & -34.04$\pm$0.04   & -33.90$\pm$0.03   & -33.83$\pm$0.03   & -33.79$\pm$0.03   & -33.77$\pm$0.03   & -33.75$\pm$0.03   \\ 
        c20ng       & -151.59$\pm$0.45  & -150.39$\pm$0.28  & -149.28$\pm$0.17  & -148.68$\pm$0.11  & -148.41$\pm$0.09  & -148.28$\pm$0.09  & -148.17$\pm$0.09  \\ 
        cr52        & -85.39$\pm$0.40   & -83.82$\pm$0.24   & -82.35$\pm$0.14   & -81.70$\pm$0.09   & -81.41$\pm$0.11   & -81.30$\pm$0.09   & -81.17$\pm$0.11   \\ 
        cwebkb      & -150.73$\pm$0.28  & -149.55$\pm$0.20  & -148.59$\pm$0.21  & -148.18$\pm$0.24  & -147.94$\pm$0.25  & -147.86$\pm$0.26  & -147.77$\pm$0.26  \\ 
        dna         & -86.79$\pm$0.25   & -86.09$\pm$0.16   & -85.49$\pm$0.09   & -85.16$\pm$0.08   & -85.01$\pm$0.07   & -84.96$\pm$0.09   & -84.91$\pm$0.09   \\ 
        
        jester      & -52.55$\pm$0.05   & -52.27$\pm$0.03   & -52.06$\pm$0.02   & -51.98$\pm$0.01   & -51.95$\pm$0.01   & -51.95$\pm$0.01   & -51.94$\pm$0.01   \\
        kdd         & -2.16$\pm$0.01    & -2.14$\pm$0.00    & -2.13$\pm$0.00    & -2.12$\pm$0.00    & -2.12$\pm$0.00    & -2.12$\pm$0.00    & -2.12$\pm$0.00    \\
        kosarek     & -10.68$\pm$0.03   & -10.63$\pm$0.02   & -10.59$\pm$0.01   & -10.57$\pm$0.01   & -10.57$\pm$0.01   & -10.56$\pm$0.01   & -10.56$\pm$0.01   \\
        msnbc       & -6.44$\pm$0.16    & -6.22$\pm$0.10    & -6.09$\pm$0.01    & -6.07$\pm$0.00    & -6.06$\pm$0.00    & -6.06$\pm$0.00    & -6.05$\pm$0.00    \\
        msweb       & -9.89$\pm$0.05    & -9.78$\pm$0.03    & -9.67$\pm$0.02    & -9.64$\pm$0.01    & -9.63$\pm$0.01    & -9.63$\pm$0.01    & -9.62$\pm$0.01    \\
        nltcs       & -6.05$\pm$0.02    & -6.03$\pm$0.02    & -6.00$\pm$0.00    & -6.00$\pm$0.00    & -6.00$\pm$0.00    & -6.00$\pm$0.00    & -5.99$\pm$0.00    \\
        plants      & -13.68$\pm$0.13   & -12.98$\pm$0.07   & -12.53$\pm$0.02   & -12.35$\pm$0.01   & -12.30$\pm$0.02   & -12.28$\pm$0.01   & -12.26$\pm$0.01   \\
        pumbs       & -26.29$\pm$0.22   & -25.22$\pm$0.22   & -24.26$\pm$0.07   & -23.92$\pm$0.02   & -23.80$\pm$0.02   & -23.76$\pm$0.03   & -23.71$\pm$0.03   \\
        tmovie      & -51.67$\pm$0.27   & -50.59$\pm$0.16   & -49.90$\pm$0.12   & -49.56$\pm$0.01   & -49.38$\pm$0.10   & -49.29$\pm$0.11   & -49.23$\pm$0.10   \\
        tretail     & -10.84$\pm$0.01   & -10.83$\pm$0.01   & -10.82$\pm$0.01   & -10.82$\pm$0.01   & -10.82$\pm$0.01   & -10.82$\pm$0.01   & -10.82$\pm$0.01   \\
        \bottomrule
     \end{tabular}
     }
    \caption{Average $\cmclt$ test log-likelihoods on 20 density estimation datasets. Mean and standard deviation across five different random seeds reported. Higher is better.}
    \label{tab:cmclt}
\end{table*}}

\newcommand{\dmmtable}{
\begin{table*}[!ht]
    \centering
    \resizebox{\textwidth}{!}{
     \begin{tabular}{l r r r r r r r}
        \toprule
        Dataset     & Mix-$2^{7}$       & Mix-$2^{8}$       & Mix-$2^{9}$       & Mix-$2^{10}$      & Mix-$2^{11}$      & Mix-$2^{12}$      & Mix-$2^{13}$      \\
        \midrule
        accidents   & -38.78$\pm$0.22   & -36.62$\pm$0.17   & -34.85$\pm$0.11   & -33.94$\pm$0.05   & -33.58$\pm$0.05   & -33.41$\pm$0.05   & -33.27$\pm$0.03   \\
        ad          & -42.89$\pm$2.39   & -32.50$\pm$1.44   & -23.79$\pm$0.93   & -20.42$\pm$0.14   & -19.59$\pm$0.11   & -19.18$\pm$0.10   & -18.71$\pm$0.15   \\
        baudio      & -40.24$\pm$0.07   & -39.76$\pm$0.03   & -39.34$\pm$0.01   & -39.14$\pm$0.01   & -39.07$\pm$0.01   & -39.05$\pm$0.01   & -39.02$\pm$0.01   \\
        bbc         & -249.58$\pm$0.66  & -245.66$\pm$0.68  & -243.06$\pm$0.34  & -241.54$\pm$0.31  & -240.78$\pm$0.36  & -240.53$\pm$0.32  & -240.19$\pm$0.29  \\
        bnetflix    & -57.35$\pm$0.09   & -56.65$\pm$0.12   & -56.02$\pm$0.01   & -55.71$\pm$0.01   & -55.58$\pm$0.01   & -55.54$\pm$0.01   & -55.49$\pm$0.02   \\
        book        & -34.51$\pm$0.05   & -34.16$\pm$0.03   & -33.92$\pm$0.04   & -33.79$\pm$0.04   & -33.72$\pm$0.04   & -33.69$\pm$0.04   & -33.67$\pm$0.04   \\
        c20ng       & -153.71$\pm$0.56  & -151.58$\pm$0.45  & -149.96$\pm$0.30  & -149.10$\pm$0.21  & -148.64$\pm$0.15  & -148.45$\pm$0.12  & -148.24$\pm$0.10  \\
        cr52        & -87.57$\pm$0.64   & -85.51$\pm$0.48   & -83.35$\pm$0.37   & -82.33$\pm$0.18   & -81.93$\pm$0.13   & -81.74$\pm$0.08   & -81.52$\pm$0.08   \\
        cwebkb      & -155.62$\pm$0.68  & -153.70$\pm$0.43  & -151.93$\pm$0.26  & -151.00$\pm$0.17  & -150.59$\pm$0.18  & -150.43$\pm$0.21  & -150.21$\pm$0.22  \\
        dna         & -99.04$\pm$0.50   & -97.80$\pm$0.17   & -96.62$\pm$0.17   & -96.11$\pm$0.25   & -95.86$\pm$0.31   & -95.77$\pm$0.33   & -95.64$\pm$0.37   \\
        
        jester      & -52.97$\pm$0.03   & -52.53$\pm$0.04   & -52.20$\pm$0.02   & -52.03$\pm$0.01   & -51.97$\pm$0.02   & -51.95$\pm$0.02   & -51.93$\pm$0.02   \\
        kdd         & -2.20$\pm$0.03    & -2.18$\pm$0.04    & -2.15$\pm$0.01    & -2.13$\pm$0.00    & -2.13$\pm$0.00    & -2.13$\pm$0.00    & -2.13$\pm$0.00    \\
        kosarek     & -11.11$\pm$0.04   & -10.93$\pm$0.03   & -10.81$\pm$0.02   & -10.75$\pm$0.01   & -10.73$\pm$0.01   & -10.72$\pm$0.01   & -10.71$\pm$0.01   \\
        msnbc       & -6.39$\pm$0.04    & -6.25$\pm$0.02    & -6.17$\pm$0.01    & -6.15$\pm$0.01    & -6.15$\pm$0.01    & -6.15$\pm$0.01    & -6.14$\pm$0.01    \\
        msweb       & -10.31$\pm$0.04   & -10.03$\pm$0.04   & -9.80$\pm$0.01    & -9.72$\pm$0.00    & -9.69$\pm$0.00    & -9.69$\pm$0.00    & -9.68$\pm$0.00    \\
        nltcs       & -6.08$\pm$0.00    & -6.03$\pm$0.00    & -6.00$\pm$0.00    & -6.00$\pm$0.00    & -6.00$\pm$0.00    & -5.99$\pm$0.00    & -5.99$\pm$0.00    \\
        plants      & -14.72$\pm$0.16   & -13.71$\pm$0.14   & -12.99$\pm$0.02   & -12.65$\pm$0.03   & -12.54$\pm$0.03   & -12.49$\pm$0.03   & -12.45$\pm$0.02   \\
        pumbs       & -39.41$\pm$0.89   & -33.67$\pm$0.52   & -30.01$\pm$0.10   & -28.50$\pm$0.04   & -28.05$\pm$0.05   & -27.85$\pm$0.02   & -27.67$\pm$0.03   \\
        tmovie      & -51.95$\pm$0.26   & -50.68$\pm$0.08   & -49.59$\pm$0.13   & -49.12$\pm$0.08   & -48.87$\pm$0.08   & -48.78$\pm$0.10   & -48.69$\pm$0.09   \\
        tretail     & -10.96$\pm$0.04   & -10.90$\pm$0.02   & -10.87$\pm$0.00   & -10.85$\pm$0.00   & -10.85$\pm$0.00   & -10.85$\pm$0.00   & -10.85$\pm$0.00   \\
        \bottomrule
     \end{tabular}
     }
    \caption{Average $\cmf$ test log-likelihoods on 20 density estimation datasets. Mean and standard deviation across five different random seeds reported. Higher is better}
    \label{tab:cmf}
\end{table*}}

\newcommand{\secondlatopttable}{
\begin{table*}[!ht]
    \centering
    \resizebox{\textwidth}{!}{
     \begin{tabular}{l | r r | r r | r r | r r}
         \toprule
         Dataset    & Mix-$2^{7}$      & Mix-$2^{7}$(LO)    & Mix-$2^{8}$       & Mix-$2^{8}$(LO)   & Mix-$2^{9}$       & Mix-$2^{9}$(LO)    & Mix-$2^{10}$      & Mix-$2^{10}$(LO)      \\
         \midrule
         accidents  & -31.00$\pm$0.15  & -29.81$\pm$0.03   & -30.06$\pm$0.09   & -29.36$\pm$0.05   & -29.28$\pm$0.05   & -29.05$\pm$0.03   & -28.93$\pm$0.02   & -28.81$\pm$0.02   \\
         ad         & -16.67$\pm$0.39  & -15.08$\pm$0.17   & -15.89$\pm$0.25   & -14.73$\pm$0.13   & -15.27$\pm$0.08   & -14.51$\pm$0.11   & -14.97$\pm$0.11   & -14.42$\pm$0.09   \\ 
         baudio     & -39.88$\pm$0.07  & -39.45$\pm$0.04   & -39.51$\pm$0.08   & -39.25$\pm$0.03   & -39.21$\pm$0.02   & -39.11$\pm$0.02   & -39.08$\pm$0.02   & -39.04$\pm$0.02   \\
         bbc        & -249.01$\pm$0.94 & -245.57$\pm$0.75  & -246.56$\pm$0.73  & -244.03$\pm$0.74  & -244.75$\pm$0.51  & -243.25$\pm$0.59  & -243.75$\pm$0.56  & -242.79$\pm$0.58  \\
         bnetflix   & -56.54$\pm$0.06  & -56.02$\pm$0.03   & -56.01$\pm$0.07   & -55.69$\pm$0.02   & -55.60$\pm$0.03   & -55.48$\pm$0.02   & -55.42$\pm$0.02   & -55.36$\pm$0.02   \\
         book       & -34.29$\pm$0.10  & -33.79$\pm$0.07   & -34.04$\pm$0.04   & -33.69$\pm$0.08   & -33.90$\pm$0.03   & -33.61$\pm$0.06   & -33.83$\pm$0.03   & -33.55$\pm$0.02   \\
         c20ng      & -151.59$\pm$0.45 & -149.95$\pm$0.31  & -150.39$\pm$0.28  & -149.16$\pm$0.26  & -149.28$\pm$0.17  & -148.65$\pm$0.17  & -148.68$\pm$0.11  & -148.28$\pm$0.11  \\
         cr52       & -85.39$\pm$0.40  & -83.28$\pm$0.23   & -83.82$\pm$0.24   & -82.25$\pm$0.22   & -82.35$\pm$0.14   & -81.67$\pm$0.18   & -81.70$\pm$0.09   & -81.31$\pm$0.15   \\
         cwebkb     & -150.73$\pm$0.28 & -148.82$\pm$0.21  & -149.55$\pm$0.20  & -148.33$\pm$0.20  & -148.59$\pm$0.21  & -147.95$\pm$0.23  & -148.18$\pm$0.24  & -147.75$\pm$0.26  \\
         dna        & -86.79$\pm$0.25  & -85.64$\pm$0.24   & -86.09$\pm$0.16   & -85.03$\pm$0.13   & -85.49$\pm$0.09   & -84.78$\pm$0.12   & -85.16$\pm$0.08   & -84.58$\pm$0.10   \\
         
         jester     & -52.55$\pm$0.05  & -52.23$\pm$0.03   & -52.27$\pm$0.03   & -52.06$\pm$0.02   & -52.06$\pm$0.02   & -51.98$\pm$0.02   & -51.98$\pm$0.01   & -51.94$\pm$0.02   \\
         kdd        & -2.16$\pm$0.01   & -2.13$\pm$0.00    & -2.14$\pm$0.00    & -2.12$\pm$0.00    & -2.13$\pm$0.00    & -2.12$\pm$0.00    & -2.12$\pm$0.00    & -2.12$\pm$0.00    \\
         kosarek    & -10.68$\pm$0.03  & -10.60$\pm$0.01   & -10.63$\pm$0.02   & -10.58$\pm$0.01   & -10.59$\pm$0.01   & -10.56$\pm$0.01   & -10.57$\pm$0.01   & -10.55$\pm$0.01   \\
         msnbc      & -6.44$\pm$0.16   & -6.08$\pm$0.01    & -6.22$\pm$0.10    & -6.06$\pm$0.00    & -6.09$\pm$0.01    & -6.05$\pm$0.00    & -6.07$\pm$0.00    & -6.05$\pm$0.00    \\
         msweb      & -9.89$\pm$0.05   & -9.67$\pm$0.01    & -9.78$\pm$0.03    & -9.64$\pm$0.02    & -9.67$\pm$0.02    & -9.61$\pm$0.01    & -9.64$\pm$0.01    & -9.60$\pm$0.01    \\
         nltcs      & -6.05$\pm$0.02   & -6.00$\pm$0.00    & -6.03$\pm$0.02    & -6.00$\pm$0.00    & -6.00$\pm$0.00    & -5.99$\pm$0.00    & -6.00$\pm$0.00    & -5.99$\pm$0.00    \\
         plants     & -13.68$\pm$0.13  & -12.76$\pm$0.04   & -12.98$\pm$0.07   & -12.50$\pm$0.02   & -12.53$\pm$0.02   & -12.34$\pm$0.02   & -12.35$\pm$0.01   & -12.27$\pm$0.01   \\
         pumbs      & -26.29$\pm$0.22  & -24.63$\pm$0.05   & -25.22$\pm$0.22   & -24.17$\pm$0.07   & -24.26$\pm$0.07   & -23.86$\pm$0.04   & -23.92$\pm$0.02   & -23.70$\pm$0.02   \\
         tmovie     & -51.67$\pm$0.27  & -50.31$\pm$0.18   & -50.59$\pm$0.16   & -49.79$\pm$0.23   & -49.90$\pm$0.12   & -49.40$\pm$0.12   & -49.56$\pm$0.01   & -49.29$\pm$0.12   \\  
         tretail    & -10.84$\pm$0.01  & -10.82$\pm$0.01   & -10.83$\pm$0.01   & -10.81$\pm$0.01   & -10.82$\pm$0.01   & -10.81$\pm$0.01   & -10.82$\pm$0.01   & -10.81$\pm$0.01   \\
         \bottomrule
     \end{tabular}
     }
    \caption{Latent Optimisation (LO) results for $\cmclt$. Integration points optimised for 150 epochs (with early stopping) using Adam with learning rate of 0.001. Mean and standard deviation across five different random seeds reported. Higher is better.}
    \label{tab:latopt}
\end{table*}}

\newcommand{\bmnisttable}{
\begin{table*}
    \centering
    \resizebox{\textwidth}{!}{
     \begin{tabular}{cc|cccccccc}
        \toprule
        &                       &           & \multicolumn{5}{c}{Number of integration points at test time}                                                                                              \\
        \cmidrule(lr{1em}){3-10}
        Model                   & N. Param  & $2^{7}$           & $2^{8}$           & $2^{9}$           & $2^{10}$          & $2^{11}$      & $2^{12}$      & $2^{13}$      & $2^{14}$  \\
        \midrule
        $\cmf$ (LO)       & 1.2M      & -167.29 (-144.00) & -150.67 (-135.89) & -138.55 (-129.15) & -129.24 (-123.44) & -121.96       & -116.42       & -112.03       & -108.69       \\
        $\cmclt$ (LO)   & 4.8M      & -127.59 (-114.02) & -119.09 (-110.02) & -113.15 (-107.14) & -108.30 (-104.37) & -104.50       & -101.55       & -99.23        & -97.48        \\
        \bottomrule
    \end{tabular}
    }
    \caption{Binary-MNIST test log-likelihoods for $\cmf$ and $\cmclt$ trained with $2^{14}$ integration points. In parentheses we report test log-likelihoods obtained via latent optimisation. Higher is better.}
    \label{tab:bmnist}
\end{table*}}

\newcommand{\secondvaetable}{
\begin{table*}[!t]
    \centering
    \resizebox{\textwidth}{!}{
     \begin{tabular}{l | r r r r r r r r | r}
        \toprule
        & \multicolumn{8}{c}{$\cmvae$} & $\cmf$ \\
        \cmidrule{2-10}
        Dataset     & ELBO              & Mix-$2^{7}$       & Mix-$2^{8}$       & Mix-$2^{9}$       & Mix-$2^{10}$      & Mix-$2^{11}$      & Mix-$2^{12}$      & Mix-$2^{13}$    & Mix-$2^{13}$  \\
        \midrule
        accidents   & -34.56$\pm$0.19   & -37.34$\pm$0.62   & -35.78$\pm$0.23   & -35.25$\pm$0.12   & -34.80$\pm$0.15   & -34.61$\pm$0.25   & -34.59$\pm$0.27   & \underline{-34.55}$\pm$0.25 & \textbf{-33.27}$\pm$0.03  \\
        ad          & -22.27$\pm$0.92   & -38.39$\pm$2.80   & -30.25$\pm$2.25   & -25.35$\pm$1.31   & -23.18$\pm$1.43   & -21.69$\pm$0.96   & -21.36$\pm$0.96   & \underline{-21.08}$\pm$1.02 & \textbf{-18.71}$\pm$0.15  \\
        baudio      & \underline{-39.36}$\pm$0.20   & -40.30$\pm$0.23   & -40.00$\pm$0.15   & -39.80$\pm$0.10   & -39.73$\pm$0.11   & -39.70$\pm$0.11   & -39.69$\pm$0.11   & -39.68$\pm$0.11 & \textbf{-39.02}$\pm$0.01   \\
        bbc         & -243.04$\pm$0.26  & -247.25$\pm$0.50  & -244.68$\pm$0.29  & -243.10$\pm$0.43  & -242.26$\pm$0.41  & -241.59$\pm$0.42  & -241.38$\pm$0.44  & \underline{-241.24}$\pm$0.41 & \textbf{-240.19}$\pm$0.29  \\
        bnetflix    & \underline{-56.38}$\pm$0.35   & -57.30$\pm$0.10   & -56.97$\pm$0.12   & -56.78$\pm$0.21   & -56.74$\pm$0.24   & -56.72$\pm$0.25   & -56.71$\pm$0.26   & -56.71$\pm$0.26  & \textbf{-55.49}$\pm$0.02 \\
        book        & \underline{-34.56}$\pm$0.55   & -34.96$\pm$0.34   & -34.83$\pm$0.43   & -34.79$\pm$0.44   & -34.73$\pm$0.46   & -34.70$\pm$0.48   & -34.68$\pm$0.48   & -34.68$\pm$0.48  & \textbf{-33.67}$\pm$0.04 \\
        c20ng       & -149.53$\pm$0.32  & -152.59$\pm$0.88  & -151.22$\pm$0.40  & -150.38$\pm$0.13  & -149.93$\pm$0.22  & -149.54$\pm$0.24  & -149.38$\pm$0.30  & \underline{-149.26}$\pm$0.32 & \textbf{-148.24}$\pm$0.10 \\
        cr52        & -82.62$\pm$0.22   & -86.08$\pm$0.57   & -84.40$\pm$0.30   & -83.05$\pm$0.22   & -82.40$\pm$0.18   & -81.94$\pm$0.10   & -81.80$\pm$0.11   & \underline{-81.66}$\pm$0.10  & \textbf{-81.52}$\pm$0.08 \\
        cwebkb      & -151.44$\pm$0.45  & -153.94$\pm$0.34  & -152.76$\pm$0.09  & -152.04$\pm$0.28  & -151.63$\pm$0.29  & -151.32$\pm$0.43  & -151.20$\pm$0.44  & \underline{-151.11}$\pm$0.47 & \textbf{-150.21}$\pm$0.22 \\
        dna         & -96.77$\pm$0.29   & -97.25$\pm$0.20   & -97.06$\pm$0.24   & -96.79$\pm$0.30   & -96.69$\pm$0.33   & -96.59$\pm$0.36   & -96.56$\pm$0.37   & \underline{-96.53}$\pm$0.38  & \textbf{-95.64}$\pm$0.37 \\
        
        jester      & \underline{-52.79}$\pm$0.33   & -53.26$\pm$0.25   & -53.04$\pm$0.21   & -53.00$\pm$0.24   & -52.99$\pm$0.25   & -52.98$\pm$0.25   & -52.98$\pm$0.25   & -52.98$\pm$0.25  & \textbf{-51.93}$\pm$0.02 \\
        kdd         & \underline{\textbf{-2.05}}$\pm$0.01    & -2.17$\pm$0.01    & -2.17$\pm$0.01    & -2.16$\pm$0.01    & -2.16$\pm$0.01    & -2.15$\pm$0.01    & -2.15$\pm$0.01    & -2.14$\pm$0.00   & -2.13$\pm$0.00 \\
        kosarek     & \underline{-10.75}$\pm$0.04   & -11.07$\pm$0.04   & -11.00$\pm$0.04   & -10.98$\pm$0.04   & -10.96$\pm$0.04   & -10.96$\pm$0.04   & -10.96$\pm$0.04   & -10.96$\pm$0.04  & \textbf{-10.71}$\pm$0.01 \\
        msnbc       & \underline{-6.32}$\pm$0.00    & -6.49$\pm$0.00    & -6.49$\pm$0.00    & -6.49$\pm$0.00    & -6.49$\pm$0.00    & -6.49$\pm$0.00    & -6.49$\pm$0.00    & -6.49$\pm$0.00   & \textbf{-6.14}$\pm$0.01 \\
        msweb       & \underline{\textbf{-9.61}}$\pm$0.04   & -10.17$\pm$0.10   & -10.08$\pm$0.06   & -10.01$\pm$0.04   & -9.99$\pm$0.04    & -9.98$\pm$0.04    & -9.98$\pm$0.04    & -9.98$\pm$0.04    & -9.68$\pm$0.00\\
        nltcs       & \underline{-6.19}$\pm$0.16    & -6.35$\pm$0.12    & -6.35$\pm$0.12    & -6.35$\pm$0.12    & -6.35$\pm$0.12    & -6.35$\pm$0.12    & -6.35$\pm$0.12    & -6.35$\pm$0.12   & \textbf{-5.99}$\pm$0.00 \\
        plants      & \underline{-12.80}$\pm$0.07   & -14.66$\pm$0.68   & -13.80$\pm$0.45   & -13.31$\pm$0.07   & -13.14$\pm$0.03   & -13.03$\pm$0.07   & -13.00$\pm$0.08   & -12.99$\pm$0.08  & \textbf{-12.45}$\pm$0.02 \\
        pumbs       & -29.21$\pm$0.37   & -37.79$\pm$0.92   & -33.54$\pm$0.47   & -30.77$\pm$0.18   & -29.78$\pm$0.19   & -29.25$\pm$0.32   & -29.11$\pm$0.39   & \underline{-28.86}$\pm$0.35  & \textbf{-27.67}$\pm$0.03 \\
        tmovie      & -49.01$\pm$0.34   & -52.07$\pm$0.87   & -50.77$\pm$0.42   & -49.83$\pm$0.08   & -49.43$\pm$0.20   & -49.15$\pm$0.25   & -49.07$\pm$0.27   & \underline{-48.98}$\pm$0.28  & \textbf{-48.69}$\pm$0.09 \\
        tretail     & \underline{\textbf{-10.79}}$\pm$0.05   & -11.01$\pm$0.01   & -11.00$\pm$0.01   & -11.00$\pm$0.01   & -11.00$\pm$0.01   & -11.00$\pm$0.01   & -11.00$\pm$0.01   & -11.00$\pm$0.01  & -10.85$\pm$0.00\\
        \bottomrule
     \end{tabular}
     }
    \caption{Results for $\cmvae$ with 4 latent dimensions and trained for 200 epochs. We compute the ELBO with 1000 Monte Carlo samples and report the mean and standard deviation of results obtained with 5 different random seeds. We underline the best result between ELBO and numerical integration in $\cmvae$ and use boldface for the best result overall.}
    \label{tab:vae}
\end{table*}}

\newcommand{\plainmixturestable}{

\begin{table*}[!t]
    \centering
    \resizebox{\textwidth}{!}{
    \begin{tabular}{lrrrr|lrrrr}
         \toprule
         Dataset    & $\dm(\struct_F)$     & dm$(\struct_F^W)$     & $\dm(\struct_F^{EM})$  & $\cmf$              & Dataset   & $\dm(\struct_F)$     & $\dm(\struct_F^W)$     & $\dm(\struct_F^{EM})$    & $\cmf$ \\
         \midrule
         accidents  & -42.58    & -40.61    & -35.38 & \textbf{-33.94}   & jester    & -55.32    & -53.54    & -52.54  & \textbf{-52.03}   \\
         ad         & -104.57   &  -97.79   & -24.91 & \textbf{-20.42}   & kdd       & -6.81     & -2.15     & -2.14 & \textbf{-2.13}    \\
         baudio     & -42.24    & -40.41    & -39.76 & \textbf{-39.14}   & kosarek   & -16.20    & -11.17    & -10.88  & \textbf{-10.75}   \\
         bbc        & -281.88   & -288.31   & -252.82 & \textbf{-241.54}  & msnbc     & -6.36     & -6.12    & \textbf{-6.03}  & -6.15    \\
         bnetflix   & -58.19    & -57.00    & -56.34 & \textbf{-55.71}   & msweb     & -18.29    & -11.36    &  -10.00 & \textbf{-9.72}    \\
         book       & -41.72    & -35.61    & -34.66 & \textbf{-33.79}   & nltcs     & -6.16     & -6.01     & -6.00 & \textbf{-6.00}    \\
         c20ng      & -163.04   & -157.80   & -151.79 & \textbf{-149.10}  & plants    & -16.66    & -14.41   & -13.44  & \textbf{-12.65}   \\
         cr52       & -104.91   & -98.79    & -87.07 & \textbf{-82.33}   & pumbs     & -46.59    & -42.90    & -32.84  & \textbf{-28.50}   \\
         cwebkb     & -176.60   & -170.90   & -154.75 & \textbf{-151.00}  & tmovie    & -66.94    & -61.64   & -52.80  & \textbf{-49.12}   \\
         dna        & -101.93   & -98.14    & \textbf{-94.46} & -96.11   & tretail   & -18.35    & -11.42    & -10.90  & \textbf{-10.85}   \\
         \bottomrule
     \end{tabular}
     }
    \caption{Test log-likelihoods on 20 standard density estimation benchmarks for plain mixtures with non-learnable equally-probable weights trained with Adam ($\dm(\struct_F)$); with learnable weights also trained with Adam ($\dm(\struct_F^W)$); with learnable weights trained via EM ($\dm(\struct_F^{EM})$); and $\cmf$ evaluated with 1024 integration points. All models had the same structure with 1024 components (integration points) and were trained for up to 300 epochs with early stopping. Higher is better.}
     \label{tab:mix}
\end{table*}
}

\newcommand{\competitorstable}{
\begin{table*}[ht!]
    \centering
    \resizebox{\textwidth}{!}{
    \begin{tabular}{lccccccccc}
        \toprule
        Dataset & HCLT & Einet & LearnSPN & ID-SPN & RAT-SPN & Strudel & LearnPSDD & $\cmf$ & $\cmclt$ \\
        \midrule
        accidents & \textbf{-26.78} & -35.59 & -40.50 & -26.98 & -35.48 & -29.46 & -28.29 & -33.27 & -28.69 \\
        ad & -16.04 & -26.27 & -19.73 & -19.00 & -48.47 & -16.52 & -20.13 & -18.72 & \textbf{-14.76} \\
        baudio & -39.77 & -39.87 & -40.53 & -39.79 & -39.95 & -42.26 & -41.51 & \textbf{-39.02} & \textbf{-39.02} \\
        bbc & -250.07 & -248.33 & -250.68 & -248.93 & -252.13 & -258.96 & -260.24 & \textbf{-240.19} & -242.83\\
        bnetflix & -56.28 & -56.54 & -57.32 & -56.36 & -56.85 & -58.68 & -58.53 & -55.49 & \textbf{-55.31} \\
        book & -33.84 & -34.73 & -35.88 & -34.14 & -34.68 & -35.77 & -36.06 & \textbf{-33.67} & -33.75 \\
        c20ng & -151.92 & -153.93 & -155.92 & -151.47 & -152.06 & -160.77 & -160.43 & -148.24 & \textbf{-148.17} \\
        cr52 & -84.67 & -87.36 & -85.06 & -83.35 & -87.36 & -92.38 & -93.30 & -81.52 & \textbf{-81.17} \\
        cwebkb & -153.18 & -157.28 & -158.20 & -151.84 & -157.53 & -160.50 & -161.42 & -150.21 & \textbf{-147.77} \\
        dna & \textbf{-79.33} & -96.08 & -82.52 & -81.21 & -97.23 & -87.10 & -83.02 & -95.64 & -84.91 \\
        jester & -52.45 & -52.56 & -75.98 & -52.86 & -52.97 & -55.30 & -54.63 & \textbf{-51.93} & -51.94 \\
        kdd & -2.18 & -2.18 & -2.18 & -2.13 & \textbf{-2.12} & -2.17 & -2.17 & -2.13 & \textbf{-2.12} \\
        kosarek & -10.66 & -11.02 & -10.98 & -10.60 & -10.88 & -10.98 & -10.99 & -10.70 & \textbf{-10.56} \\
        msnbc & -6.05 & -6.11 & -6.11 & -6.04 & \textbf{-6.03} & -6.05 & -6.04 & -6.14 &  -6.05 \\
        msweb & -9.90 & -10.02 & -10.25 & -9.73 & -10.11 & -10.19 & -9.93 & -9.68 & \textbf{-9.62} \\
        nltcs & -6.00 & -6.01 & -6.11 & -6.02 & -6.01 & -6.06 & -6.03 & \textbf{-5.99} & \textbf{-5.99} \\
        plants & -14.31 & -13.67 & -12.97 & -12.54 & -13.43 & -13.72 & -13.49 & -12.45 & \textbf{-12.26}\\
        pumbs & -23.32 & -31.95 & -24.78 & \textbf{-22.40} & -32.53 & -25.28 & -25.40 & -27.67 & -23.71 \\
        tmovie & -50.69 & -51.70 & -52.48 & -51.51 & -53.63 & -59.47 & -55.41 & \textbf{-48.69} & -49.23 \\
        tretail & -10.84 & -10.91 & -11.04 & -10.85 & -10.91 & -10.90 & -10.92 & -10.85 & \textbf{-10.82} \\
        \bottomrule
    \end{tabular}
    }
    \caption{
    Results on 20 density estimation benchmarks. Average test-set log-likelihood of all baselines as reported on the respective papers: Strudel \citep{dang2020strudel}, LearnPSDD \citep{liang2017learning}, Einets \citep{peharz2020einsum}, LearnSPN \citep{gens2013learning}, ID-SPN \citep{rooshenas2014learning}, and RAT-SPN \citep{peharz2020random}. Results for $\cmf$ and $\cmclt$ are computed with $2^{13}$ integration points. Higher is better.}
    \label{tab:20datasets-extend}
\end{table*}}

\newcommand{\imagedatasetstable}{
\begin{table}
    {
        \fontsize{9}{10}\selectfont
        \centering
         \begin{tabular}{ l c c c}
             \toprule
             Model & `Small Einet' & `Big Einet' & Ours ($2^{14}$) \\
             \midrule
             Binary-MNIST   & 0.206   & 0.184  & \textbf{0.179}   \\
             MNIST          & 1.490    & 1.415   & \textbf{1.282}    \\
             Fashion-MNIST  & 3.938    & 3.737   & \textbf{3.546}  \\
             SVHN           & 6.442   & \textbf{5.961}  & 6.307   \\
             \bottomrule
         \end{tabular}
    }
     \caption{Bits per dim. (bpd) for image data. Lower is better.}
     \label{tab:imagedatasets}
\end{table}
}

\newcommand{\imagedatasetsnormaltable}{
\begin{table}[!hb]
    \small
    \centering
     \begin{tabular}{ l c c c}
         \toprule
         Model & `Small Einet' & `Big Einet' & Ours ($2^{14}$) \\
         \midrule
         MNIST          & 2.651    & \textbf{2.057}   & 2.762    \\
         Fashion-MNIST  & 4.309    & \textbf{3.938}   & 4.485  \\
         SVHN           & 6.332   & \textbf{5.972}  & 6.400   \\
         \bottomrule
     \end{tabular}
    \caption[Bits per dimension (bpd) for image data and models using normal distributions at the leaves.]{Bits per dimension (bpd) for image data and models using normal distributions at the leaves. Lower is better.}
    \label{tab:imagedatasetsnormal}
\end{table}
}

\newtheorem{definition}{Definition}

\begin{document}

\maketitle

\begin{abstract}
Probabilistic models based on continuous latent spaces, such as variational autoencoders, can be understood as uncountable mixture models where components depend continuously on the latent code. They have proven to be expressive tools for generative and probabilistic modelling, but are at odds with tractable probabilistic inference, that is, computing marginals and conditionals of the represented probability distribution. Meanwhile, tractable probabilistic models such as probabilistic circuits (PCs) can be understood as hierarchical discrete mixture models, and thus are capable of performing exact inference efficiently but often show subpar performance in comparison to continuous latent-space models. In this paper, we investigate a hybrid approach, namely continuous mixtures of tractable models with a small latent dimension. While these models are analytically intractable, they are well amenable to numerical integration schemes based on a finite set of integration points. With a large enough number of integration points the approximation becomes de-facto exact. Moreover, for a finite set of integration points, the integration method effectively compiles the continuous mixture into a standard PC. In experiments, we show that this simple scheme proves remarkably effective, as PCs learnt this way set new state of the art for tractable models on many standard density estimation benchmarks.
\end{abstract}

\section*{Introduction}
Probabilistic modelling typically aims to capture the data-generating joint distribution, which can then be used to perform probabilistic inference to answer queries of interest. A recurring scheme in probabilistic modelling is the use of an \emph{uncountable mixture model}, that is, the data generating distribution is approximated by
\begin{equation}    \label{eq:continuous_mixture}
    p(\rvx) 
    = \E_{p(\rvz)} \left [ p(\rvx \cbar \rvz ) \right ] 
    = \int  p(\rvx \cbar \rvz ) \, p(\rvz) \, \mathrm{d} \rvz
\end{equation}
where $p(\rvz)$ is a mixing distribution (prior) over latent variables $\rmZ$, $p(\rvx \cbar \rvz)$ is a conditional distribution of $\rvx$ given $\rvz$ (mixture components), and $p(\rvx)$ is the modelled density over variables $\rmX$, given by marginalising $\rmZ$ from the joint distribution defined by $p(\rvx \cbar \rvz)\,p(\rvz)$.

Some successful recent examples of uncountable mixtures are variational autoencoders (VAEs) \cite{kingma2013auto}, generative adversarial networks (GANs) \cite{goodfellow2014generative}, and normalising Flows \cite{Rezende2015}. All three of these models use a simple prior $p(\rvz)$, e.g.~an isotropic Gaussian, and represent the mixture components with a neural network. In the case of VAEs, the mixture component is a proper density $p(\rvx \cbar \rvz)$ with respect to the Lebesgue measure, represented by the so-called decoder, while for GANs and Flows the mixture component is a point measure, i.e.~a deterministic function $\rvx = f(\rvz)$\footnote{In normalising Flows $\rvz$ is not truly latent since it relates to $\rvx$ via a bijection.}. The use of continuous neural networks topologically relates the latent space and the observable space with each other, so that these models can be described as \emph{continuous} mixture models. The use of continuous mixtures allows, to a certain extent, the interpretation of $\rmZ$ as a (latent) embedding of $\rmX$, but also seems to benefit generalisation, i.e.~to faithfully approximate real-world distributions with limited training data.

However, while continuous mixture models have achieved impressive results in density estimation and generative modelling, their ability to support probabilistic inference remains limited.
Notably, the key inference routines of \emph{marginalisation} and \emph{conditioning}, which together form a consistent reasoning process \cite{Ghahramani2015,Jaynes2003}, are largely intractable in these models, mainly due to the integral in {(\ref{eq:continuous_mixture})} which forms a hard computational problem in general.

Meanwhile, the area of \emph{tractable probabilistic modelling} aims at models which allow a wide range of exact and efficient inference routines. One of the most prominent frameworks to express tractable models are \emph{probabilistic circuits} (PCs) \cite{Vergari2020}, which are computational graphs composed of (simple) tractable input distributions, factorisations (product nodes) and discrete mixture distributions (sum nodes). PCs describe many tractable models such as \emph{Chow-Liu trees} \cite{chow1968approximating}, \emph{arithmetic circuits} \cite{darwiche2003differential}, \emph{sum-product networks} \cite{poon2011sum}, \emph{cutset networks} \cite{rahman2014cutset},  \emph{probabilistic sentential decision diagrams} \cite{Kisa2014}, and \emph{generative forests} \cite{Correia2020}. The distribution represented by a PC depends both on its network structure $\struct$ and its parameters $\param$, which contains all weights of its sum nodes and parameters of its input distributions.

From a representational point of view, PCs can be interpreted as hierarchical, discrete mixture models \cite{peharz2016latent,zhao2016unified}, i.e.~they can be generally written as 
\begin{equation}    \label{eq:discrete_mixture}
    p(\rvx) = \sum_{\rvz'} p(\rvx \cbar \rvz') \, p(\rvz')
\end{equation}
where $\rmZ'$ is a \emph{discrete} latent vector, but otherwise the form is similar to the continuous mixture in {(\ref{eq:continuous_mixture})}. The number of states of $\rmZ'$ and thus the number of represented mixture components $p(\rvx \cbar \rvz')$ grows exponentially in the depth of the PC \cite{Peharz2015a,zhao2016unified}.
Moreover, recent vectorisation-based implementations \cite{peharz2020einsum} have enabled large PC architectures ($>$100M of parameters) at execution speeds comparable to standard neural networks. These endeavours evidently boosted the performance of tractable models, yet there is still a notable gap to intractable models like VAEs. One reason for this performance gap is likely the structural constraints in PCs, which are required to maintain tractability but are at odds with expressivity. On the other hand, a \emph{huge} discrete mixture model of the form (\ref{eq:discrete_mixture}) should in principle be able to outperform a moderately sized uncountable mixture (\ref{eq:continuous_mixture}). Yet, on standard benchmarks, we do not see this result. For instance, a vanilla VAE with a few million parameters gets test log-likelihoods higher than -90 nats on Binary MNIST \cite{tomczak2018vae}, while an Einet with 84 million parameters \cite{peharz2020einsum} barely gets above -100 nats (or 0.184 bpd) as shown in Table~\ref{tab:imagedatasets}. Thus, a complementary explanation is that discrete (and hierarchical) mixtures---like PCs---are hard to learn (or generalise poorly), while continuous mixtures---like VAEs---are easier to learn (or generalise well).

In this paper, we follow a hybrid approach and consider \emph{continuous mixtures of tractable models.}
In particular, we consider continuous mixtures of two very simple tractable models, namely \emph{completely factorised distributions} and \emph{tree-shaped graphical models}, which can both be easily expressed as PCs.
Specifically, we consider models of the form 
\begin{equation}    \label{eq:cm_model}
    p(\rvx) = \E_{p(\rvz)} \left [ p(\rvx \cbar \param(\rvz)) \right ],
\end{equation}
where $p(\rvz)$ is an isotropic Gaussian prior and $p(\rvx \cbar \param(\rvz))$ is a PC with its parameters depending on $\rvz$ via some neural-network $\param(\rvz)$. That has certain parallels to previous works in \emph{Conditional SPNs} \cite{shao2020conditional} and \emph{HyperSPNs} \cite{shih2021hyperspns}. While continuous mixtures are analytically intractable, we can approximate the marginalisation of $\rvz$  arbitrarily well with numerical techniques, such as \emph{Gaussian quadrature rules} and \emph{(quasi) Monte Carlo}. The common principle of these methods is that they select a finite set of \emph{integration points} $\rvz_{1}, \dots, \rvz_{N}$ in either a deterministic or (partially) random manner and construct a corresponding weight function $w(\rvz)$ such that (\ref{eq:cm_model}) is approximated as
\begin{equation}    \label{eq:numerical_approximation}
    p(\rvx) = 
    \E_{p(\rvz)} \left [ p(\rvx \cbar \param(\rvz)) \right ]
    \approx
    \sum_{i=1}^N  w(\rvz_i) \, p(\rvx \cbar \param(\rvz_i)).
\end{equation}
All integration methods we consider become exact for $N {\rightarrow} \infty$ and, under certain conditions on $\param(\rvz)$, one can derive guarantees for the approximation error for $N<\infty$. In particular, for (quasi) Monte Carlo integration it is straightforward to derive probabilistic error guarantees for the approximation quality by leveraging concentration bounds. Moreover, an empirical observation is that numerical integration works reasonably well for low dimensional spaces, but tends to deteriorate for larger dimensionality. Thus, in this paper we keep the dimensionality of $\rmZ$ relatively small ($\leq 16$), so that our continuous mixtures of tractable models remain `morally tractable'.

Specifically, the integration weights $\{w(\rvz_i)\}_{i=1}^{N}$ typically sum to one\footnote{For Monte Carlo, the weights are simply $w(z_i) = \frac{1}{N}$. For Gaussian quadratures the sum of the weights is a function of the domain of integration, but knowledge that $p(\rvx)$ is a probability distribution gives licence to re-normalise the weights in this case.}, so that the approximation in (\ref{eq:numerical_approximation}) can be interpreted as a discrete mixture model. For fixed $\rvz_i$, each $p(\rvx \cbar \param(\rvz_i))$ is simply a PC with fixed parameters $\param_i = \param(\rvz_i)$, so that (\ref{eq:numerical_approximation}) is in fact a mixture of PCs, which in turn can be  interpreted as a larger PC \cite{Vergari2020}. Thus, we can convert a learnt intractable model from (\ref{eq:cm_model}) into a PC which facilitates exact inference, that is, `performing exact inference in an accurately approximate model'.

To the best of our knowledge, this simple idea for constructing tractable models has not been explored before. Yet, it delivers astonishing results: on 16 out of 20 commonly used density estimation benchmarks, we set new state of the art (SOTA) among tractable models, outperforming even the most sophisticated PC structure and parameter learners. We also achieve competitive results on image datasets, where in comparison to other PCs, our models produced better samples and often attained better test log-likelihoods.

\section*{Background and Related Work}
In this paper we are interested in \emph{tractable} probabilistic models, i.e.~models which allow for exact and efficient inference. \emph{Probabilistic circuits} (PCs) are a prominent language for tractable models and are defined as follows.

\begin{definition}[Probabilistic Circuit]   
Given a set of random variables $\rmX$, a \emph{probabilistic circuit} (PC)  is based on an acyclic directed graph $\struct$ containing three types of nodes: tractable \emph{distribution nodes} over a subset of the random variables in $\rmX$, e.g.~Gaussian, Categorical, or other exponential families; \emph{sum nodes}, which compute a convex combination (a mixture) of their inputs; and \emph{product nodes}, which compute the product (a factorisation) of their inputs. All leaves of $\struct$ are distribution nodes and all internal nodes are either sum or product nodes.
We assume that $\struct$ has a single root, which is the output of the PC, computing a density over $\rmX$.
\end{definition}

As mentioned in the introduction, PCs can express many different tractable models \cite{Vergari2020}. Of particular interest in this paper are PCs representing \emph{completely factorised} distributions, $p(\rvx) = \prod_{d=1}^D p(x_d)$, and \emph{Chow-Liu trees} (CLTs) \cite{chow1968approximating}, tree-shaped graphical models that can be learnt in polynomial time. CLTs can also be easily converted into PCs \cite{dang2020strudel, di2021random}. We will denote PC structures corresponding to factorised distributions as $\structf$ and CLT structures as $\structclt$. While these structures are arguably simple, we show in the experiment section that continuous mixtures of such PCs outperform all state-of-the-art PC learners on 16 out of 20 common benchmark datasets.

The perhaps most widely known \emph{continuous mixture model} is the \emph{variational autoencoder} (VAE) \cite{kingma2013auto,Rezende2015}, specifying the model density as 
$
    p(\rvx) = \int p(\rvx \cbar \param(\rvz)) \, p(\rvz) \mathrm{d}\rvz
$
where $p(\rvz)$ is an isotropic Gaussian and $p(\rvx \cbar \param(\rvz))$ is typically a fully factorised distribution of Gaussians or Binomials, whose parameters are provided by a neural network $\param(\rvz)$---the so-called decoder---taking $\rvz$ as input. Since the latent code in VAEs is usually relatively high-dimensional, learning and inference is done via amortised inference \cite{kingma2013auto}.

When using $\structf$, our models specify in fact the same model as VAEs, which has originally been introduced by McKay  \cite{Mackay1995} under the name \emph{density network}. Our work essentially re-visits McKay's work, who already mentions: \emph{For a hidden vector of sufficiently small dimensionality, a simple Monte Carlo approach to the evaluation of these integrals can be effective.} For this paper, we considered various numerical integration methods and found that randomised quasi-Monte Carlo (RQMC) performs best for our purposes. Moreover, we also use numerical approximation as a \emph{`compilation approach'}, whereby the continuous mixture is converted into a tractable discrete mixture that sets new state of the art among tractable models in a number of datasets.

Similar approaches to our method are HyperSPNs \cite{shih2021hyperspns} and conditional SPNs \cite{shao2020conditional}, both of which use neural nets to compute the weights of PCs. However, HyperSPNs are primarily a regularisation technique that applies to a single PC, whereas we use neural nets to learn continuous mixtures of PCs. In conditional SPNs the parameters are a function of observed variables, while in this work they are a function of a continuous \emph{latent} space.

\section*{Inference and Learning}
Our model as specified in (\ref{eq:cm_model}) consists of a continuous latent space $\rmZ$ and a given PC structure $\struct$, whose parameters $\param(\rvz)$ are a differentiable function of the latent variables. We will broadly refer to function $\param$ as \emph{decoder} and to the model as a whole as \emph{continuous mixtures}. We use $\cm(\structf)$ and $\cm(\structclt)$ to denote continuous mixtures with factorised structure and CLT structure, respectively. 

Amortised inference \cite{kingma2013auto,Rezende2015} is the de-facto standard way to learn continuous mixture models. In this approach, a separate neural network---the so-called \emph{encoder}---represents an approximate posterior $q(\rvz \cbar \rvx)$. The encoder and decoder are learnt simultaneously by maximising the \emph{evidence lower bound} (ELBO) 
\begin{equation}    \label{eq:elbo}
    \E_{q}[\log p(\rvx \cbar \rvz) - \log q(\rvz \cbar \rvx) + \log p(\rvz)] 
\end{equation}
which is a lower bound of the (marginal) log-likelihood $\log p(\rvx)$ and thus a principled objective. At the same time, maximising the ELBO is moving $q$ closer to the true posterior in Kullback-Leibler sense, hence tightening the ELBO.

In this paper, we investigate numerical integration as an alternative inference and learning method. In particular, we do not require an encoder or any other parametric form of approximate posterior.

\subsection*{Inference via Numerical Integration}
Given some function $f$, a numerical integration method consists of a set of $N$ integration points $\{\rvz_i\}_{i=1}^{N}$ and a weight function $w : \rvz \mapsto \mathbb{R}$ such that the integration error $\varepsilon = \left|\int f(\rvz) \, \mathrm{d}\rvz - \sum_{i=1}^N w(\rvz_i) \, f(\rvz_i) \right|$ is as small as possible. In this paper, we are interested in approximating the density $p(\rvx) = \int p(\rvx \cbar \rvz) \, p(\rvz) \, d\rvz$ of a $\cm$ model, so that the integration error with integration points $\{\rvz_i\}_{i=1}^{N}$ is given as
\begin{align} \label{eq:int_error}
    &\nonumber \varepsilon_{\cm}(\rvx, \{\rvz_i\}_{i=1}^{N}) = \\ 
    &\left| \int p(\rvx \cbar \! \param(\rvz)) \, p(\rvz) \, d\rvz - \sum_{i=1}^N w(\rvz_i) \, p(\rvx \cbar \! \param(\rvz_i)) \right|.
\end{align}

\paragraph{Quadrature Rules} divide the integration domain into sub-intervals and approximate the integrand on these intervals with polynomials, which are easy to integrate. They yield a set of deterministic integration points and weights as a function of the degree of the interpolating polynomial. Common quadrature rules like trapezoidal and Simpson's rule achieve error bounds of $\mathcal{O}(N^{-2})$ and $\mathcal{O}(N^{-4})$, respectively. \emph{Gaussian quadrature rules} go a step further and allow us to take into account the distribution of $\rmZ$; e.g. Gauss-Hermite quadrature is designed for indefinite integrals of the form (\ref{eq:continuous_mixture}) with $\rvz\sim\gN(\bm{0}, \bm{1})$. Gaussian quadratures integrate exactly any polynomial of degree $2N{-}1$ or less, which makes them attractive for general integrands, since by the Weierstrass approximation theorem, any function can be approximated by a polynomial to arbitrary precision under mild regularity conditions \cite{weierstrass1885analytische}. 

\paragraph{Sparse Grids\textnormal{.}}
Unfortunately, quadrature rules do not scale well to high dimensions, since multi-dimensional quadrature rules are usually constructed as the tensor product of univariate rules. If a univariate quadrature rule has an error bound of $\mathcal{O}(N^{-r})$, the corresponding rule in $d$ dimensions with $N^d$ integration points would achieve an error bound $\mathcal{O}(N^{-r/d})$, which degrades quickly due to the curse of dimensionality. Sparse grids \cite{bungartz2004sparse,smolyak1960interpolation} try to circumvent that by a special truncation of the tensor product expansion of univariate quadrature rules. This reduces the number of integration points to $O(N(\log N)^{d-1})$ without significant drops in accuracy \cite{gerstner2010sparse}. However, even if the underlying quadrature formulas are positive, sparse grids can yield negative weights $w(\rvz_i)$. In our preliminary experiments, this property of sparse grids was highly problematic, in particular when inference was used as part of a learning routine.

\paragraph{Monte Carlo (MC)} methods cast the integral as an expectation such that we can compute $\int p(\rvx \cbar \rvz) \, p(\rvz) \, \mathrm{d}\rvz$ as $\E_{p(\rvz)}[p(\rvx \cbar \rvz)] = \sum_{i=1}^N \frac{1}{N} p(\rvx \cbar \rvz_i)$. MC is easy to use and especially attractive for high-dimensional problems since its convergence rate $\mathcal{O}(N^{-1/2})$ is not directly dependent on the problem dimensionality. However, this convergence rate decelerates quickly as one increases the number of integration points, which can be too slow. Quasi-Monte Carlo (QMC) methods \cite{caflisch1998monte} replace the (pseudo-)random sequences of integration points of standard MC with low-discrepancy ones, with the intent of reducing the variance of the estimator and converging faster than $\mathcal{O}(N^{-1/2})$. Crucially, QMC is deterministic, which makes it hard to estimate the integration error in practice. Randomised quasi-Monte Carlo (RQMC) reintroduces randomness into the low-discrepancy sequences of integration points, enabling error estimation (via multiple simulations) and essentially turning QMC into a variance reduction method \cite{l2016randomized}.

In our experiments, we opt for RQMC for two reasons. First, its convergence does not depend directly on the dimensionality, meaning we have more freedom to define the latent space. In fact, we empirically observe that increasing the latent dimensionality does not hurt performance (see additional results in Appendix B). We conjecture that training via numerical integration sufficiently regularises the decoder so that it remains amenable to numerical integration, even when using a relatively large latent dimensionality in the order of tens. Second, in contrast to Monte Carlo, RQMC produces integration points of lower variance \cite{l2016randomized}, which often facilitates training (see Appendix G); and, in comparison to QMC, RQMC reintroduces randomness into the sets of integration points which helps to avoid overfitting.

\subsection*{Learning the Decoder}
In principle, continuous mixture models can be learnt in many ways, such as amortised variational inference \cite{kingma2013auto} or adversarial training \cite{goodfellow2014generative}. However, these methods do not encourage the decoder to be amenable to numerical integration, and thus their approximation by (or compilation to) a mixture of tractable models is subpar. Perhaps not surprisingly, we find that training via numerical integration is the best way to learn and extract expressive mixtures of tractable probabilistic models. We compare numerical integration and variational inference in Figure~\ref{fig:average-trend} and Appendix C.

Training via integration simply amounts to selecting a set of integration points $\{\rvz_i\}_{i=1}^{N}$ using any numerical integration method of choice and, for some training data $\{\rvx_j\}_{j=1}^{M}$, maximising the log-likelihood (LL) of the approximate model with respect to the decoder parameters:
\begin{equation} \label{eq:loss}
    LL = \sum_{j=1}^M \log \sum_{i=1}^N w(\rvz_i) \, p(\rvx_j \cbar \! \param(\rvz_i)).
\end{equation}
For $N \rightarrow \infty$, this objective converges to the exact log-likelihood of the continuous mixture, and for $1 \ll N < \infty$ it serves as a reasonable approximation.

In particular, when using (RQ)MC methods the inner sum of (\ref{eq:loss}) is unbiased, yielding a \emph{negatively biased} estimate of the true log-likelihood due to Jensen's inequality---i.e.~a `noisy lower bound'---justifying (\ref{eq:loss}) as training objective for similar reasons as the variational ELBO (\ref{eq:elbo}). However, (\ref{eq:loss}) should not be confused with the standard ELBO as it does not involve a posterior approximation and, unlike the ELBO, it becomes tight for $N \rightarrow \infty$. We further note that (RQ)MC methods to estimate the log-likelihood of latent variable models is not a new idea and has been widely explored either directly \cite{Mackay1995} or to improve ELBO techniques \cite{burda2015importance,mnih2016variational,buchholz2018quasi}. In this paper, however, we are specifically interested in combining continuous mixtures with \emph{tractable models} via numerical integration, as this direction has been explored rather little.

One can interpret our model in two distinct ways. The first is to interpret it as a `factory' method, whereby each fixed set of latent variables $\{\rvz_i\}_{i=1}^{N}$ yields a tractable model supporting exact likelihood and marginalisation, namely a PC (trivially a mixture of PCs is a PC). The second is to take it as an intractable continuous latent variable model, but one that is amenable to numerical integration. At test time, we are free to choose the set of integration points, possibly changing the integration method and number of integration points $N$ if more or less precision is needed.

\subsection*{Efficient Learning} \label{sec:eff}
When using a neural network to fit $p(\rvx \cbar \! \param(\rvz))$, computing the backward pass with respect to the log-likelihood objective in (\ref{eq:loss}) can be memory intensive. We can circumvent that by first finding the $K$ integration points that are most likely to have generated each training instance. We do so via a forward pass (with no gradient computation) that allows us to identify the $K$ values of $\rvz$ (among the $N$ points $\{\rvz_i\}_{i=1}^{N}$ defined by the integration method) that maximise $w(\rvz) \, p(\rvx_j \cbar \! \param(\rvz))$ for each $\rvx_j$ in a training batch. We then run backprop to optimise a cheaper estimate of the log-likelihood (\ref{eq:loss2}) which only requires $K$ values $\{\rvz_{ij}\}_{i=1}^{K}$ for each $\rvx_j$, instead of $N$
\begin{equation} \label{eq:loss2}
    LL' = \sum_{j=1}^M \log \sum_{i=1}^K w(\rvz_{ij}) \, p(\rvx_j \cbar \! \param(\rvz_{ij})).
\end{equation}
For $K \ll N$ this results in large improvements in memory efficiency. In our experiments, we only use this approximation for non-binary image datasets for which $K=1$ was already sufficient to get good results.

\subsection*{Latent Optimisation}    \label{sec:learning_integration_points}
As mentioned in the last section, once the decoder is trained, we can compile a continuous mixture into a PC by fixing a set of integration points (selected via some integration method), leading to a discrete mixture of PCs.
However, rather than using a fixed integration scheme, one might also treat the integration points as `parameters' and optimise them. More precisely, given training instances $\{\rvx_j\}_{j=1}^{M}$, we might find suitable $\{\rvz_i\}_{i=1}^{N}$ by maximising the log-likelihood:
\begin{align} \label{eq:lo_obj}
    \argmax_{\{\rvz_i\}_{i=1}^{N}} \sum_{j=1}^M \log \sum_{i=1}^N w(\rvz_i) \, p(\rvx_j \cbar \! \param(\rvz_i)).
\end{align}

\binarydatasetstable

Due to the similarity to \cite{bojanowski2018optimizing, park2019deepsdf} we refer to this technique as \emph{Latent Optimisation (LO)}. There are, however, a few differences in spirit: in \cite{bojanowski2018optimizing, park2019deepsdf} the decoder and individual latent representations, one for each training instance, are jointly learnt to minimise the reconstruction error. The latent space is regularised to follow a Gaussian prior, but otherwise is devoid of any probabilistic interpretation. 

In our approach, however, the goal is an accurate yet compact approximation to the true continuous mixture model. Training the decoder and integration points together would lead to overfitting, meaning that the decoder would not perform well with different integration points or methods. For that reason, we only optimise the integration points as a \emph{post-processing step}, i.e.~the \emph{decoder parameters remain fixed} throughout the optimisation process.

Our LO approach can also be interpreted as a way to learn (or compile) PCs, using continuous mixtures as a teacher model or regularizer. In our experiments, we see that this approach is remarkably effective. LO yields test log-likelihoods similar to that obtained with RQMC while using considerably fewer integration points, and thus delivering smaller PCs.

\section*{Experiments}

We evaluated our method on common benchmarks for generative models, namely 20 standard density estimation datasets \cite{lowd2010learning,van2012markov,bekker2015tractable} as well as 4 image datasets (Binary MNIST \cite{larochelle2011neural}, MNIST \cite{lecun1998gradient}, Fashion MNIST \cite{xiao2017fashion} and Street View House Numbers (SVHN) \cite{netzer2011reading}). All models were developed in python 3 with PyTorch \cite{paszke2019pytorch} and trained with standard commercial GPUs. We used RQMC in all experiments ($w(\rvz_i) = 1/N$), with samples generated by QMCpy \cite{QMCPy}. Further experimental details can be found in Appendix A, and our source code is available at \url{github.com/alcorreia/cm-tpm}.

\subsection*{Standard Density Estimation Benchmarks}
As a first experiment, we compared \emph{continuous mixtures of factorisations}, denoted $\cmf$, and \emph{continuous mixtures of CLTs}, denoted $\cmclt$, as density estimators on a series of 20 standard commonly used benchmark datasets.
In this set of experiments, we fixed the mixing distribution $p(\rvz)$ to a 4-dimensional standard Gaussian and used $N=2^{10}$ integration points during training. For the decoder we used 6-layer MLPs with LeakyReLUs activations.

At test time, the trained models can be evaluated with any number of integration points $N$, yielding a mixture of PCs and consequently indeed a standard PC \cite{Vergari2020}. In Table \ref{tab:20datasets} we report the test log-likelihoods for $\cm$ models with $2^{13}$ components, averaged over the 5 random seeds, and for current SOTA PCs. Our results set SOTA log-likelihoods for tractable models on 16 out of 20 datasets and are competitive on the remaining 4. For each dataset, we ranked the performance of the considered models from 1 to 4, and reported the average rank at the bottom of the first half of the table. In particular, we notice a substantial gap in performance between $\cmclt$ and $\cmf$ for the datasets \texttt{accidents}, \texttt{ad}, \texttt{dna} and \texttt{pumbs}, which are known to be highly structured. We emphasise that we used the exact same hyperparameters for all datasets, and hence our SOTA results do not stem from extensive tuning efforts.

In Figure \ref{fig:average-trend}, we plot the performance of our models relative to the best results in Table~\ref{tab:20datasets}, averaged over all 20 datasets. This shows the effect of the number of integration points at test time and indicates $\cmclt$ generally outperforms $\cmf$, especially for small numbers of integration points.

\subsection*{Latent Optimisation}
\begin{figure}
\centering
\includegraphics[width=\linewidth]{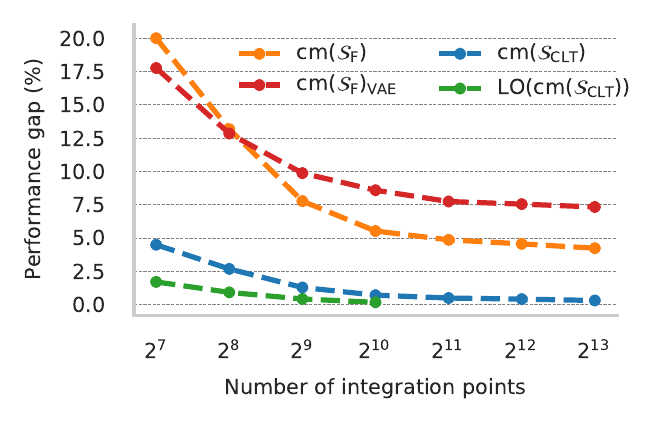}
\caption{Relative performance gap to the best log-likelihood in Table~\ref{tab:20datasets} as a function of the number of integration points at test time and averaged over all 20 datasets. Latent Optimisation is run (on purpose) for fewer number of integration points yet performs best. Lower is better.}
\label{fig:average-trend}
\end{figure}

Next, we showcase the effect of Latent Optimisation (LO), i.e.~learning the integration points after having fit the decoder, as previously discussed.
More precisely, we run LO to search for a good set of integration points for a trained $\cmclt$ by maximising (\ref{eq:lo_obj}). We show the results under $\locmclt$ in Table~\ref{tab:20datasets} and Figure~\ref{fig:average-trend}.

We see that $\locmclt$ achieves essentially the same performance as $\cmclt$ but with  \emph{8 times fewer integration points,} leading to much smaller PCs. However, as can be seen in Figure~\ref{fig:average-trend}, for a large number of integration points, (RQ)MC estimates already have low-variance and there is little room for improvement with LO. In fact, in this setting LO is prone to overfitting. This can be expected when the number of integration points becomes too large (in comparison to the training data), since they are treated as trainable parameters. For that reason, we limit our LO experiments to $2^{10}$ integration points in all datasets in Table~\ref{tab:20datasets} and~\ref{tab:bmnist}.

\subsection*{Comparison with Variational Learning}
As the main idea of this paper is to relate continuous mixtures to PCs via numerical integration, we used (\ref{eq:loss}) to train our models thus far. However, the training of a continuous mixture model and its subsequent conversion to a PC are orthogonal to each other, and we might also use VAE training to learn continuous mixtures. That is, we can train a standard VAE with a \emph{small latent dimension}, maximising the ELBO (\ref{eq:elbo}) with amortised variational inference, and then convert the resulting model into a PC using RQMC or LO.

We evaluated this alternative with the same experimental setup and 20 density estimation datasets. We learnt continuous mixtures with VAE training \cite{kingma2013auto} and subsequently numerically integrated the resulting model, denoted $\cmvae$, with RQMC. Note that we used exactly the same architecture for both $\cmf$ and $\cmvae$, with the only difference being the training method. Figure~\ref{fig:average-trend} shows that $\cmvae$ is outperformed by $\cmf$, even for large numbers of integration points at test time. We offer two explanations for this result: First, it might be that $\cmvae$ models are less amenable to our numerical approximation techniques and that their true log-likelihood is actually higher, or conversely, that models \emph{trained} with numerical integration are more amenable to numerical integration at test time. Second, it might also be that numerical integration leads to better model training for \emph{small latent dimensionality}. We provide affirmative evidence for the latter in  Appendix C. However, evidently, VAE training is superior for large latent dimensionality, as numerical integration degrades quickly in high dimensional spaces. See Appendix C for more comprehensive experimental details and further results.

\subsection*{Binary MNIST}
\bmnisttable
We also evaluated our models on Binary MNIST \cite{larochelle2011neural}. We followed the same experimental protocol as in the previous experiments, except that we employed a larger latent dimensionality of 16 and increased the number of integration points during training to $2^{14}$. We did \emph{not} use convolutions and stuck to 6-layer MLPs. We ran $\cm(\structf)$ and $\cm(\structclt)$ and applied LO to both final models for up to 50 epochs, using early stopping on the validation set to avoid overfitting. Table \ref{tab:bmnist} shows that $\cm(\structclt)$ outperforms $\cm(\structf)$ overall and that LO is remarkably effective when few integration points are used.
In Table~\ref{tab:imagedatasets}, we compare our models against Einets \cite{peharz2020einsum}, large scale PCs designed to take advantage of GPU accelerators. We considered Einets of different sizes\footnote{For Binary MNIST, `Small Einet' and `Big Einet' had respectively 5 and 84 million parameters; for MNIST, 11 and 90 million parameters; and for SVHN, 28 and 186 million parameters.} and used Poon-Domingos architectures \cite{poon2011sum}, which recursively partition the image into contiguous square blocks.

\begin{figure}
\centering
\begin{subfigure}{.32\linewidth}
    \centering
    \includegraphics[width=\textwidth]{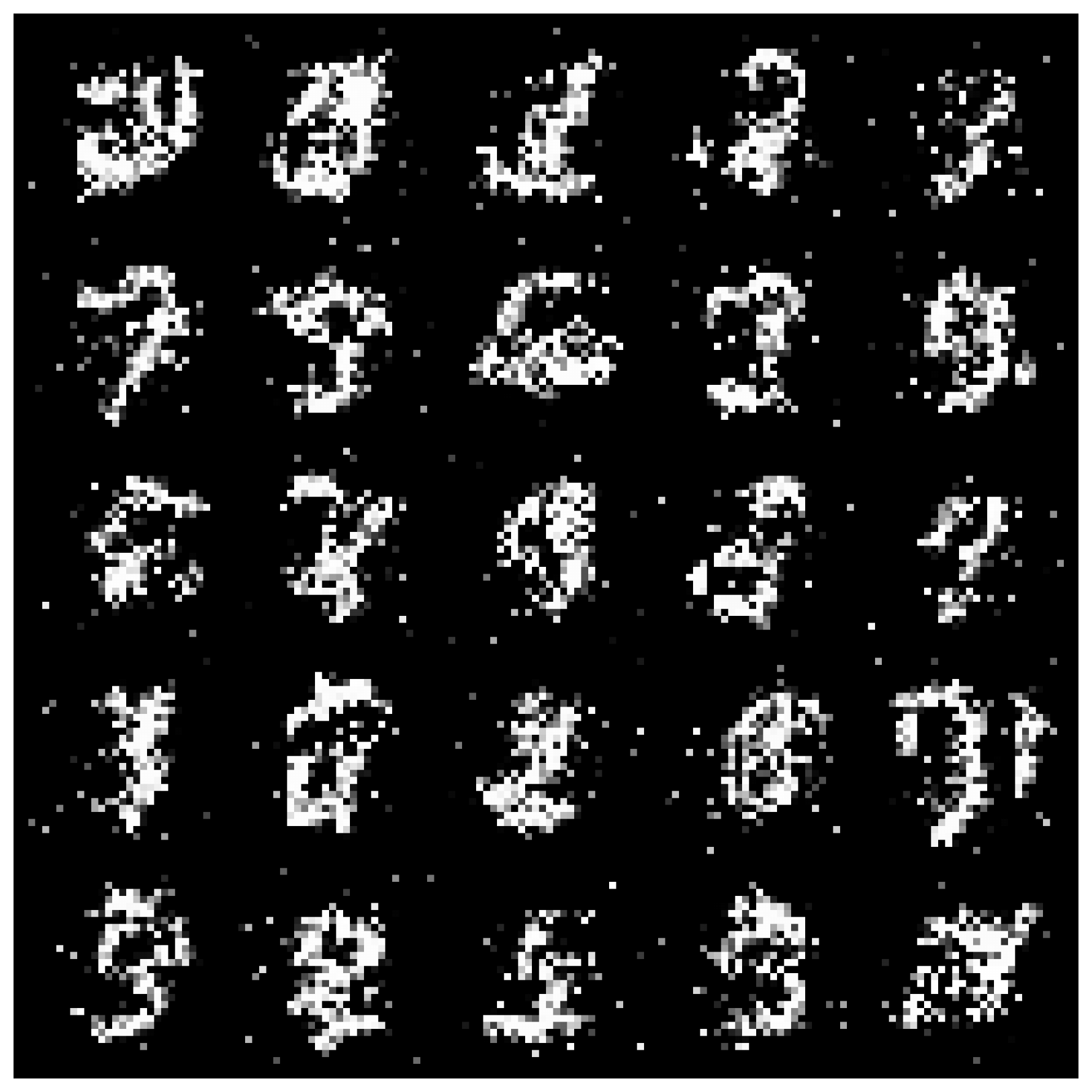}
\end{subfigure}
\begin{subfigure}{.32\linewidth}
    \centering
    \includegraphics[width=\textwidth]{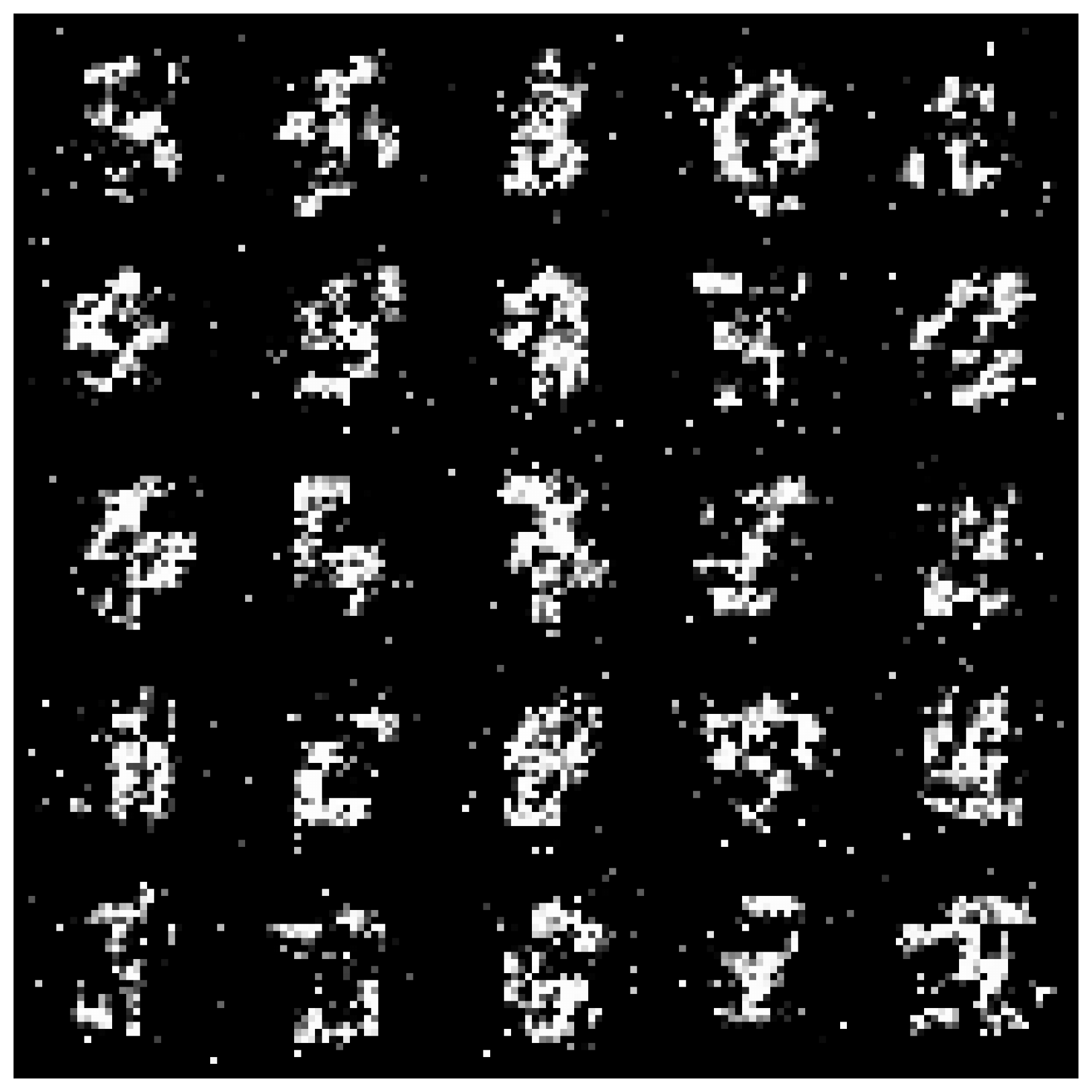}
\end{subfigure}
\begin{subfigure}{.32\linewidth}
    \centering
    \includegraphics[width=\textwidth]{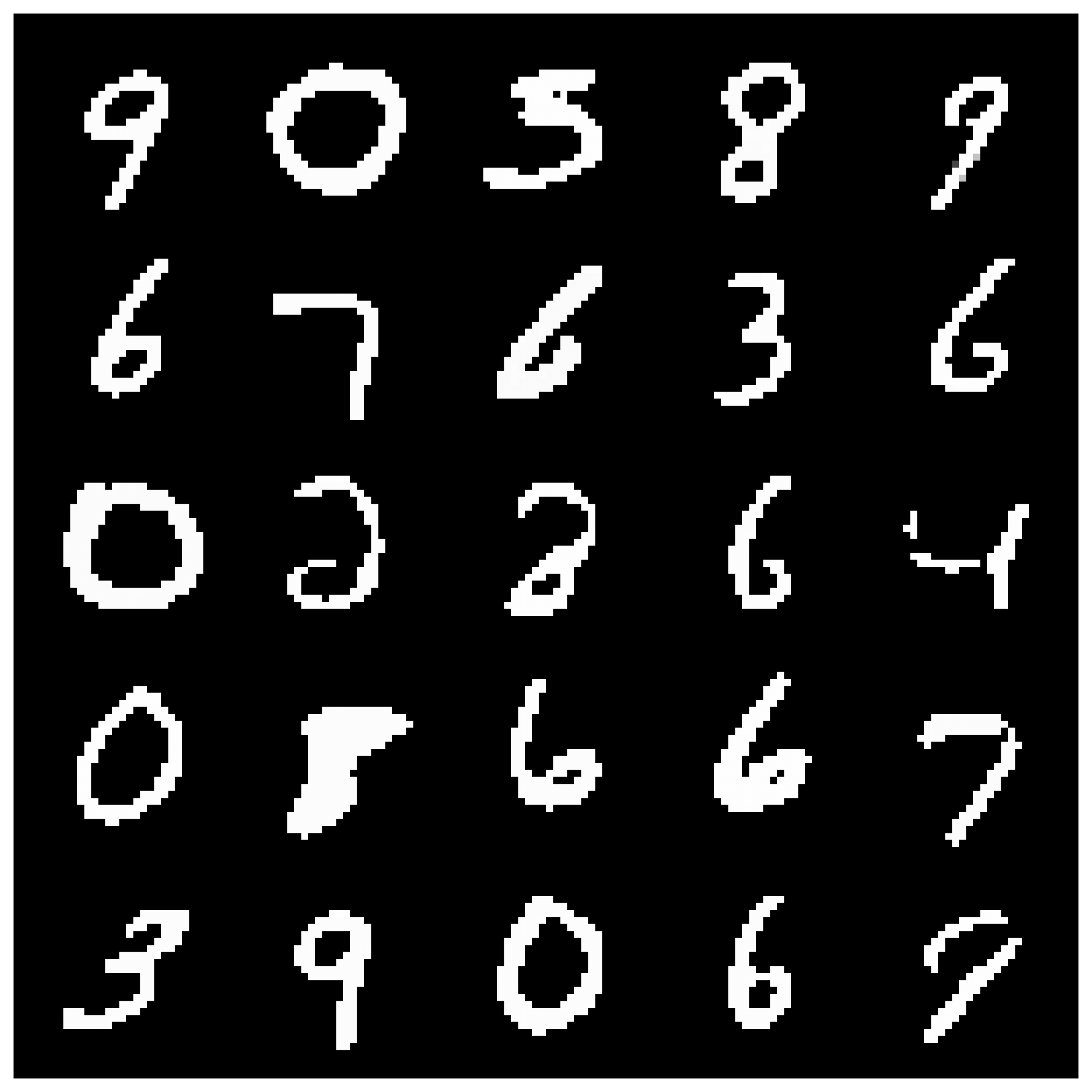}
\end{subfigure}

\smallskip
\begin{subfigure}{.32\linewidth}
    \centering
    \includegraphics[width=\textwidth]{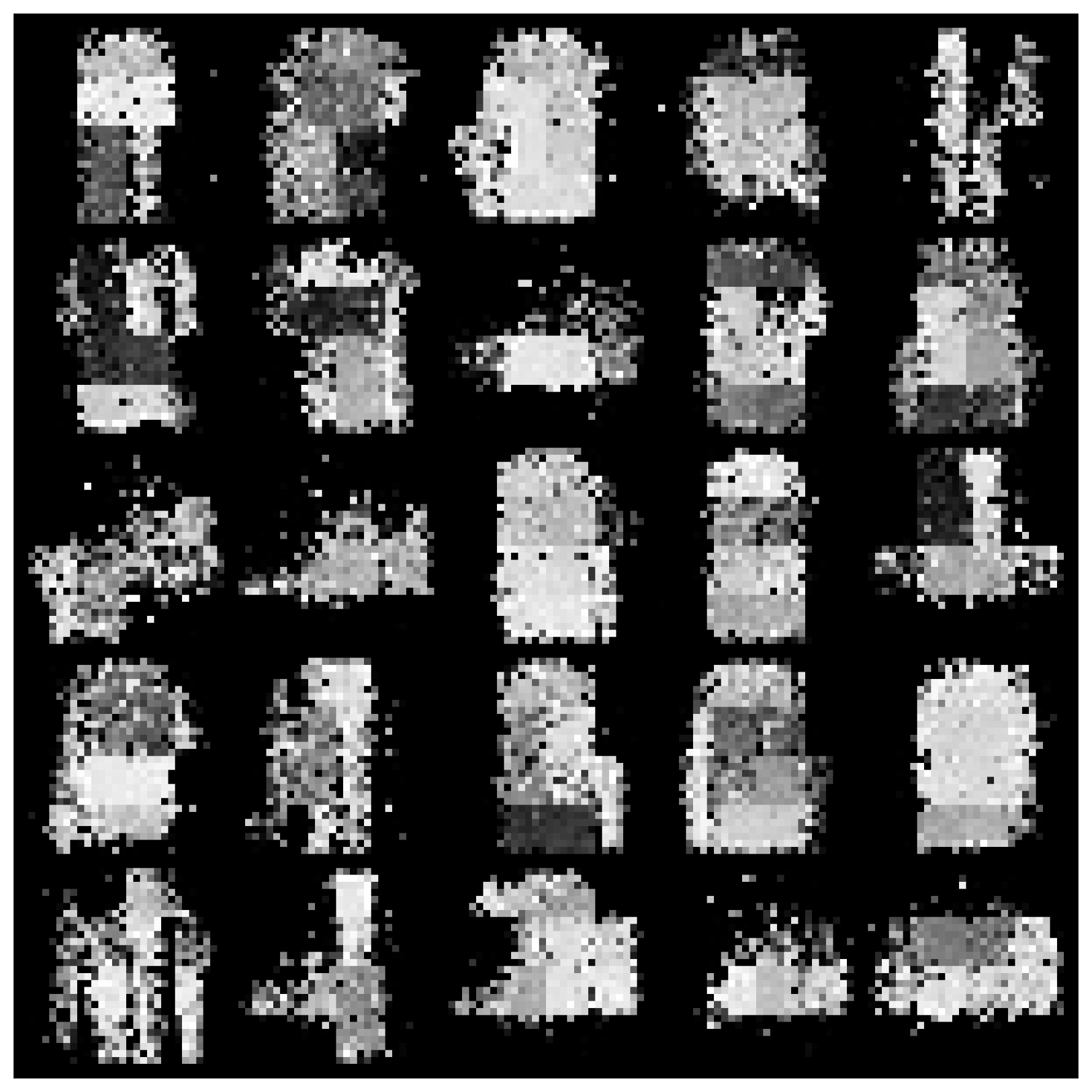}
\end{subfigure}
\begin{subfigure}{.32\linewidth}
    \centering
    \includegraphics[width=\textwidth]{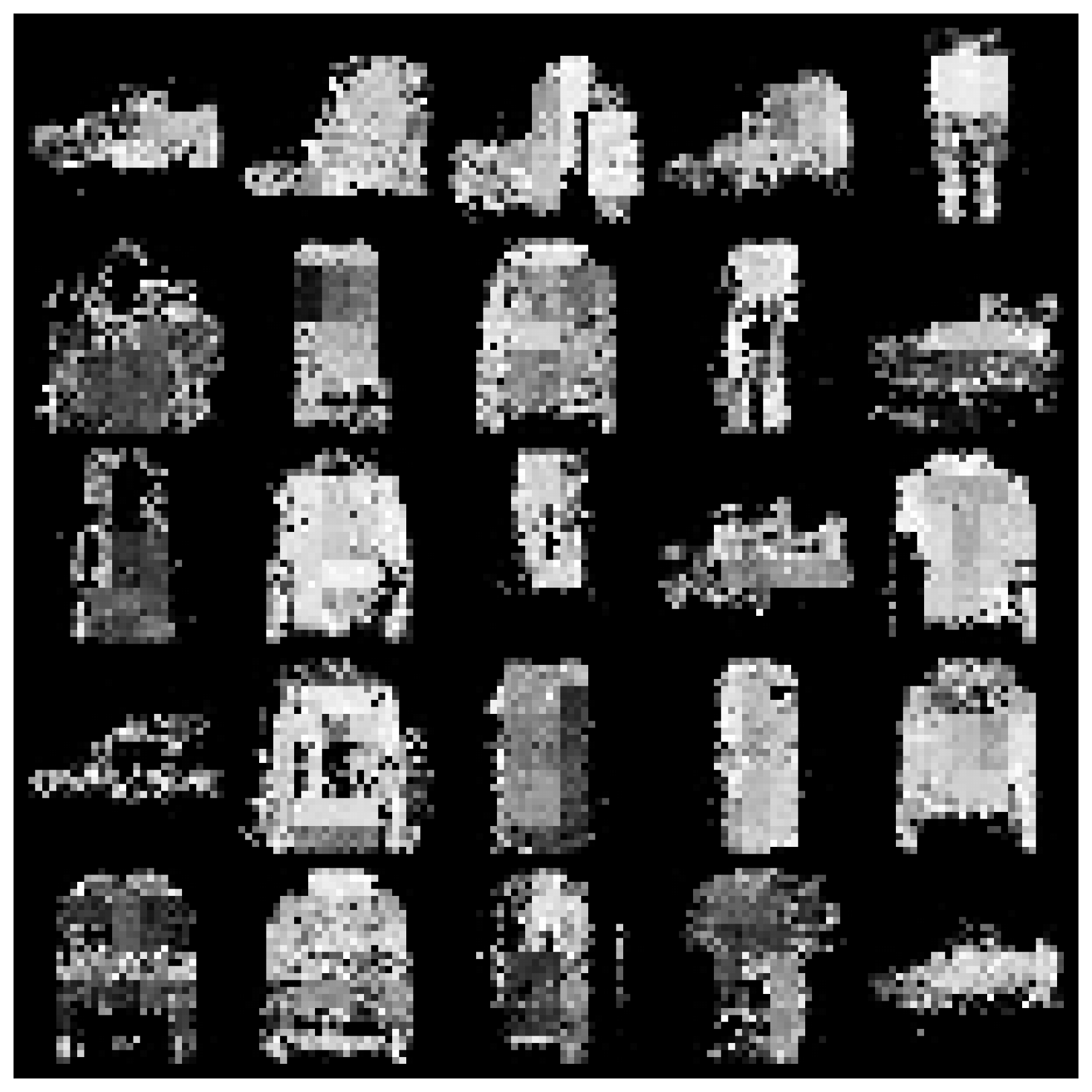}
\end{subfigure}
\begin{subfigure}{.32\linewidth}
    \centering
    \includegraphics[width=\textwidth]{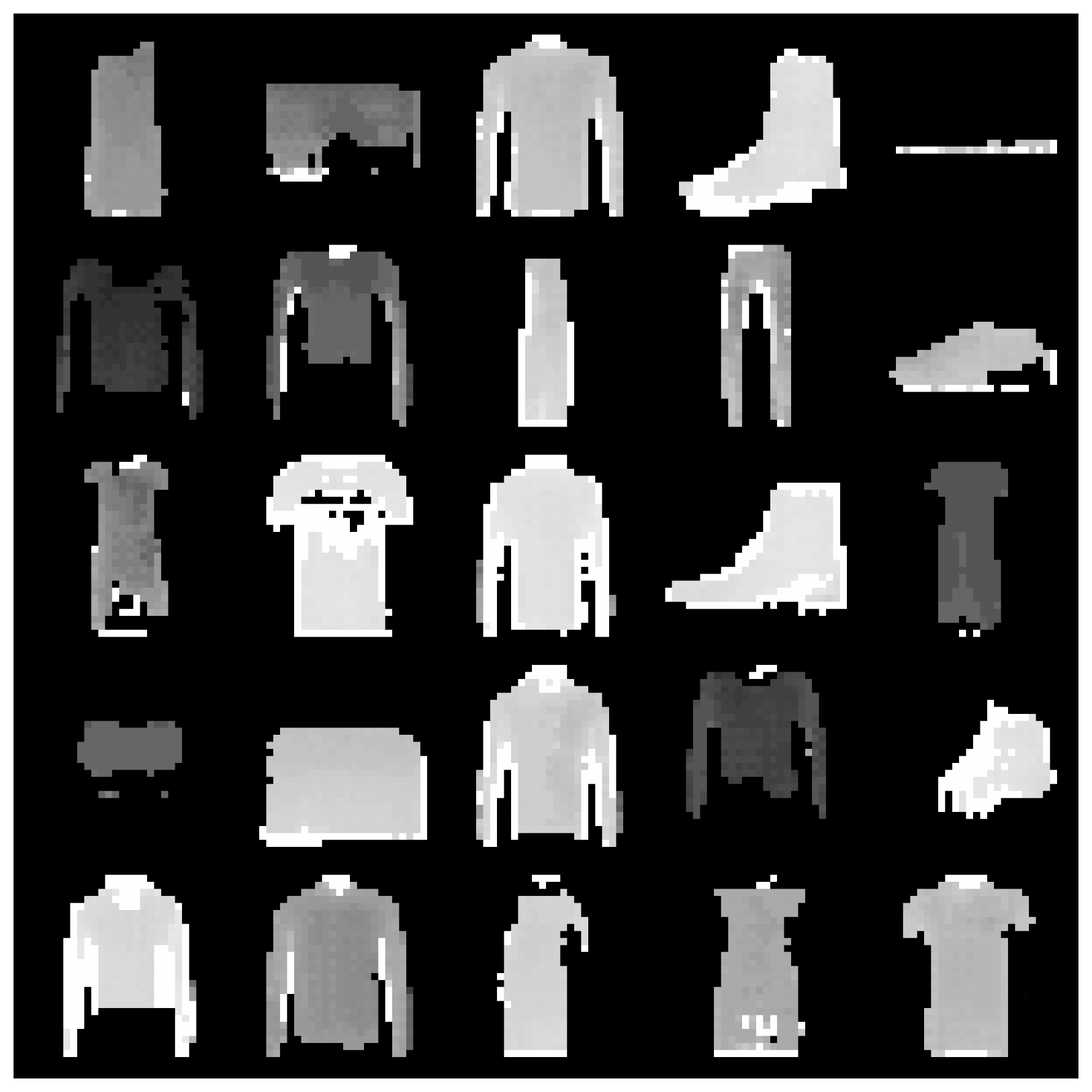}
\end{subfigure}

\smallskip
\begin{subfigure}{.32\linewidth}
    \centering
    \includegraphics[width=\textwidth]{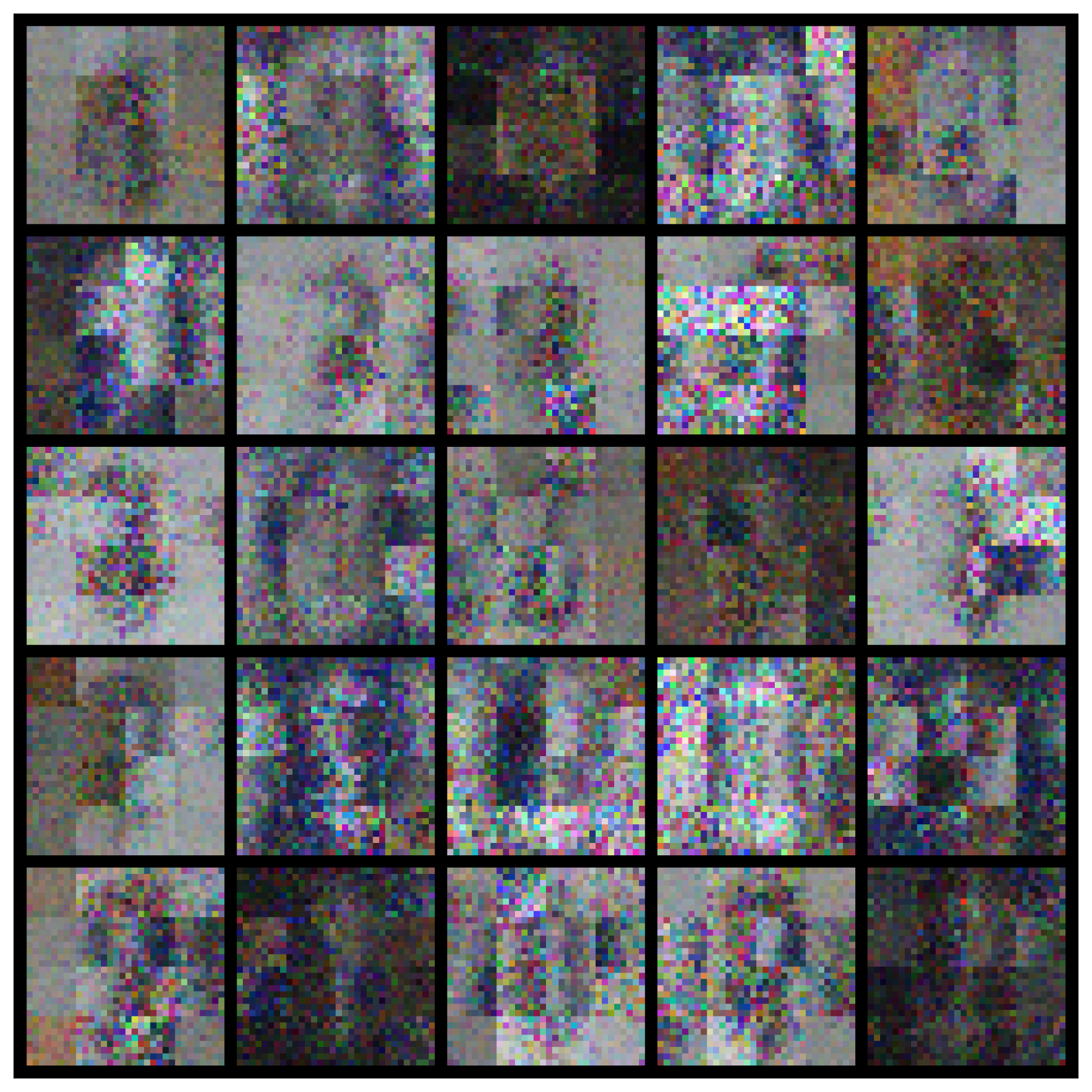}
\end{subfigure}
\begin{subfigure}{.32\linewidth}
    \centering
    \includegraphics[width=\textwidth]{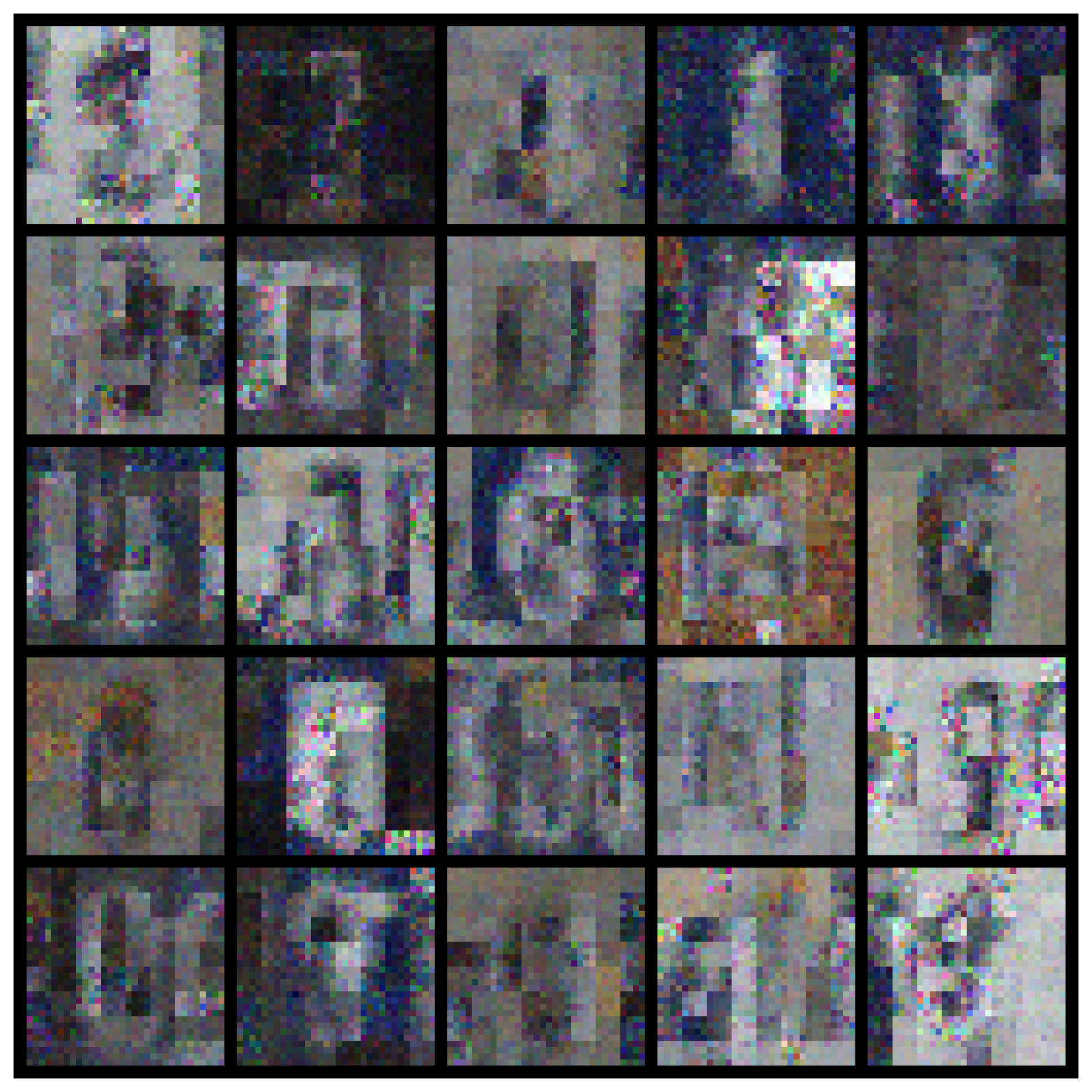}
\end{subfigure}
\begin{subfigure}{.32\linewidth}
    \centering
    \includegraphics[width=\textwidth]{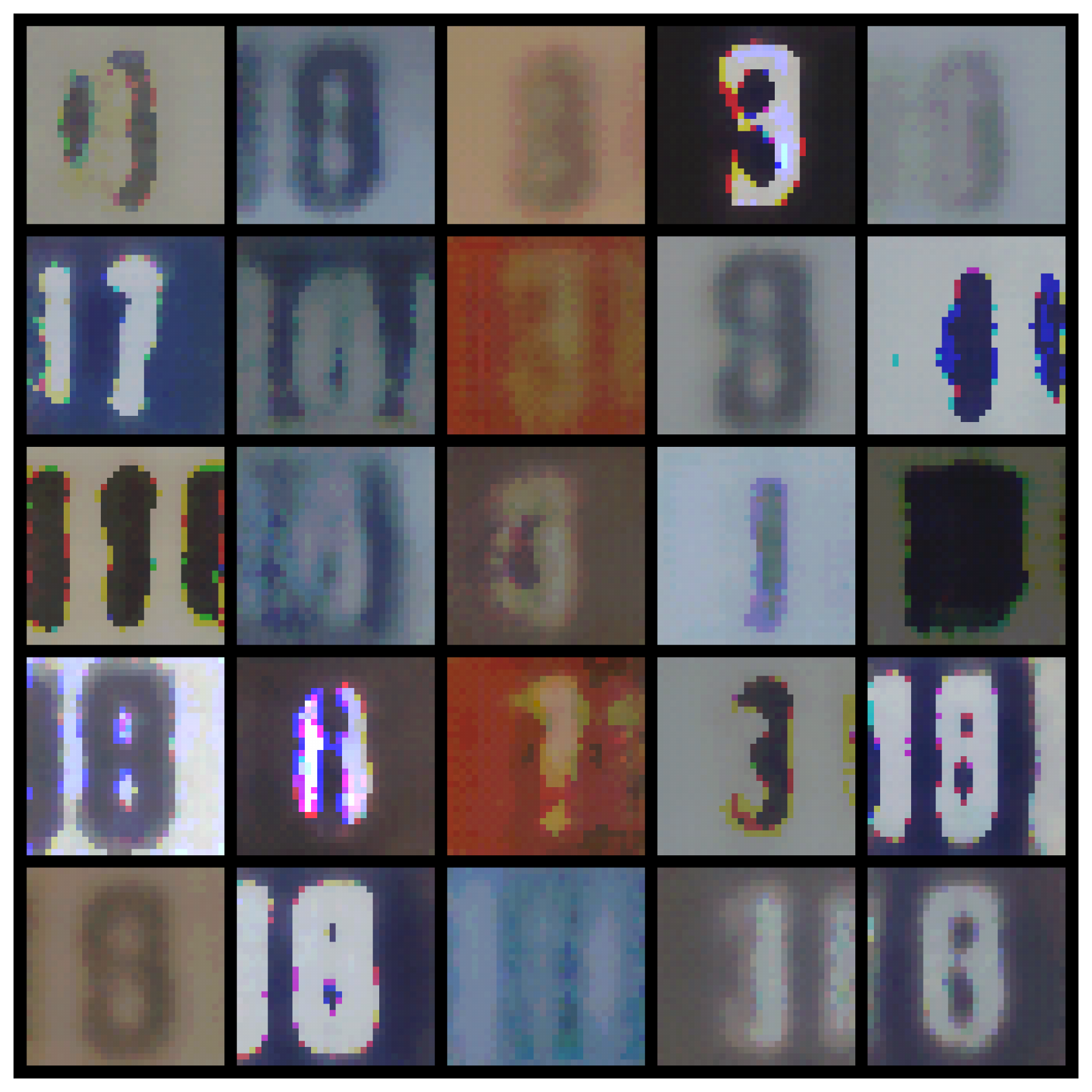}
\end{subfigure}
\caption{Samples from `Small Einet' (left column), `Big Einet' (middle column) and $\cmf$ (right column).}
\label{fig:samples}
\end{figure}

\subsection*{Image Datasets}
\imagedatasetstable
For non-binary image data we used a convolutional architecture similar to that of DCGAN \cite{radford2015unsupervised} but also included residual blocks as in \cite{van2017neural}. For both MNIST and SVHN data, we used the same architecture and trained $\cmf$ models with 16 latent dimensions and $K{=}1$ (see Efficient Learning).
Once more, we compared against Einets of different sizes. For both Einets and our models, pixels were modelled with 256-dimensional categorical distributions\footnote{See Appendix~\ref{sec:extra_res} for results using Normal distributions instead.}. In all cases, we use no auxiliary clustering algorithm to assign datapoints to components of a sum node. Such a pre-processing step is applicable to any method and does not add to the analysis in this paper. That is why sample quality in Figure~\ref{fig:samples} is worse than that reported in \cite{peharz2020einsum}, where images were clustered and a dedicated Einet was trained on each cluster.

As seen in Table~\ref{tab:imagedatasets}, continuous mixtures outperform Einets in all image datasets but SVHN. That is remarkable since our models are extremely compact with the decoder given by a light convolutional architecture of approximately 100K free parameters for MNIST data, and 300K for SVHN; orders of magnitude smaller than the competing Einets. Moreover, our models also achieve better sample quality. In Figure~\ref{fig:samples}, we see samples from $\cmf$ are clearly sharper and do not suffer from intense pixelation like those from Einets.

\subsection*{Other Tractable Queries}
Our models also support efficient marginalisation, since the discrete approximation obtained via numerical integration is a PC in itself. That allows us to handle missing data and perform tasks like inpainting out-of-the-box, without any extra modelling steps. While this is not the focus of the paper---these queries are well-established for PCs, and it is not surprising that our models support them as well---we do present a couple of interesting experiments in Appendix F. We successfully trained our model on MNIST with substantial parts of the data missing. Note that such a training procedure is delicate for intractable models like VAEs. Furthermore, we included inpainting experiments on Binary MNIST, MNIST and Fashion MNIST, i.e. reconstructing missing data at test time, using a model trained on complete data.

\section*{Discussion}
Our experiments show that continuous mixtures of PCs (or actually their discrete approximations yielding again PCs) outperform most previous PC methods on several datasets. At first this might appear surprising. For fixed $N$, a discrete mixture with $N$ components is at least as expressive as a continuous mixture approximated by $N$ integration points, since the former has mixture components with free (private) parameters, while the latter has components which are determined via a shared neural network, and thus entangled in a complex way. Moreover, the PCs of previous works have been deeper and used more sophisticated architectures than our continuous mixtures. A comparison between our models and discrete mixtures with the same shallow structures is deferred to Appendix D.

The main reason for the efficacy of our approach might be the continuity of the neural network, which topologically relates the latent and observable space, thus identifying some underlying latent structure; this is in fact one of the attractive and widely appreciated properties of VAEs. Yet, the effect of continuity on generalisation has not been much studied, and our results provide an interesting pointer in this regard. Why does continuity promote generalisation, or act as some form of regularisation? For the one, there might be an Occam's razor effect at work, since our models are usually much smaller in terms of free parameters, yet they are expressive due to the non-linear nature of neural nets. Furthermore, dependence among components introduced via the latent space might effectively facilitate learning by avoiding redundant or `dead' components, which have been observed in vanilla PCs \cite{dangsparse}.

These results have two important consequences for future work on tractable probabilistic models: (i) continuous latent spaces seem to be a valuable tool for learning tractable models and (ii) PCs in general seem to have untouched potential not yet exploited by existing learning methods.

\section*{Conclusion}
In this paper, we have investigated the marriage of continuous mixtures and tractable probabilistic models. We have observed that, even with simple structures and standard numerical integration methods, continuous latent variables facilitate the learning of expressive PCs, as confirmed by SOTA results on many datasets. Moreover, we have proposed latent optimisation as an effective way to derive competitive mixture models with relatively few components (integration points). We believe continuous mixtures are a promising tool for learning tractable probabilistic models as well as developing new hybrid inference models \cite{peharz2019hierarchical}.

Our model is not without limitations, however. In particular, numerical integration is a computationally expensive training approach, and we assume fixed PC structures (independent of the latent variables) that have to be defined or learnt a priori. These two issues are promising avenues for future work, especially with extensions to more complex structures, like HCLTs \cite{liu2021tractable}.

\section*{Acknowledgements}
We thank the Eindhoven Artificial Intelligence Systems Institute (EAISI) for its support. This research was supported by the Graz Center for Machine Learning (GraML). This work was partially funded by the EU European Defence Fund Project KOIOS (EDF-2021-DIGIT-R-FL-KOIOS). We also thank Guy Van den Broeck and team for their feedback that helped us improve the paper.

\bibliography{cmtpm}
\clearpage
\appendix
\onecolumn

\section{Model Description} \label{sec:model}

\subsection{Latent Space} We use latent variables of dimension $d{=}4$ for the 20 density estimation benchmarks \cite{lowd2010learning,van2012markov,bekker2015tractable}, and $d{=}16$ for the image datasets, Binary MNIST \cite{larochelle2011neural}, MNIST \cite{lecun1998gradient}, Fashion MNIST \cite{xiao2017fashion} and Street View House Numbers (SVHN) \cite{netzer2011reading}.
In all cases, the latent space was distributed as a standard normal, $\rvz \sim \gN(0, 1)^d$, and integration points $\{{\rvz_i}\}_{i=1}^N$ were generated via a low-discrepancy lattice sequence in a randomised quasi-Monte Carlo (RQMC) framework. In practice, quasi-Monte Carlo methods are designed for sequences $\{{\rvu_i}\}_{i=1}^N$ mimicking a uniform distribution $\gU(0, 1)^d$. However, for most commonly used distributions, it is easy to map uniform samples to the random variable of interest. For instance, for $\rmZ$ distributed as a multivariate normal with mean $\mu$ and covariance $\Sigma$, we can construct a sequence $\{{\rvz_i}\}_{i=1}^N$ from $\{{\rvu_i}\}_{i=1}^N$ as
\begin{equation}
    \rvz_i = \Phi^{-1}(\rvu_i)\Sigma^{1/2} + \mu,
\end{equation}
where $\Phi^{-1}$ is the inverse CDF of a standard normal distribution.
We chose a standard normal prior to facilitate the comparison to plain variational autoencoders \cite{kingma2013auto}. We have experimented with other distributions in preliminary studies, especially $\gU(0, 1)^d$, but have not observed significant differences in performance.

\subsection{Decoder architecture}
In all experiments, only two architectures were considered depending on the type of data. For binary datasets, the decoder was a multi-layer perceptron (MLP) with 6 layers of progressively increasing hidden size (from latent dimension to input dimension). For non-binary image datasets we used a convolutional architecture similar to that of DCGAN \cite{radford2015unsupervised}, but we also included residual convolutional blocks as in \cite{van2017neural}. The hyperparameters of these architectures were kept the same for all experiments, only varying input and output dimensions to accommodate different data dimensionality or parametrisation at the leaves (normal vs. categorical distributions). In all cases, we used LeakyReLU activations and batch normalisation \cite{ioffe2015batch}.

\subsection{Structure in Tractable Probabilistic Models}
Our method consists of a continuous mixture of tractable probabilistic models. The parameters of these models are a function of the continuous latent variables, and thus learnt, but their structure is assumed and has to be defined a priori. As discussed in the main text, we consider two types of structure: fully factorised distributions $\structf$ and Chow-Liu Trees $\structclt$, which are learnt a priori using the training data with the classic algorithm \cite{chow1968approximating} and kept fixed thereafter. Ultimately, both structures are a composition of univariate distributions. For all these models, we use Bernoulli distributions with learnable parameters for binary data, and 256-dimensional categorical distributions for non-binary image data. 
We also considered treating images as continuous data, in which case we use normal distributions at the leaves and normalise the images to $[0, 1]$ as is common in the literature \cite{theis2016note}: we added uniform noise to each image and divided pixel values by 256. See Appendix~\ref{sec:cont_image} for results in this last setting.

In the case of Einets \cite{peharz2020einsum}, we used the publicly available implementation by Peharz et al.\footnote{https://github.com/cambridge-mlg/EinsumNetworks.} with the exact same leaf distributions as just described above, and trained via EM. We used Poon-Domingos architectures \cite{poon2011sum} of different sizes: a `Small Einet' partitioning the image space into 16 ($4 \times 4$) contiguous square blocks and a `Big Einet' partitioning the image space into 49 ($7 \times 7$) blocks in the case of MNIST and into 64 ($8 \times 8$) blocks in the case of SVHN.

\subsection{Training}
We train via numerical integration with randomised quasi-Monte Carlo. At each training step, we generate a new low-discrepancy lattice sequence via \emph{random shifting}. That is, we generate a lattice sequence $\{{\rvu_i}\}_{i=1}^N$ with each $\rvu_i$ in $(0, 1)^d$ and, at each iteration, we shift it by adding a single random point to all other points in the sequence modulo 1. 
\begin{equation*}
    \rvu_i' = \rvu_i + \rvu_{shift} (\bmod 1), \enspace \rvu_{shift} \sim \gU(0, 1)^d.
\end{equation*}
After shifting, $\{{\rvu_i}\}_{i=1}^N$ is mapped to $\{{\rvz_i}\}_{i=1}^N$ via a suitable transformation as described above.
This is the simplest form of randomisation in RQMC and works well in practice. We recommend \cite{l2016randomized} for a good overview of RQMC methods. 
We train all models (including $\cmvae$) for 300 epochs using Adam \cite{kingma2014adam} with learning rate of $1e^{-3}$, $\beta_1{=}0.9$ and $\beta_2{=}0.999$. 
We also track performance on the validation set and use early-stopping with a patience of 15 to avoid overfitting.

In the experiments with binary datasets, we used a batch size of 128 and $2^{10}$ integration points to train both $\cmf$ and $\cmclt$. We report results across 5 random seeds, which we set to $\{0, 1, 2, 3, 4\}$. At test time, RQMC sequences are still stochastic, so we evaluate models with random seed set to 42 for reproducibility.
In the experiments with image datasets, we used a larger batch size of 512 and also increased the number of integration points to $2^{14}$.

\section{Effect of Latent Space Dimensionality}
\begin{figure*}[!h]
    \centering
    \includegraphics{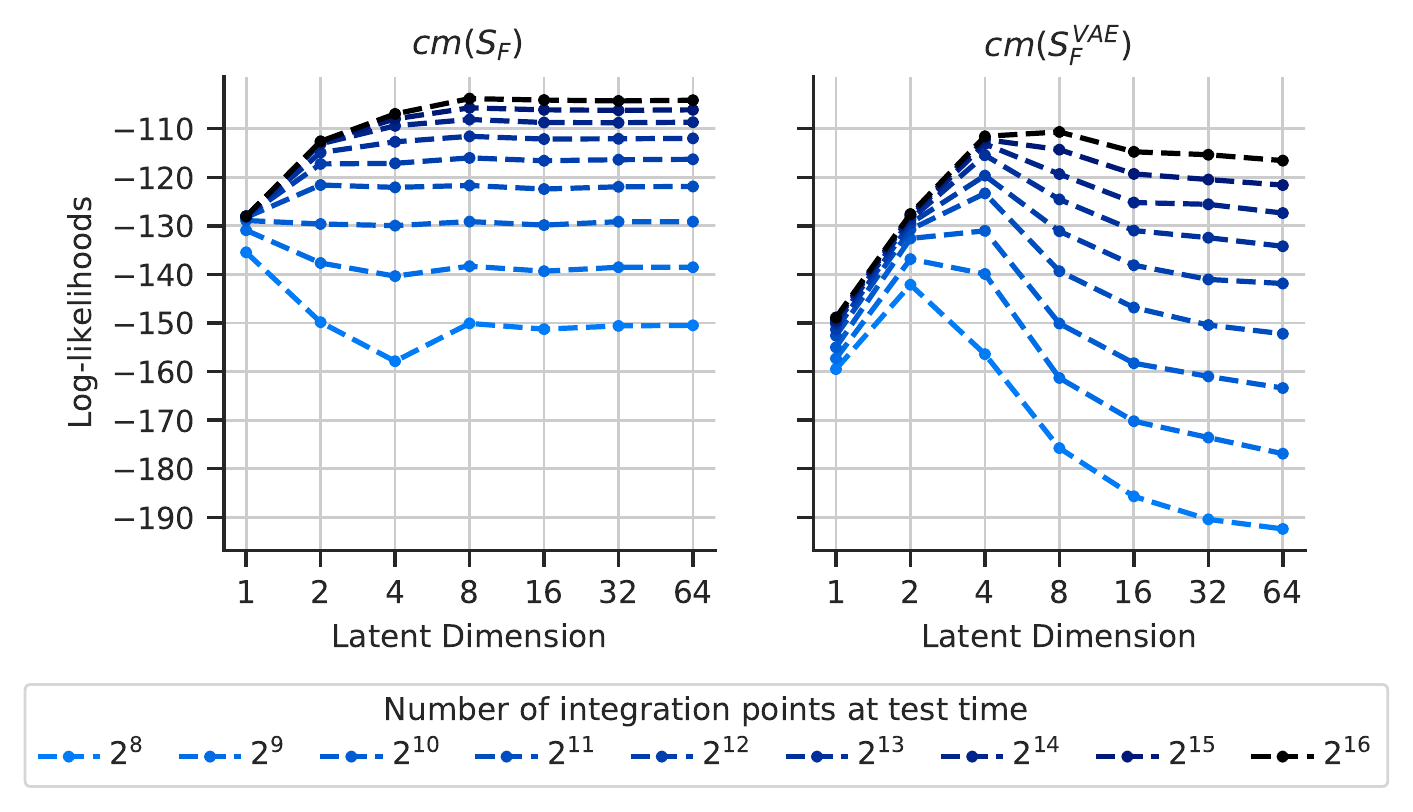}
    \caption{Test log-likelihood on the Binary MNIST dataset against latent dimensionality for $\cmf$ (left) and $\cmvae$ (right) models evaluated with different numbers of integration points.}
    \label{fig:latent_dims}
\end{figure*}

We also investigate the effect of the latent space dimensionality on the overall performance of the model.
We trained a $\cmf$ on Binary MNIST \cite{larochelle2011neural} with different latent dimensions $d$, namely $d{=}2^i$ for $i \in \{0, 1, 2, 3, 4, 5, 6\}$. 
We train all models with $2^{14}$ integration points but vary the number of integration points at test time from $2^{8}$ to $2^{16}$, as shown in Figure~\ref{fig:latent_dims}. In the following discussion, when referring to integration points we mean those used at \emph{test time}.

For low numbers of integration points (up to 1024), models with small latent spaces perform best. That is as expected, since small latent spaces are more amenable to numerical integration. However, as we increase the number of integration points, it is clear the model benefits from larger latent dimensions, eventually converging for 8 latent dimensions or higher. That is somewhat surprising, since numerical integration in high dimensions is notoriously difficult and we would expect performance to deteriorate as we increase the latent dimensionality. We hypothesise training via numerical integration strongly regularises the decoder so that the learned function is smooth enough to allow for reliable numerical integration irrespective of the latent dimensionality.

In Figure~\ref{fig:latent_dims}, we also report the same analysis for $\cm$ models trained via variational inference, $\cmvae$. We use the same decoder architecture and training protocol (see Appendix~\ref{sec:model}) as for $\cmf$. Here we see that numerical integration struggles with high latent spaces, as test log-likelihoods rapidly decline for $d\geq4$. That is simply because numerical integration becomes harder, since the ELBO actually improves for higher latent dimensions as shown in Table~\ref{tab:vae_bmnist}.

In all cases, it seems easier to numerically integrate---or extract good mixtures of tractable models from---models learned via numerical integration than models trained via variational inference. For $d\leq4$, numerical integration outperforms variational inference, whereas for $d>4$ we still get better mixtures from $\cmf$ than from $\cmvae$, although the latter is more expressive as indicated by the ELBO.

\begin{table*}[!h]
    \centering
    \begin{tabular}{l| c c c c c c c}
        Latent dimensions & 1 & 2 & 4 & 8 & 16 & 32 & 64 \\
        \midrule
        ELBO (VAE) & -154.13 & -132.21 & -112.77 & \textbf{-97.26} & \textbf{-93.66} & \textbf{-92.24} & \textbf{-92.74} \\
        LL ($\cmvae$) & -148.87 & -127.57 & -111.57 & -110.67 & -114.76 & -115.36  & -116.57 \\
        LL ($\cmf$) & \textbf{-128.05} & \textbf{-112.60} & \textbf{-106.97} & -103.82 & -104.08 & -104.29 & -104.21 \\
    \end{tabular}
    \caption{Test log-likelihoods for $\cmf$ and $\cmvae$. Evidence Lower Bound (ELBO) computed with 1000 Monte Carlo samples and log-likelihoods (LL) computed via RQMC with $2^{16}$ points.}
    \label{tab:vae_bmnist}
\end{table*}
\newpage

\section{Training via Amortised Variational Inference} \label{sec:variational}
In this section, we expand the discussion on (amortised) variational inference \cite{kingma2013auto,rezende2014stochastic} by considering binary density estimation benchmarks and limiting the latent dimensionality to 4. In Table~\ref{tab:vae} we report the results obtained using a $\cmvae$ with 4 latent dimensions and standard normal prior. In all but 3 datasets, a $\cm$ model trained via numerical integration ($\cmf$), outperformed $\cmvae$, indicating numerical integration is more effective than variational inference for small latent dimensions.

Interestingly, although $\cmf$ outperformed $\cmvae$ in most datasets in Table~\ref{tab:vae}, we still could not effectively approximate the log-likelihood of the continuous mixtures induced by $\cmvae$ via numerical integration. For 11 out of the 20 datasets, RQMC could not match the ELBO of $\cmvae$ (computed with 1000 MC samples) even when integrating with $2^{13}$ points. That indicates variational inference induces a complex decoder that is not amenable to numerical integration, even for a relatively small latent space of dimensionality $d=4$. Naturally, numerically integrating $\cmvae$ becomes even harder for higher latent dimensions, as shown in Figure~\ref{fig:latent_dims}.

\secondvaetable

These results indicate numerical integration is the best method to learn continuous mixtures of tractable probabilistic models. However, amortised variational inference is probably the most popular way to train continuous latent variables models and is more efficient than numerical integration: for each data point $\rvx$, variational inference computes $p(\rvx \cbar \param(\rvz_i))$ for only a few samples $\rvz_i$ from the posterior (as few as one \cite{kingma2013auto}), whereas numerical integration has to compute it once for every integration point. 
Therefore, it would be interesting to be able to extract good mixtures from models learned via variational inference. We believe one can improve on this by sufficiently regularising the decoder (e.g. penalising high Lipschitz constants) and thus facilitating numerical integration, but we leave that for future research.

\section{Plain Mixture Models}
For Monte Carlo integration with a fixed number of integration points $N$, the approximate model resembles a plain mixture model with equally-probable weights:
\begin{align}
    \hat p(\rvx) = \frac{1}{N}\sum_{i=1}^N p(\rvx \cbar \param(\rvz_i)) \approx p(\rvx) = \int p(\rvx \cbar \param(\rvz))p(\rvz)d\rvz .
\end{align}

Thus it makes sense to compare our models to plain mixture models. That is, instead of having mixture components as a function of some latent variable, each component $p_i$ is fully independent with
\begin{equation}
    p_{\text{mix}}(\rvx) = \frac{1}{N}\sum_{i=1}^N p_i(\rvx).
\end{equation}
\clearpage
In Table~\ref{tab:mix}, we compare plain mixtures to a $\cmf$ with 4 latent dimensions. We use a fully factorised structure for both models, that is
\begin{align}
    p_i(\rvx) &= \prod_{j=1}^{|\rmX|} p_i(x_j) \enspace \text {for plain mixture models,} \\
    p(\rvx \cbar \param(\rvz_i)) &= \prod_{j=1}^{|\rmX|} p(x_j | \param(\rvz_i)) \enspace \text {for $\cmf$}.
\end{align}

We train both models using the exact same protocol (see Appendix~\ref{sec:model}).
The results in Table~\ref{tab:mix} show that our model outperformed plain mixtures ($\dm(\struct_F)$) in all datasets, even though both models were trained the exact same way and have the exact same structure if we consider a $\cmf$ compiled to a 1024 components mixture. The only difference is how we parametrise the mixture components. 

Similarly, we also considered mixture models with learnable weights, that is 
\begin{align}
    p_{\text{mix}}(\rvx) &= \sum_{i=1}^{N} w_i p_i(\rvx),
\end{align}
where $\{w_i\}_{i=1}^{N}$ are learnable parameters with $w_i \geq 0$ and $\sum_i w_i = 1.$ We train these mixture models both via gradient descent ($\dm(\struct_F^{W})$) and Expectation Maximisation ($\dm(\struct_F^{EM})$). As shown in Table~\ref{tab:mix}, our model outperforms plain mixtures in all datasets but \texttt{dna} and \texttt{msnbc}, where learning via EM produces better test log-likelihoods. However, existing PC architectures (namely HCLT and RAT-SPNs) already performed better than our models in these datasets, so it is not that surprising that a discrete mixture works well for these two datasets. It is also worth mentioning that $\cmf$ results could be further improved via latent optimisation (see the main paper for a discussion on latent optimisation and some results for $\cmclt$).

\plainmixturestable

For a fixed number of integration points (mixture components) discrete mixture models are strictly more expressive than continuous ones, since in the former the parameters of each component are completely independent. Nonetheless, continuous mixtures outperform discrete ones in almost all cases, indicating the regularisation introduced by the shared `decoder' does facilitate learning. This result, albeit not surprising, has never been fully exploited in the tractable probabilistic models literature, to the best of our knowledge.

\section{Additional Experimental Results} \label{sec:extra_res}
\subsection{Binary Density Estimation Benchmarks}
\begin{figure*}[htbp]
    \centering
    \includegraphics{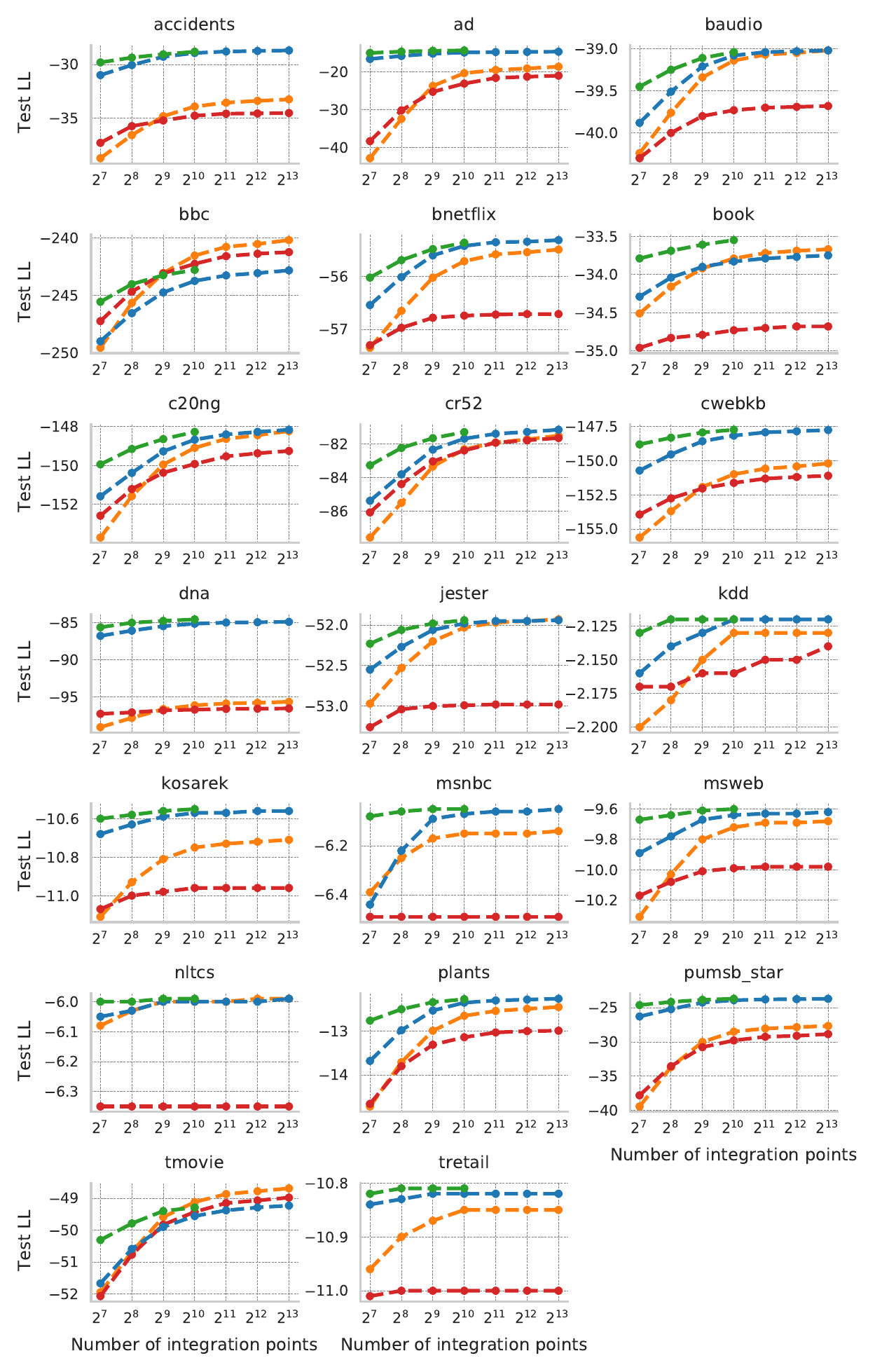}
    \caption{Test log-likelihood against number of integration points for $\cmf$ (orange), $\cmclt$ (blue), $\locmclt$ (green) and $\cmvae$ (red). Results are averaged over 5 random seeds.}
    \label{fig:loplots}
\end{figure*}

In this section, we present additional results on continuous mixtures, showing test log-likelihoods on the 20 binary density estimation benchmarks for different numbers of integration points in Tables~\ref{tab:cmf} and \ref{tab:cmclt}. 
We also include latent optimisation results for $\cmclt$ in Table~\ref{tab:latopt}, showing significant improvements especially when using few integration points.
Table~\ref{tab:20datasets-extend} extends the results presented in Table~\ref{tab:20datasets} detailing the performance of each PC method, while Table~\ref{tab:20datasetscomplete} includes the standard deviation of $\cm$ results across 5 random seeds.
Finally, in Figure~\ref{fig:loplots} we show the performance of $\cmf$, $\cmclt$, $\locmclt$ and $\cmvae$ with an individual plot for each of the 20 density estimation benchmarks for different numbers of integration points.

\dmmtable
\dmmclttable
\secondlatopttable
\competitorstable
\clearpage
\binarydatasetstablecomplete

\subsection{Non-binary Image Data} \label{sec:cont_image}

We also show results obtained when modelling non-binary images (MNIST, Fashion MNIST and SVHN) as continuous data. In that case, we followed best practices from the literature in generative models \cite{theis2016note} and applied a `jittering' process whereby one adds uniform noise to pixel values and subsequently divides the result by 256. We kept the same architectures and training protocol from previous experiments with $K=1$ (see Efficient Learning section), but this time we parametrised the leaves as normal distributions with learnable mean and variance. To avoid numerical issues, we bounded variance values between $1\mathrm{e}{-6}$ and ${1.0}$ during training. This parametrisation results in slightly more compact models: for MNIST, `Small Einet' and `Big Einet' had respectively 5 and 84 million parameters; and for SVHN, 5 and 163 million parameters.

Interestingly, as seen in Table~\ref{tab:imagedatasetsnormal}, in that scenario $\cmf$ is outperformed by both Einets. One possible explanation is that the optimisation problem is intrinsically more complicated in the continuous case, with the variance of normal distributions being particularly hard to optimise \cite{dai2018diagnosing}. Einets mitigate that by using online EM \cite{peharz2016latent,peharz2020einsum}, which performs a form of  natural gradient descent \cite{sato1999fast}, and thus greatly facilitates learning \cite{amari1998natural}. Still, there remains a  gap in performance between PCs learnt on discrete data (Table~\ref{tab:imagedatasets}) and PCs learnt on continuous data (Table~\ref{tab:imagedatasetsnormal}). Curiously, that gap is much smaller on SVHN than on MNIST, possibly because it is easier to scale normal distributions, which require learning only two parameters in comparison to the 256 parameters of categorical distributions.

\imagedatasetsnormaltable

\begin{figure}[htbp]
\centering
\begin{subfigure}{.325\linewidth}
    \centering
    \includegraphics[width=\textwidth]{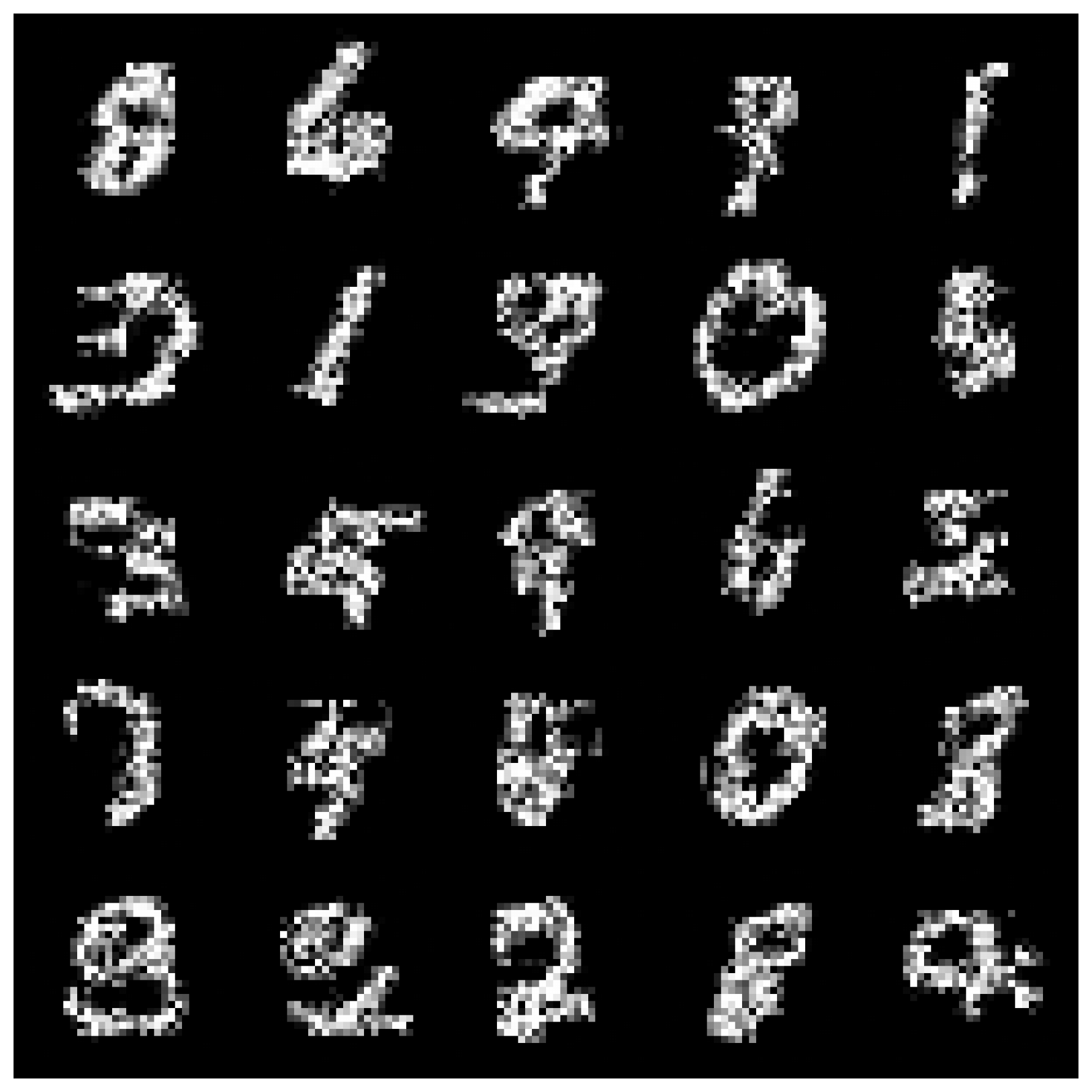}
\end{subfigure}
\begin{subfigure}{.325\linewidth}
    \centering
    \includegraphics[width=\textwidth]{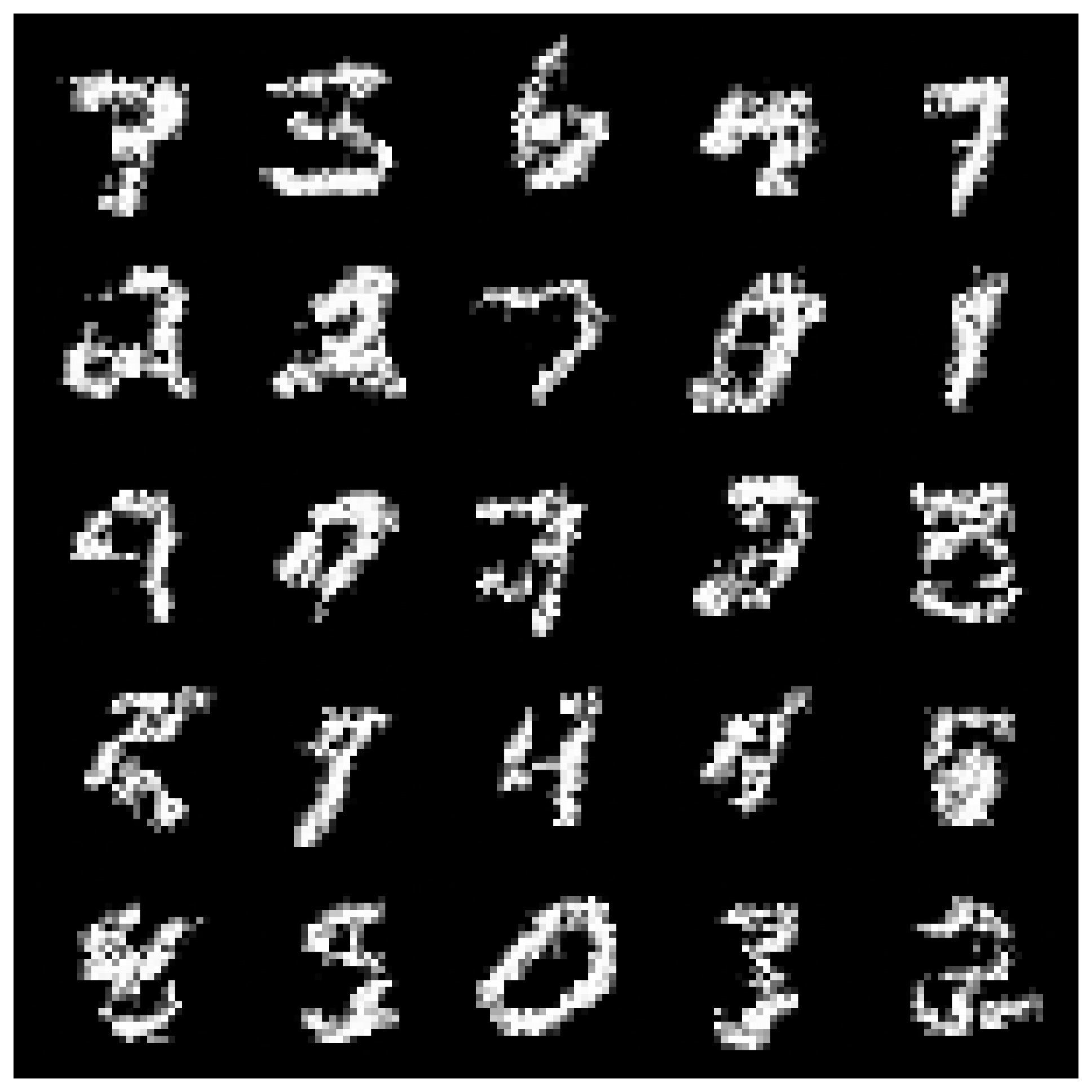}
\end{subfigure}
\begin{subfigure}{.325\linewidth}
    \centering
    \includegraphics[width=\textwidth]{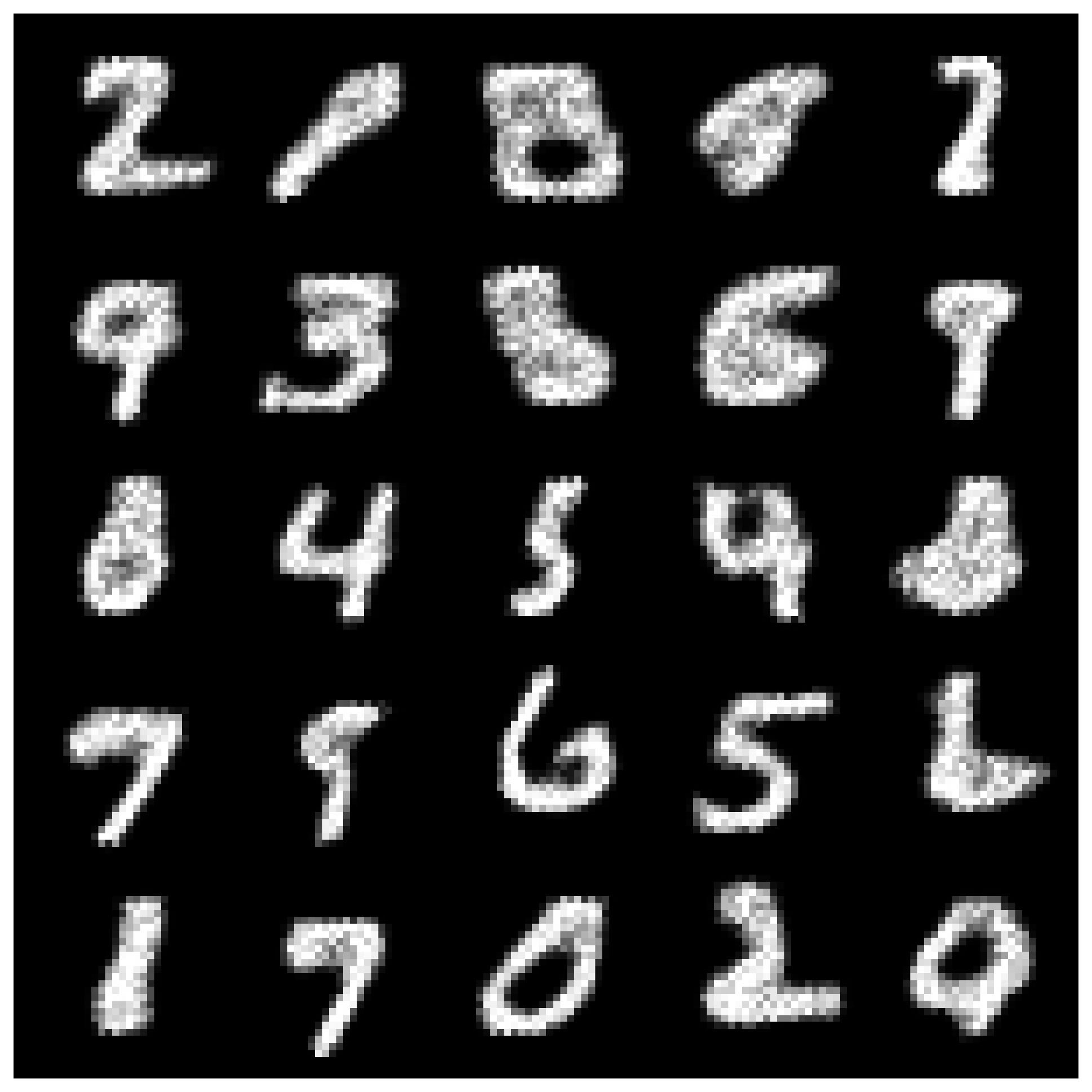}
\end{subfigure}

\smallskip
\begin{subfigure}{.325\linewidth}
    \centering
    \includegraphics[width=\textwidth]{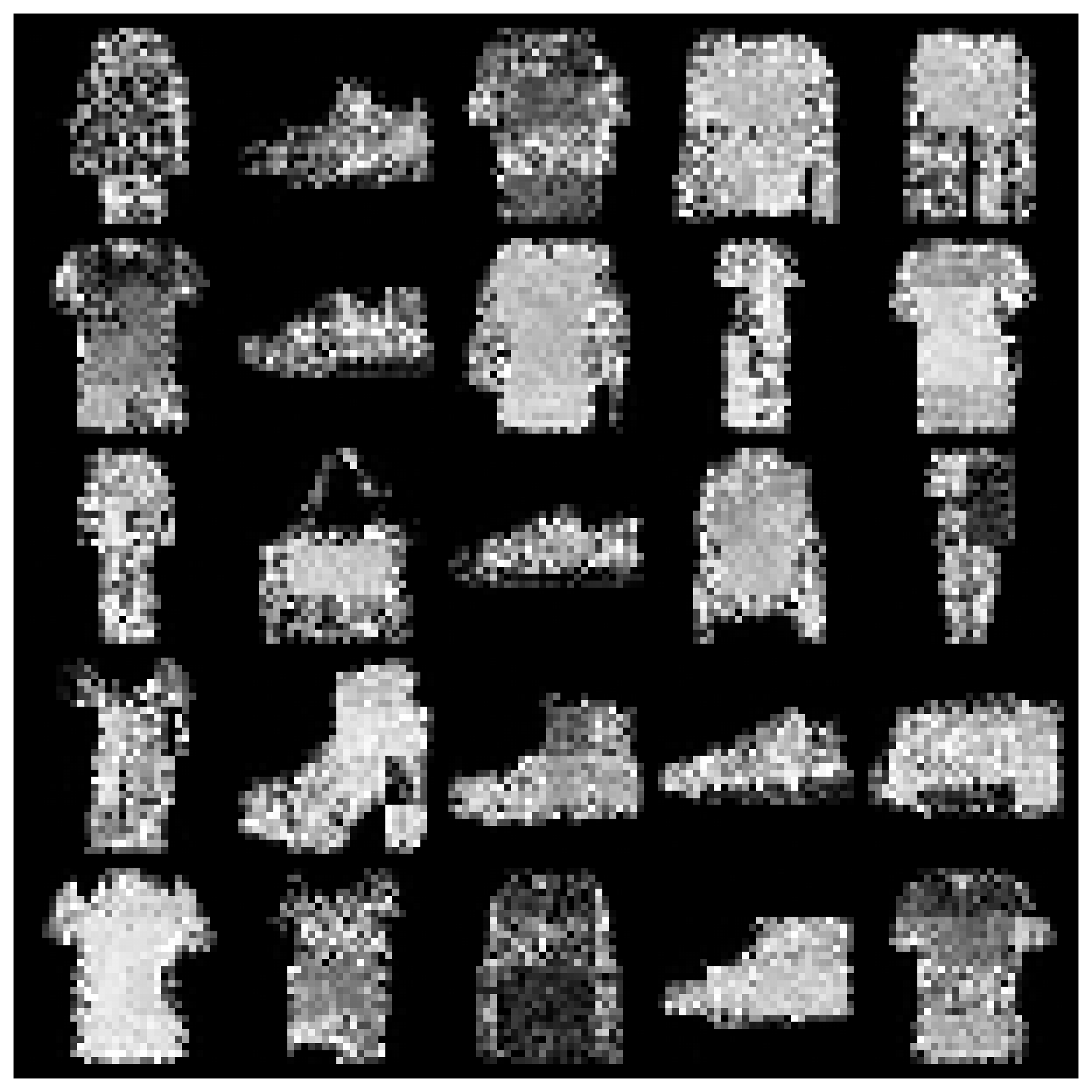}
\end{subfigure}
\begin{subfigure}{.325\linewidth}
    \centering
    \includegraphics[width=\textwidth]{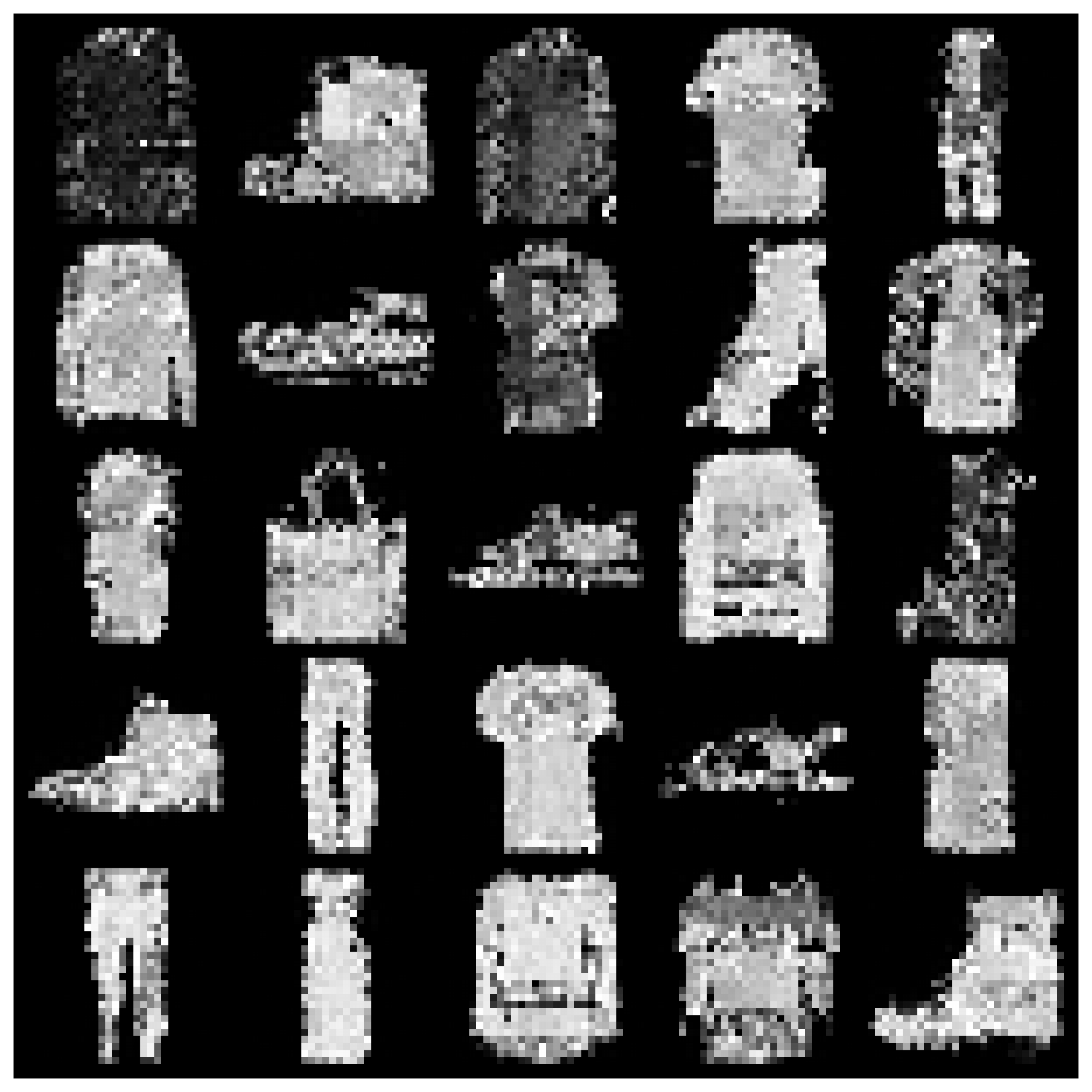}
\end{subfigure}
\begin{subfigure}{.325\linewidth}
    \centering
    \includegraphics[width=\textwidth]{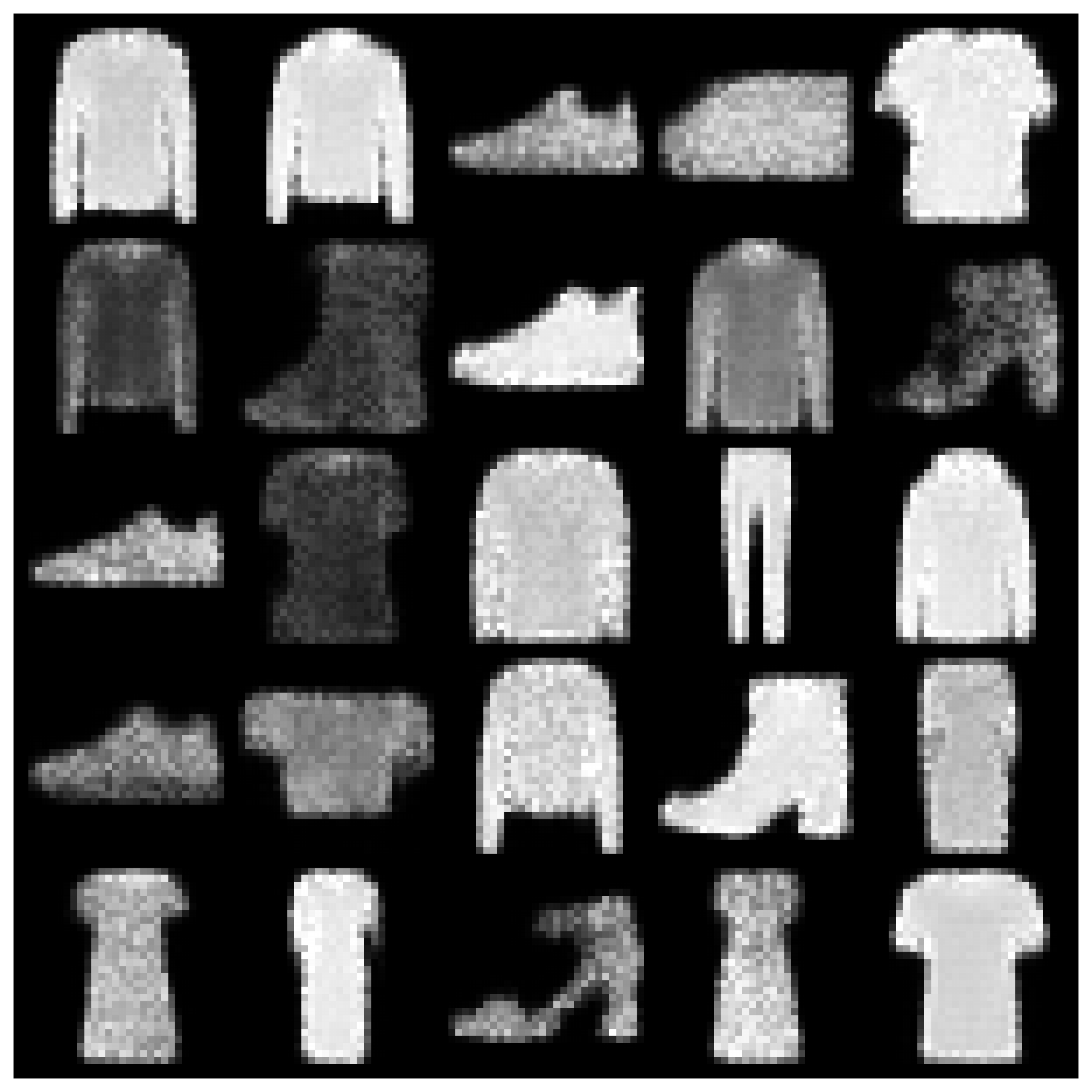}
\end{subfigure}

\smallskip
\begin{subfigure}{.325\linewidth}
    \centering
    \includegraphics[width=\textwidth]{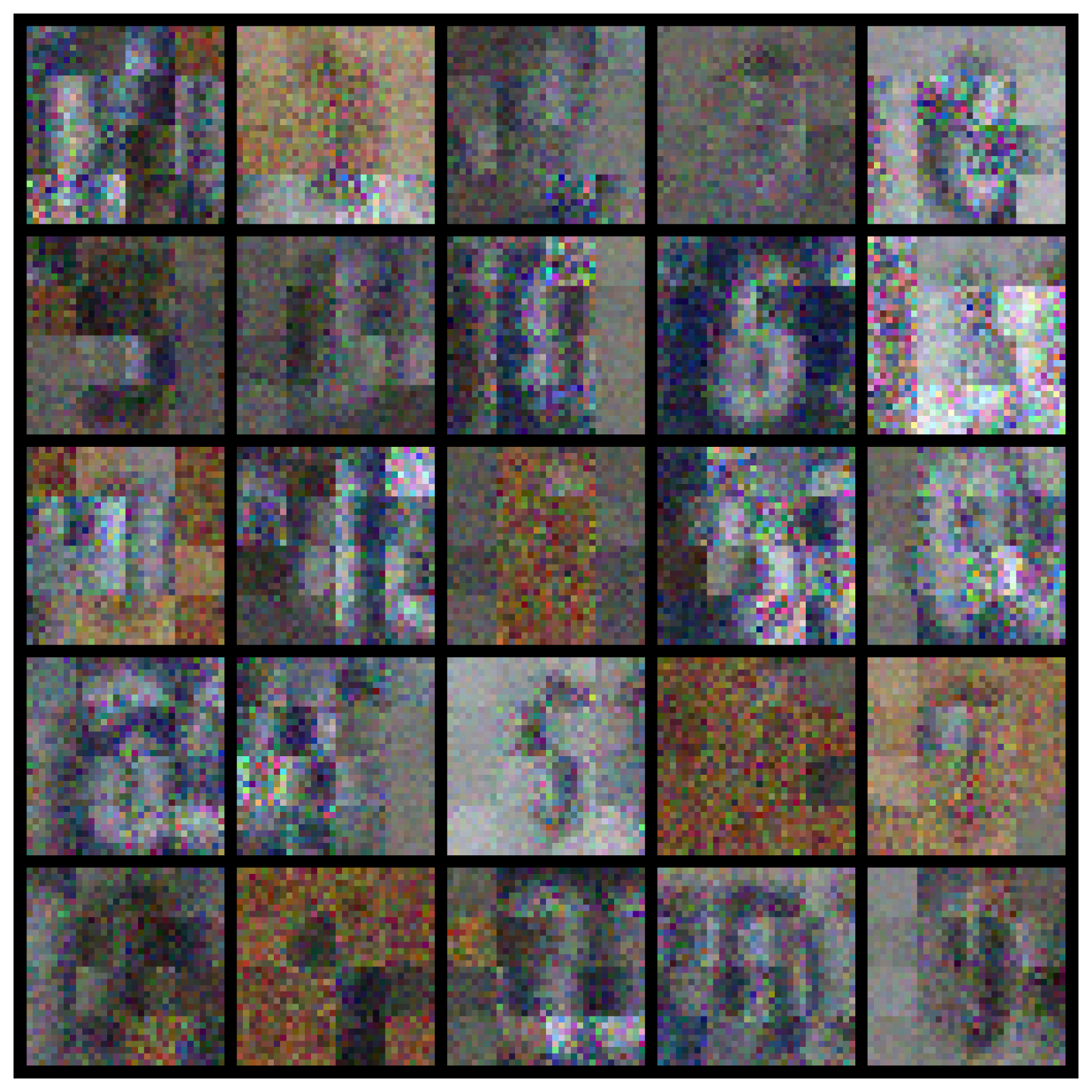}
\end{subfigure}
\begin{subfigure}{.325\linewidth}
    \centering
    \includegraphics[width=\textwidth]{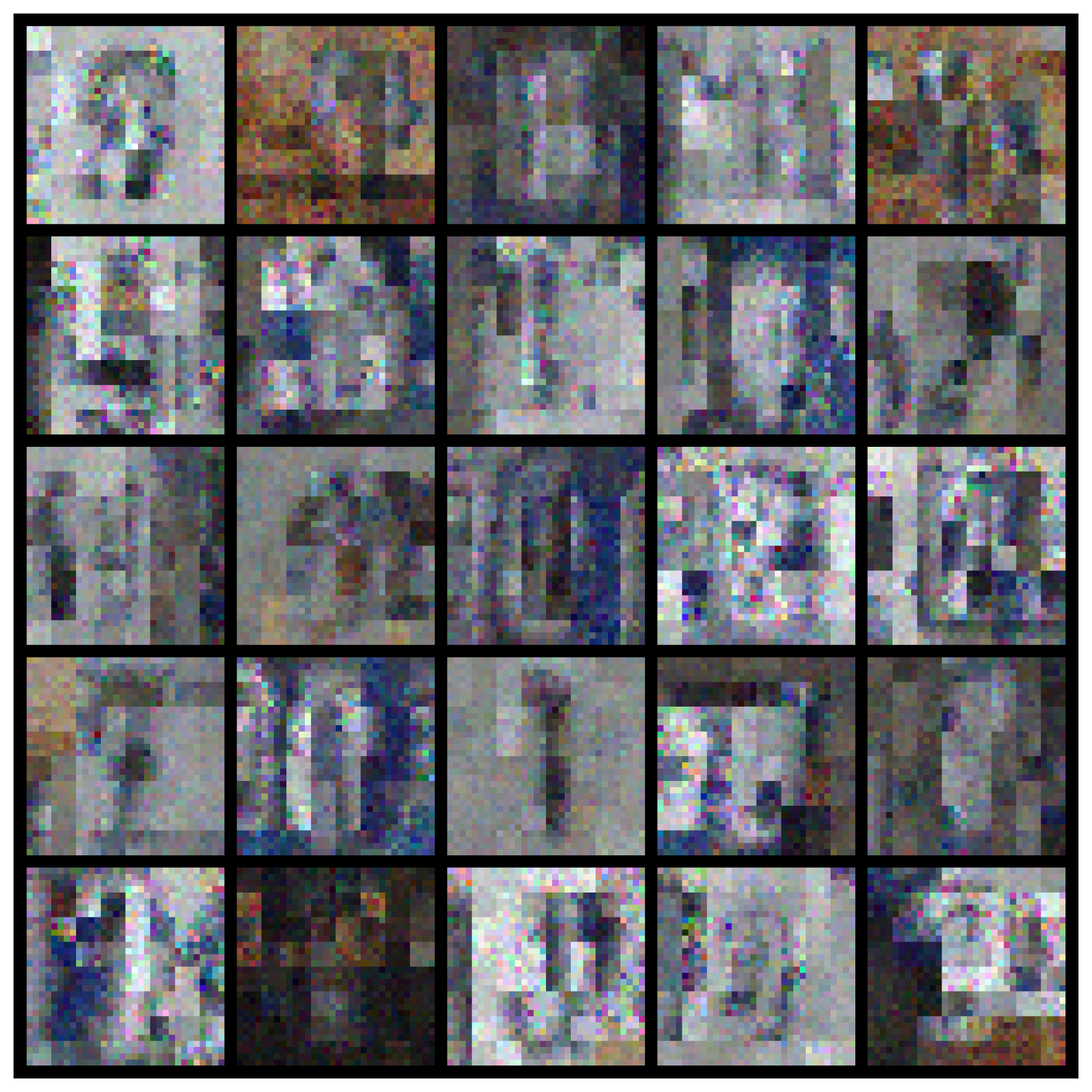}
\end{subfigure}
\begin{subfigure}{.325\linewidth}
    \centering
    \includegraphics[width=\textwidth]{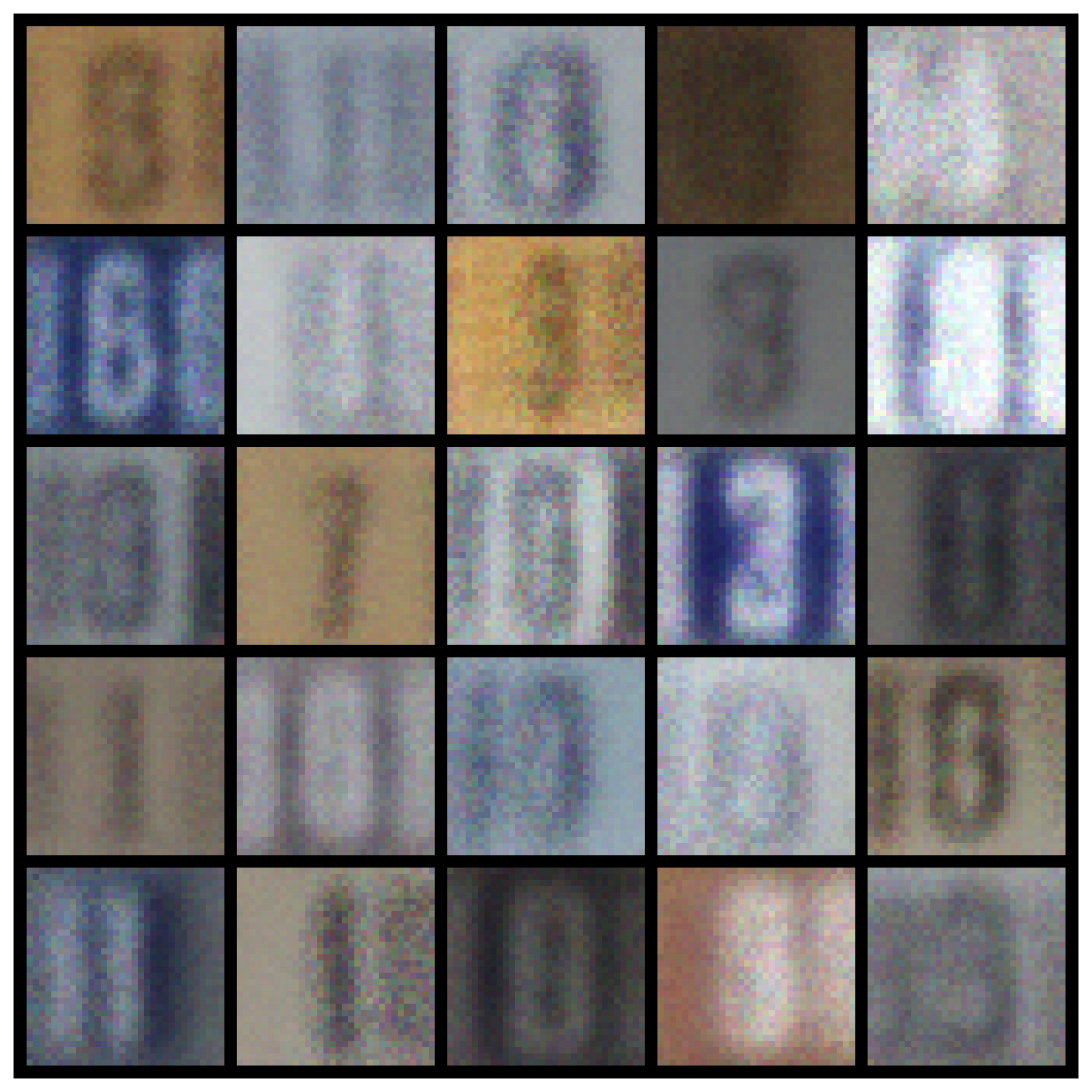}
\end{subfigure}
\caption[Image samples from Einets and $\cmf$ with normal distributions at the leaves.]{Image samples from models with \emph{normal distributions} at the leaves: `Small Einet' (left column), `Big Einet' (middle column) and $\cmf$ (right column). Once more we see $\cmf$ offers better sample quality than Einets.}
\label{fig:normal_samples}
\end{figure}

In terms of sample quality, $\cmf$ models still generate smoother and less pixelated images than Einets as seen in Figure~\ref{fig:normal_samples}. That is a clear example that sample quality and log-likelihood values are weakly correlated (Theis et al. 2016). Moreover, by comparing Figures~\ref{fig:samples} and~\ref{fig:normal_samples}, we can infer normally distributed leaves produce images that are smoother but also more blurry.

\begin{figure}[htbp]
\centering
\begin{subfigure}{.325\linewidth}
    \centering
    \includegraphics[width=\textwidth]{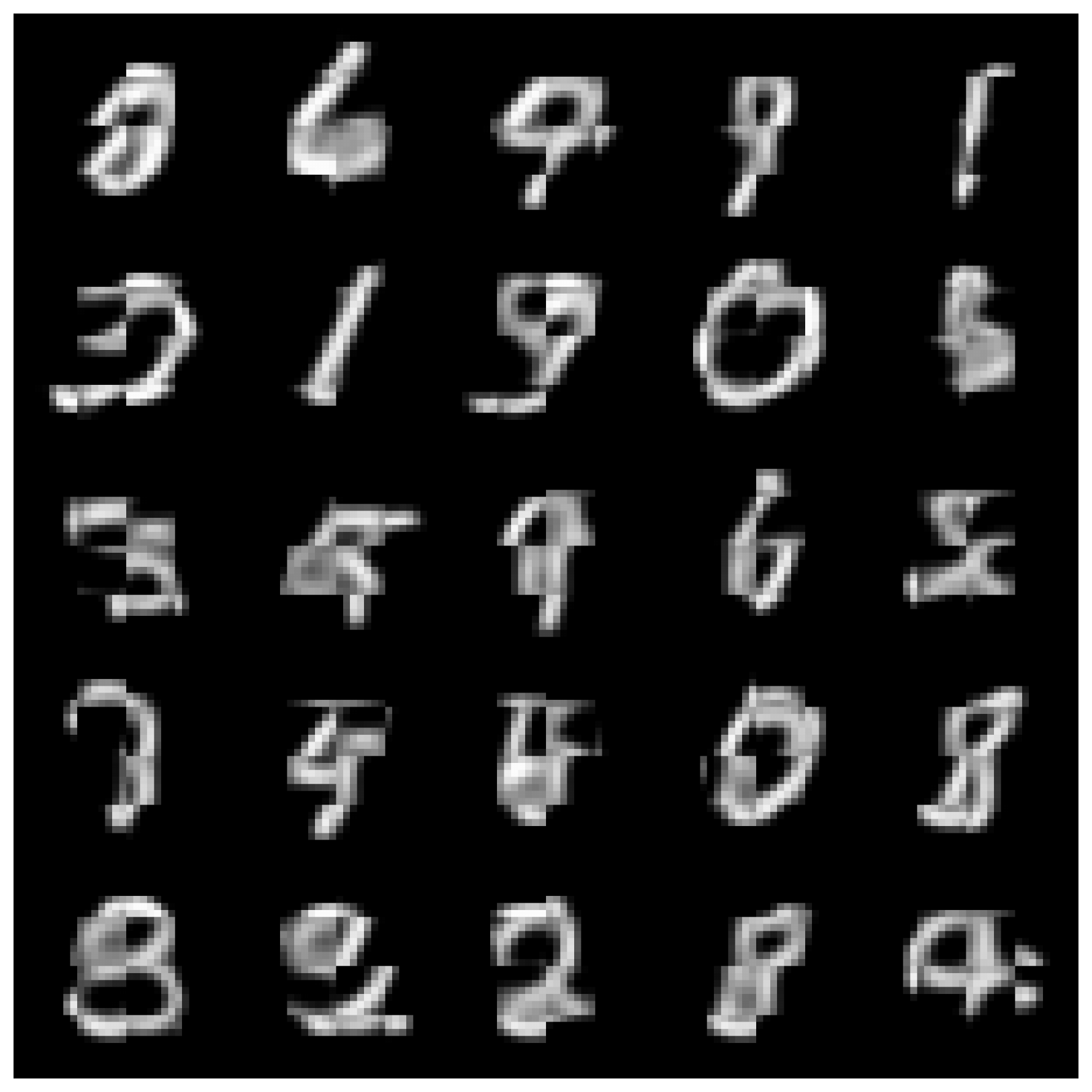}
\end{subfigure}
\begin{subfigure}{.325\linewidth}
    \centering
    \includegraphics[width=\textwidth]{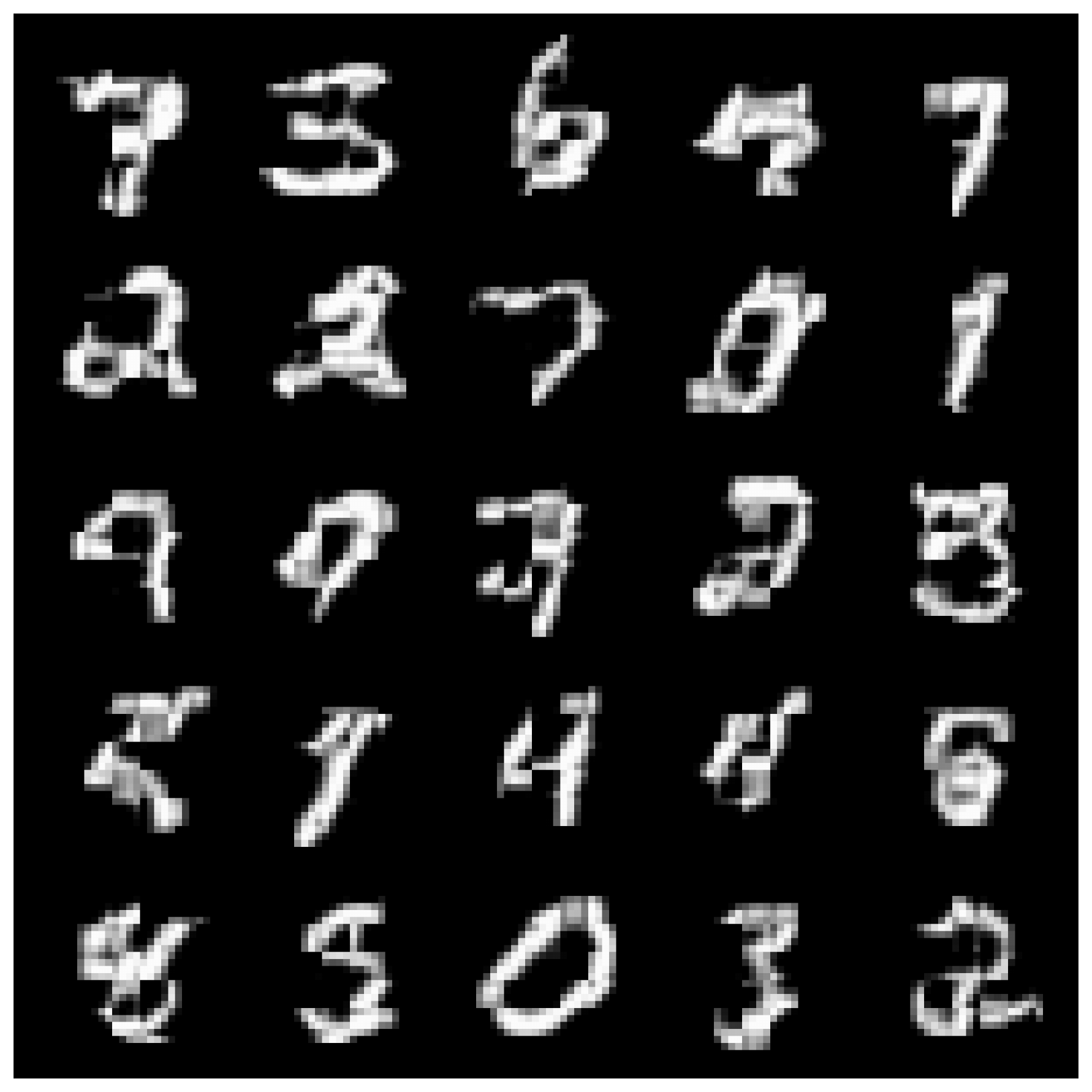}
\end{subfigure}
\begin{subfigure}{.325\linewidth}
    \centering
    \includegraphics[width=\textwidth]{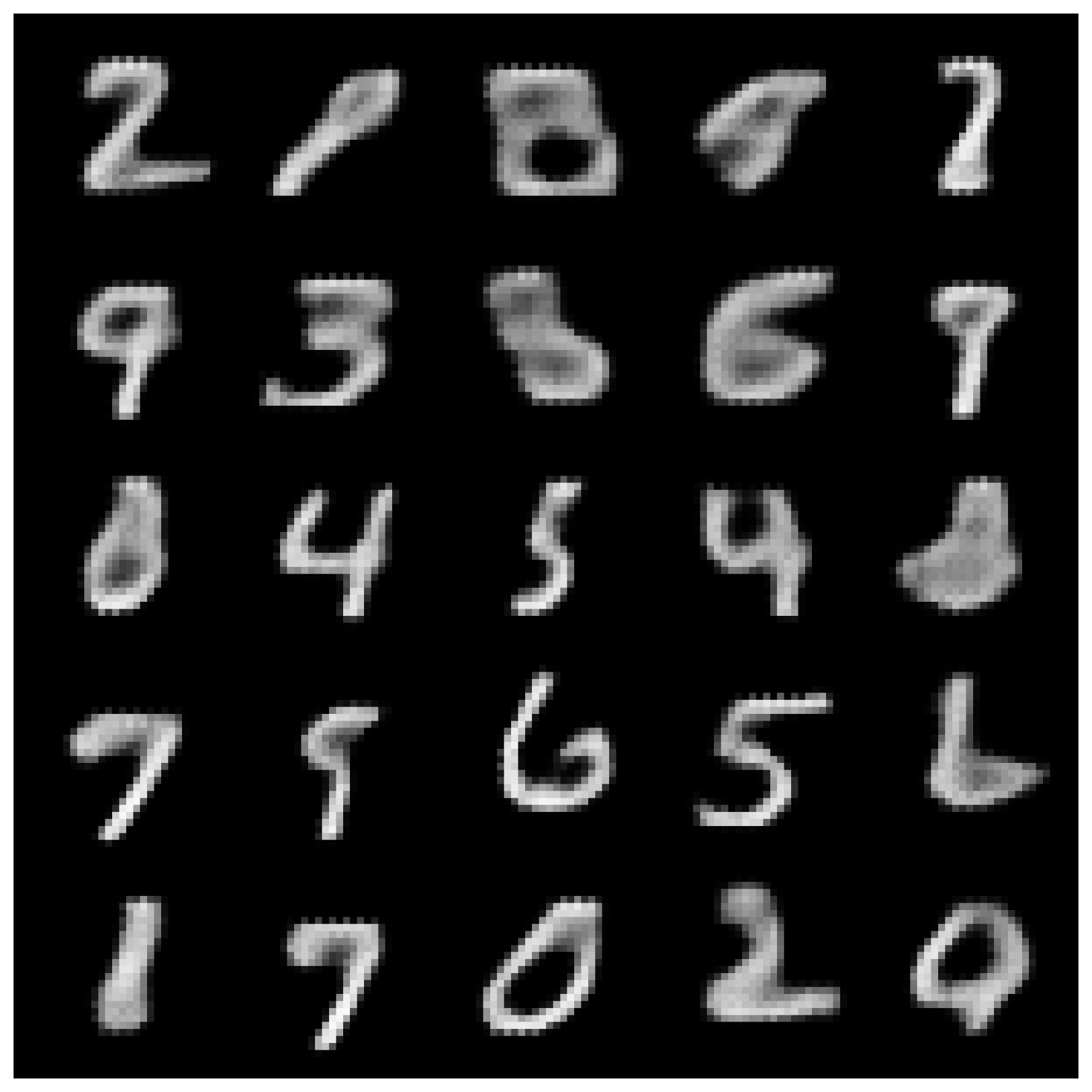}
\end{subfigure}

\smallskip
\begin{subfigure}{.325\linewidth}
    \centering
    \includegraphics[width=\textwidth]{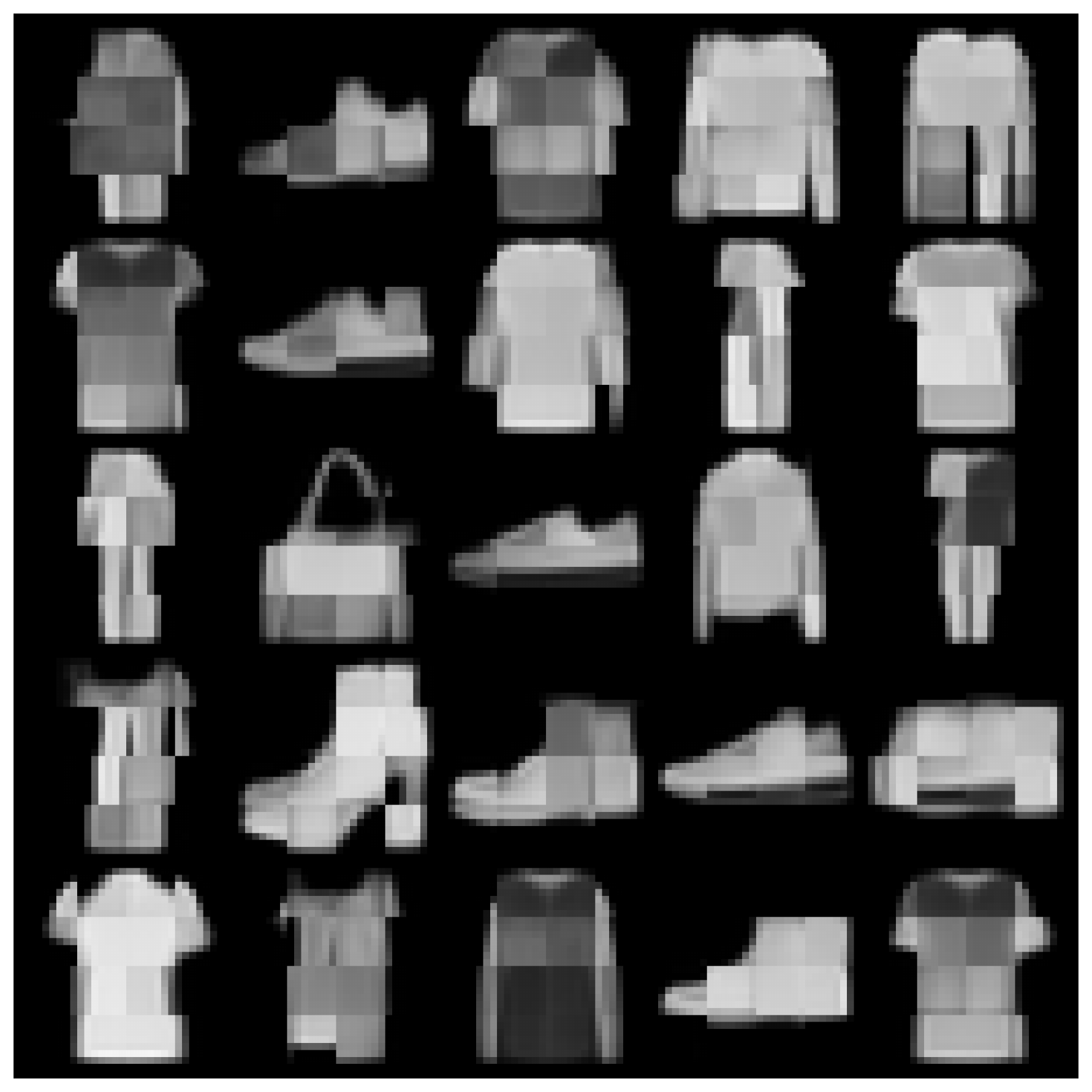}
\end{subfigure}
\begin{subfigure}{.325\linewidth}
    \centering
    \includegraphics[width=\textwidth]{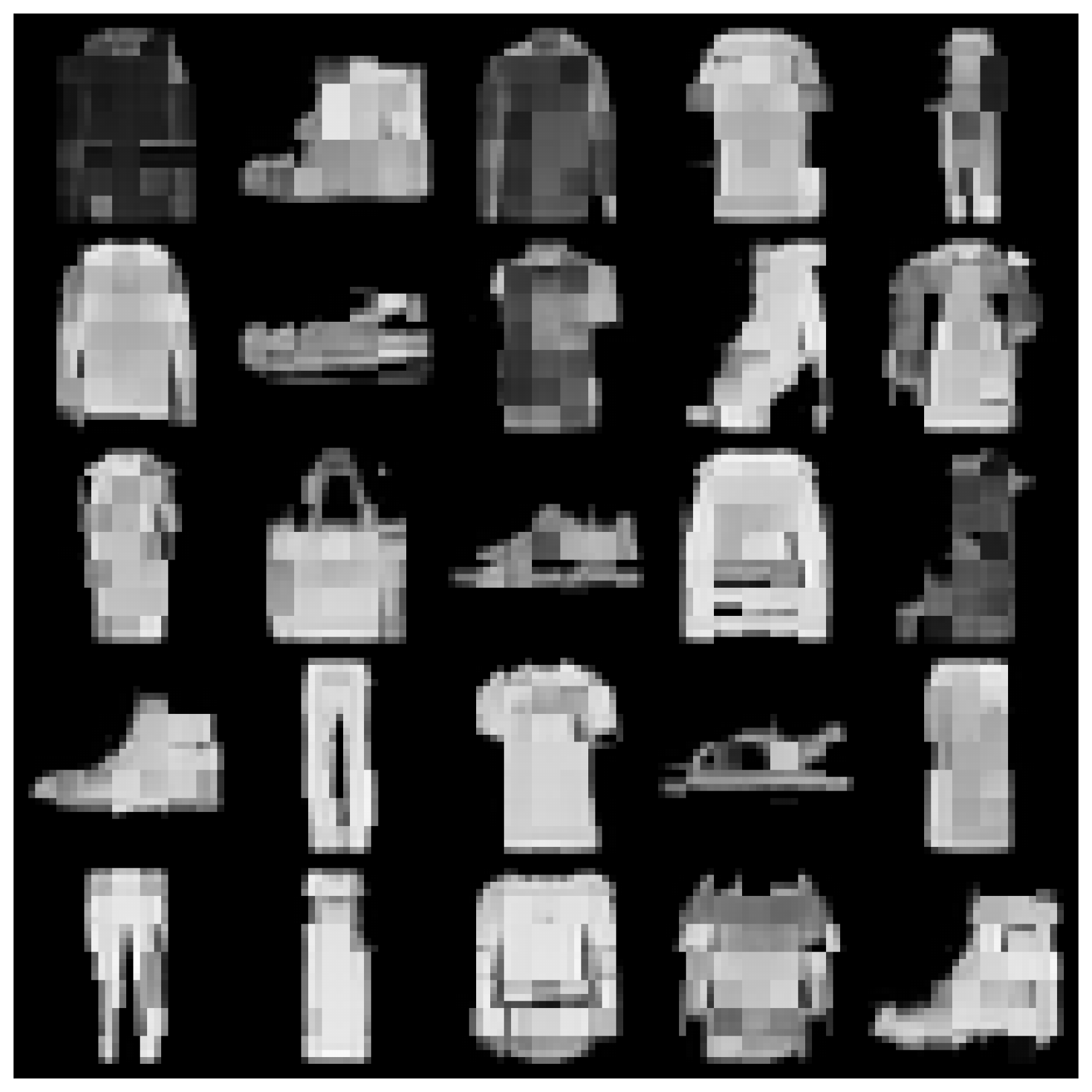}
\end{subfigure}
\begin{subfigure}{.325\linewidth}
    \centering
    \includegraphics[width=\textwidth]{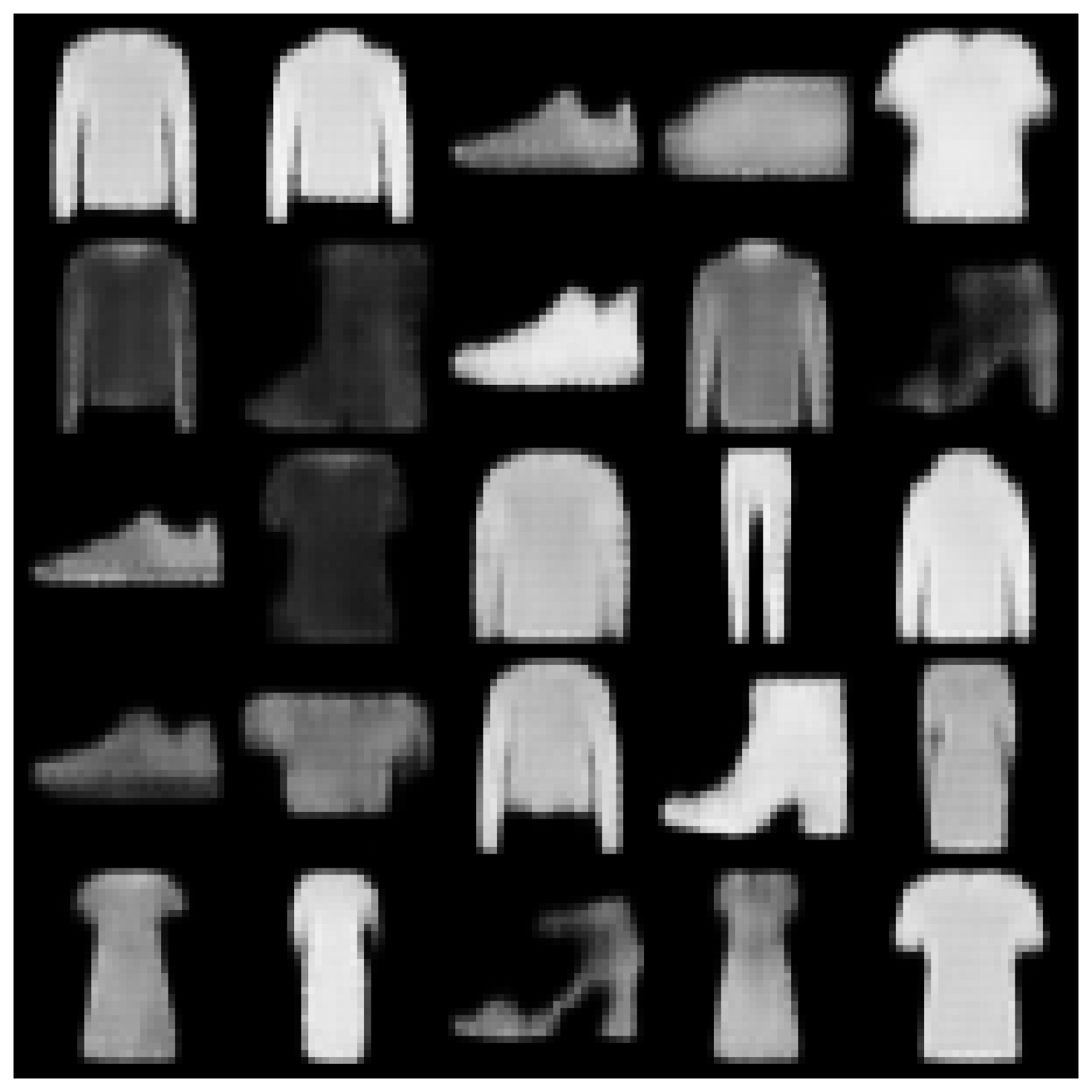}
\end{subfigure}

\smallskip
\begin{subfigure}{.325\linewidth}
    \centering
    \includegraphics[width=\textwidth]{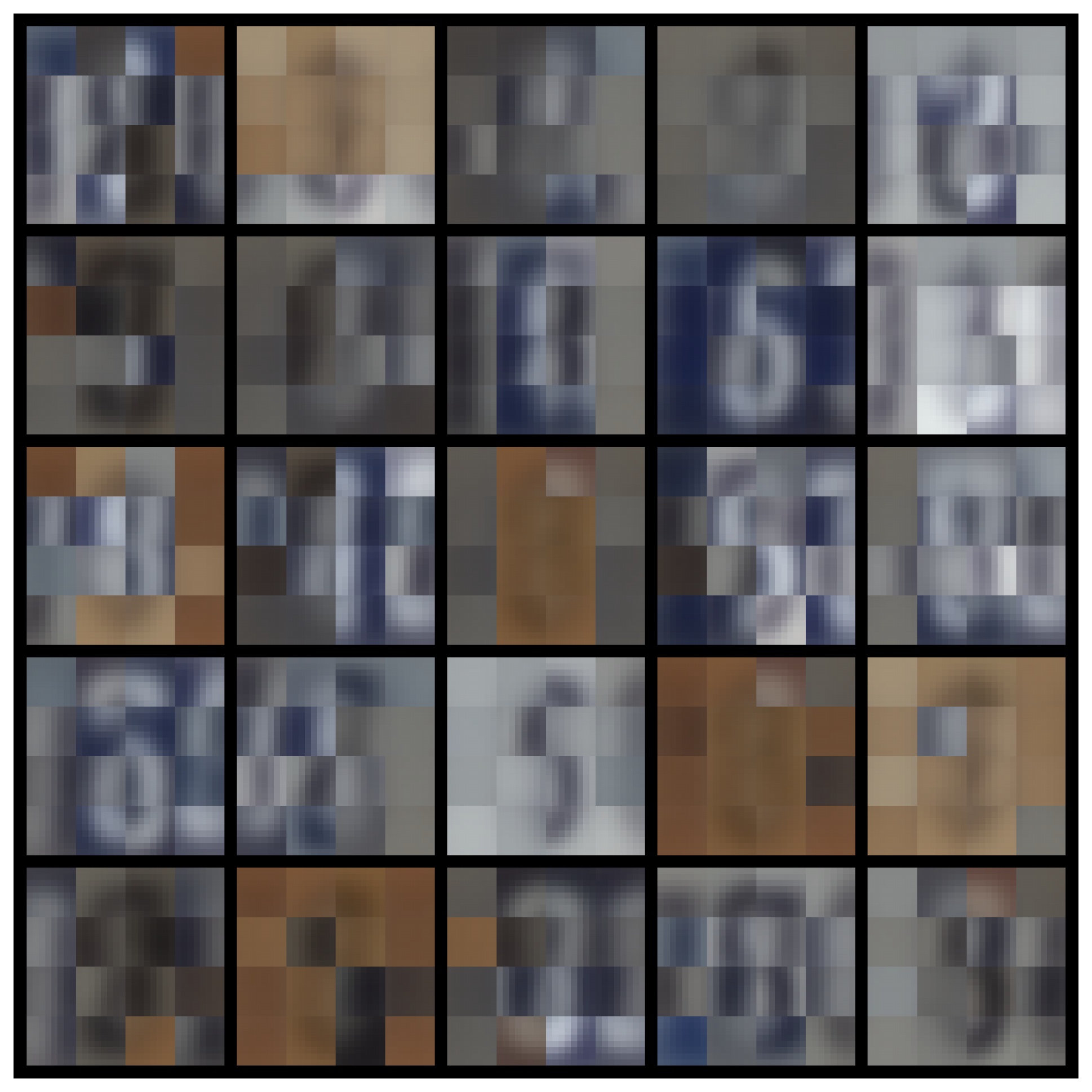}
\end{subfigure}
\begin{subfigure}{.325\linewidth}
    \centering
    \includegraphics[width=\textwidth]{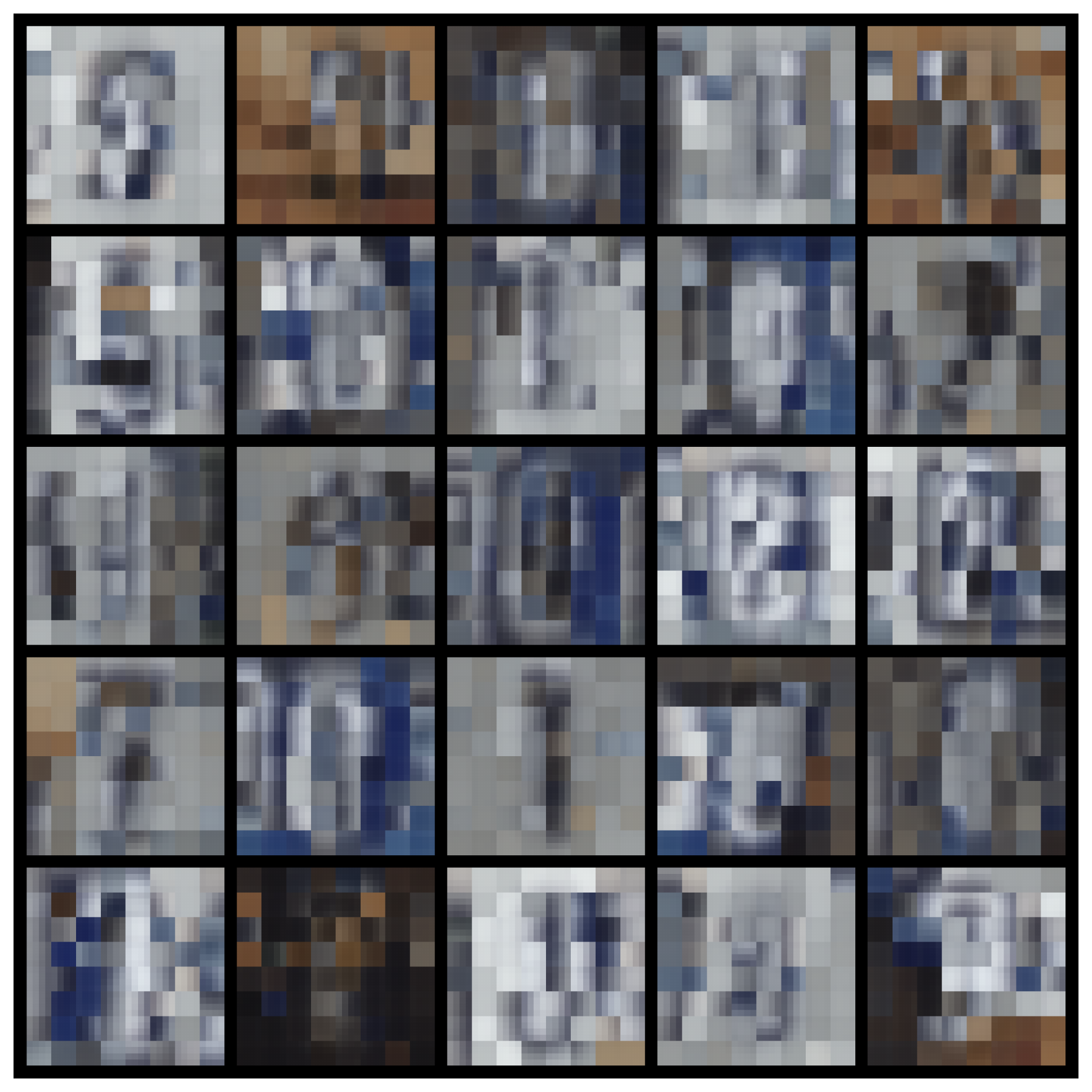}
\end{subfigure}
\begin{subfigure}{.325\linewidth}
    \centering
    \includegraphics[width=\textwidth]{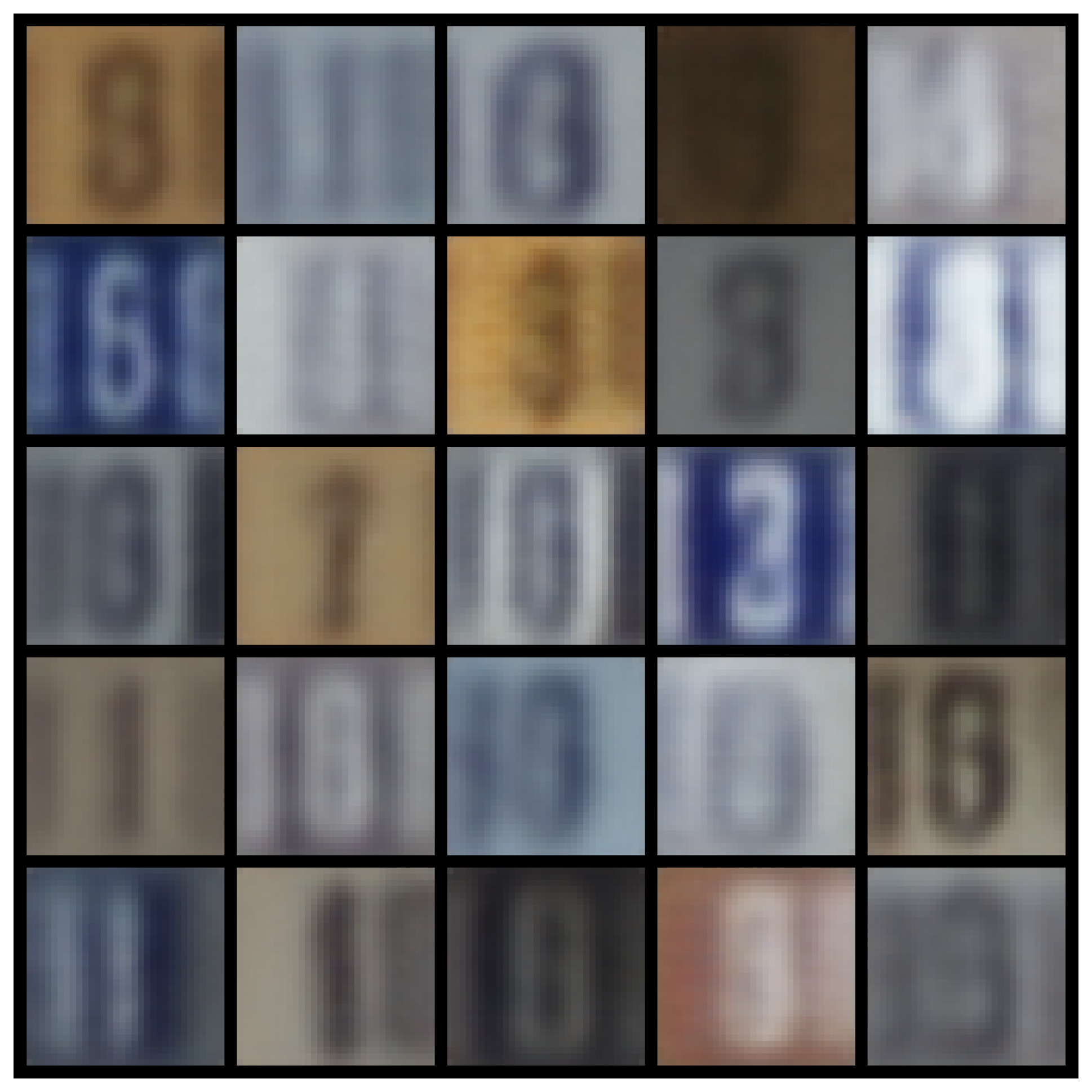}
\end{subfigure}
\caption[Image samples from Einets and $\cmf$ with normal distributions at the leaves but ignoring the variance of individual pixels.]{Image samples from models with \emph{normal distributions} at the leaves but ignoring the variance of individual pixels: `Small Einet' (left column), `Big Einet' (middle column) and $\cmf$ (right column).}
\label{fig:normal_zerovar_samples}
\end{figure}

Yet, there is a caveat in the variability of samples generated by a $\cmf$ with either normal or categorical leaves. Consider the PC induced by the numerical integration of a $\cmf$ with $N$ integration points. Such a PC is essentially a shallow mixture model and is thus only capable of generating $N$ meaningfully different samples. That is because in any given mixture component of a $\cmf$, any variation in the generated samples comes from univariate distributions at the leaves which only amounts to some added noise, as visible in the images of Figures~\ref{fig:samples} and~\ref{fig:normal_samples}. That is particularly apparent in a $\cmf$ with normal distributions, for which it is possible to plot the mean values of the sampled leaves, ignoring the variance of individual pixels. In Figure~\ref{fig:normal_zerovar_samples} we see images thus generated are more visually appealing, suggesting the variance at the leaves is not useful for sample generation.

Naturally, a $\cmf$ might be capable of capturing more variance in the data and, in the context of this experiment, represent a larger collection of images. That is, we can get a more diverse set of samples if we resample $\rvz \sim p(\rvz)$ each time, instead of using a compiled PC approximating the underlying $\cmf$. Unfortunately, if we rely on the continuous latent space for sampling, we are not, strictly speaking, sampling from a tractable model, as we must rely on approximate numerical integration methods to compute likelihoods. As mentioned previously, these can be made arbitrarily precise, but that is still not directly comparable to fully tractable models like Einets. At any rate, the comparisons in Figures~\ref{fig:samples}, \ref{fig:normal_samples} and \ref{fig:normal_zerovar_samples} remain valid (we sample from a compiled PC), although we recognise they are not enough to fully evaluate the variance of the images generated by each model.

\clearpage
\clearpage

\section{Demonstration of Tractable Queries}
\subsection{Training with missing data}
Our models support efficient marginalisation out-of-the-box both at training and test time, like any other Probabilistic Circuit (PC). More precisely, for any partition of the domain variables into observed $\rmX_{o}$ and non-observed variables $\rmX_{\neg o}$, one can compute $p(\rmX_o) = \int_{\rvx_{\neg o}} p(\rmX_o, \rvx_{\neg o}) d\rvx_{\neg o}$, with the same computational complexity of computing $p(\rmX)$.
This is formally demonstrated for Sum-Product Networks \cite{Peharz2015b}; a class of PCs to which our approximate model $\sum_{i=1}^N w(\rvz_i)p(\rvx \cbar \param(\rvz_i))$ belongs to. 

\begin{figure}[!ht]
    \centering
    \begin{subfigure}{\linewidth}
        \centering
        \includegraphics[width=0.9\textwidth]{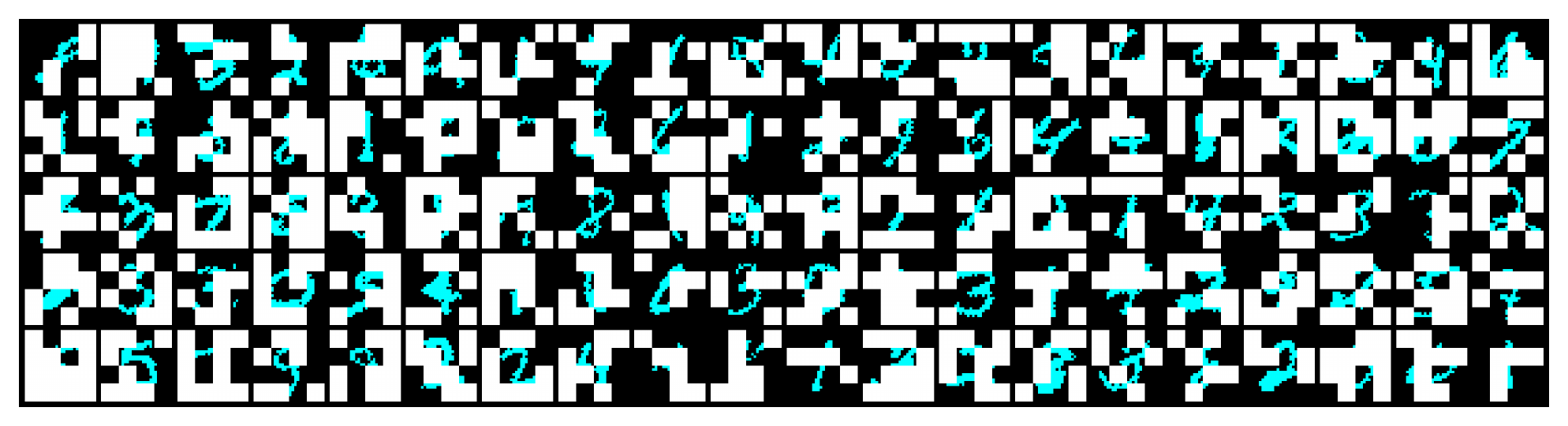}
    \end{subfigure}
    \begin{subfigure}{\linewidth}
        \centering
        \includegraphics[width=0.9\textwidth]{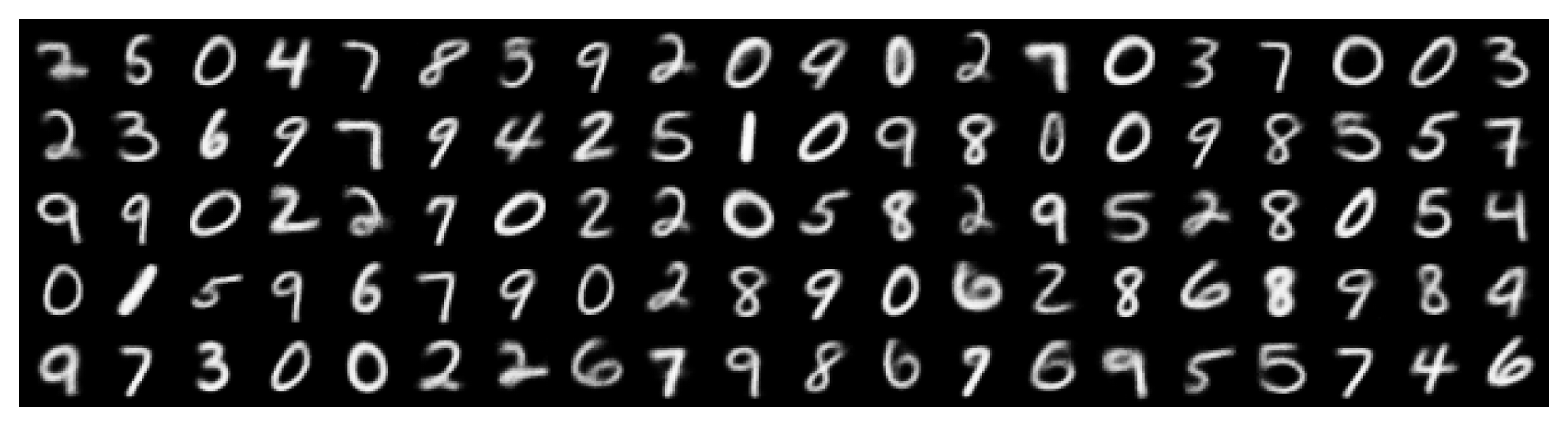}
    \end{subfigure}
    \caption{
        Illustration of the ability of $\cmf$ to handle missing data via marginalisation. On top we have Binary MNIST training images with blocks of pixels masked at random. We display the training images in colour to distinguish pixel values from non-observed pixels (depicted in white). On the bottom, we have samples (continuous Bernoulli) from a $\cmf$ trained on incomplete Binary MNIST images, like the ones shown above. The models always generates complete digits, despite never observing one during training.
    }
    \label{fig:missing_training}
\end{figure}

In a nutshell, marginalisation in PCs boils down to evaluating marginals at the leaves, which is often simple to compute (e.g. with univariate leaves). 
That is the same for continuous mixtures, where inference is performed via the approximate model above; itself a PC supporting efficient exact marginalisation. That means, numerical integration can approximate any marginal query in continuous mixtures in the exact same principled manner, without requiring imputation or ad-hoc techniques to handle missing values.
We illustrate this by showing samples from a $\cmf$ (with Bernoulli leaves) trained only with incomplete Binary MNIST samples, as depicted in Figure~\ref{fig:missing_training}. At training time, we omit blocks of pixels from each image completely at random, meaning the model never observes a complete digit. Nonetheless, the model generates complete and reasonable digits, indicating proper marginalisation of the missing pixels at training time. 

\subsection{Inpainting}
\begin{figure}[htbp]
    \centering
    \begin{subfigure}{\linewidth}
        \centering
        \includegraphics[width=0.75\textwidth]{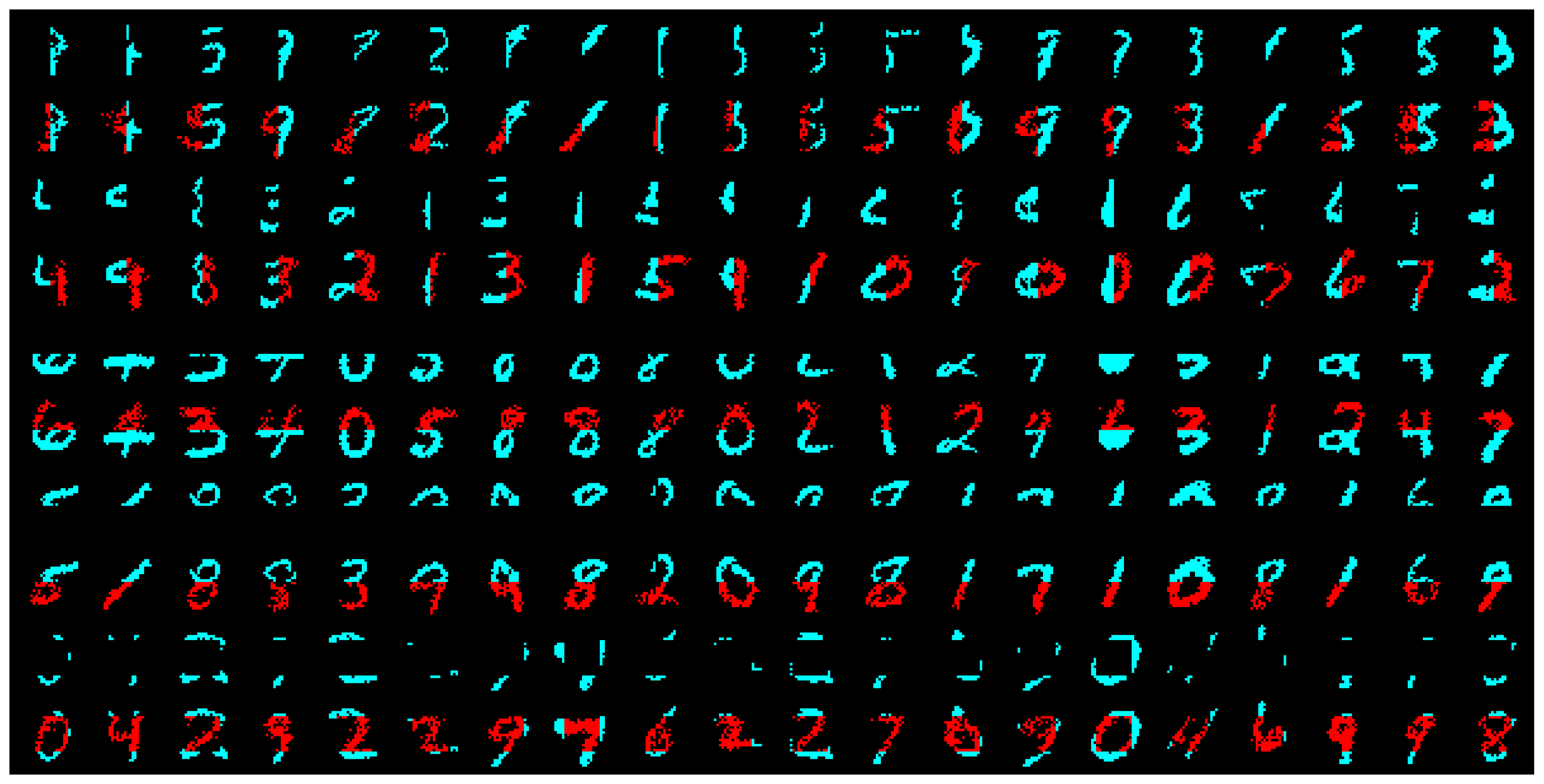}
    \end{subfigure}
    \begin{subfigure}{\linewidth}
        \centering
        \includegraphics[width=0.75\textwidth]{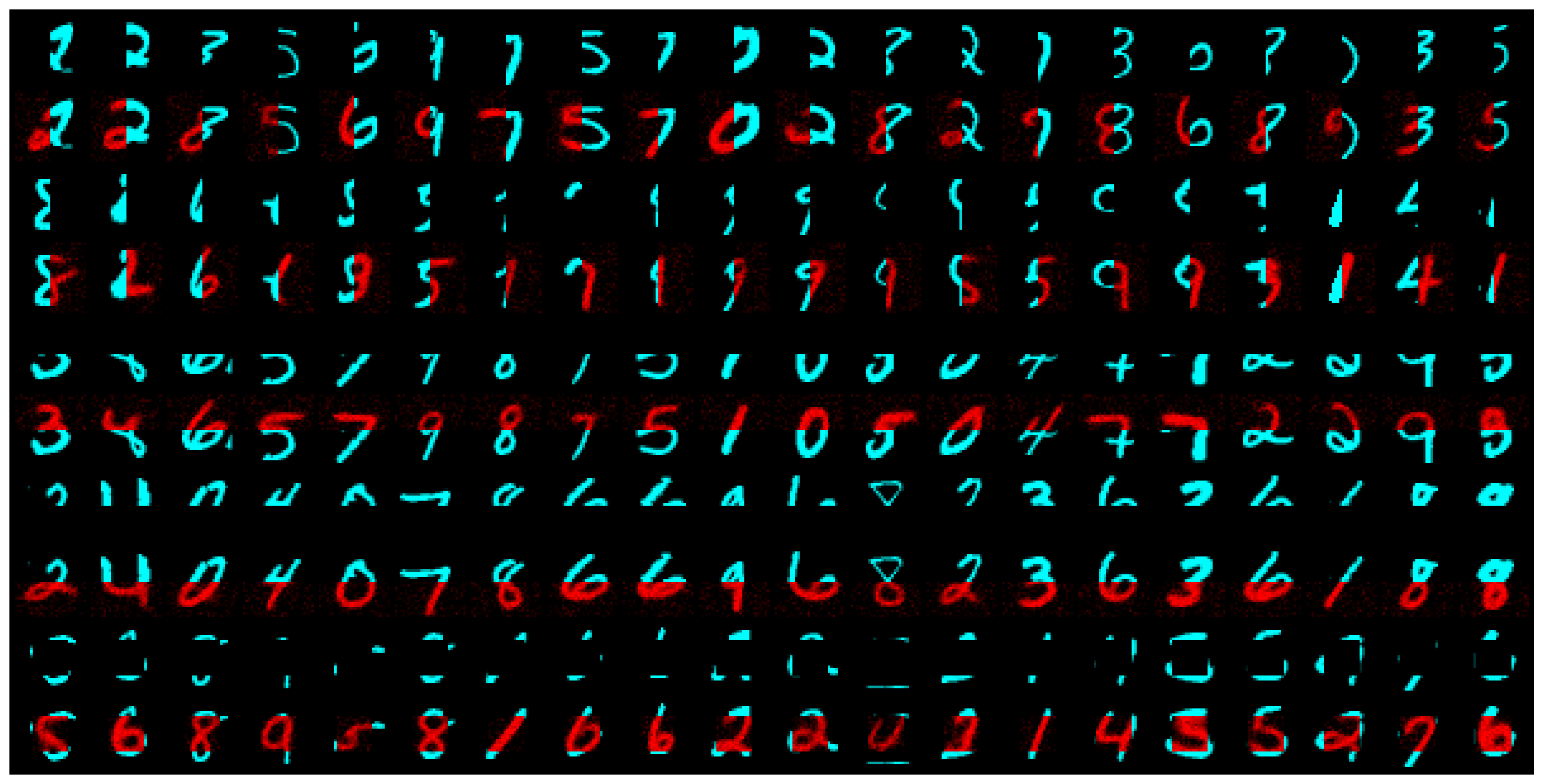}
    \end{subfigure}
    \begin{subfigure}{\linewidth}
        \centering
        \includegraphics[width=0.75\textwidth]{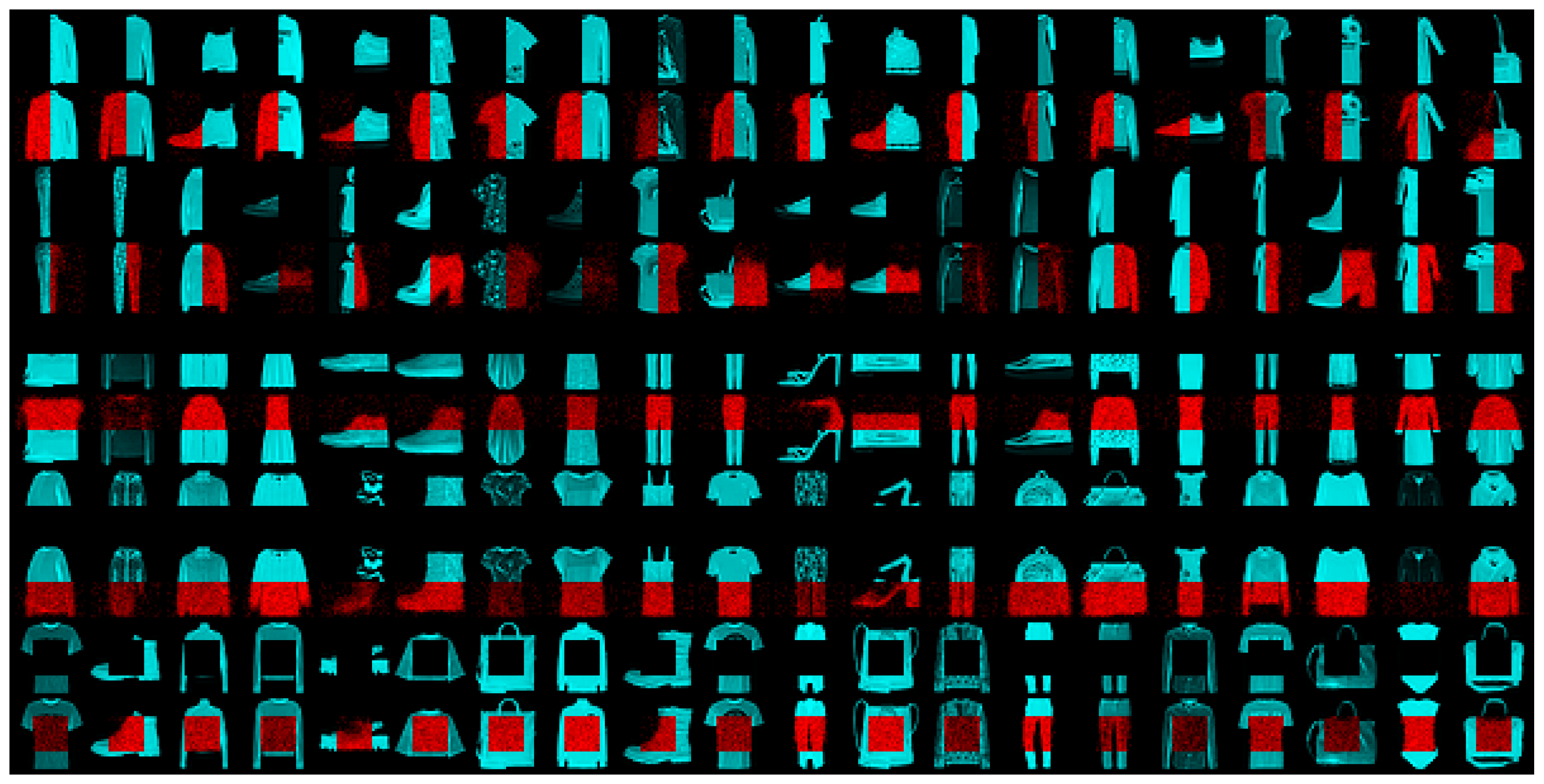}
    \end{subfigure}
    \caption{
        Inpainting results on Binary MNIST (top), MNIST (middle) and Fashion MNIST (bottom) obtained with $\cmf$. In alternating rows, we have first the original image with a missing part, and below the reconstructed image output by $\cmf$. Pixels from the original (reconstructed) images are depicted in blue (red).}
    \label{fig:inpainting}
\end{figure}

Our models also support efficient \emph{approximate} Most Probable Explanation (MPE) queries. This property also stems directly from the underlying PC in the approximate model $\sum_{i=1}^N w(\rvz_i)p(\rvx \cbar \param(\rvz_i))$.
In Figure~\ref{fig:inpainting}, we show an application of this type of query via inpainting with a $\cmf$ model on Binary MNIST (Bernoulli leaves) and on MNIST and Fashion MNIST (normal leaves). 

To compute an MPE for the images in Figure~\ref{fig:inpainting}, we first compile the model to a discrete mixture with $2^{14}$ integration points and evaluate the approximate model on the incomplete images (possible thanks to efficient marginalisation in PCs as discussed above). This allows us to perform a simple (approximate) posterior inference by selecting, for each image, the component (integration point) that is the most likely to have generated it. Finally, we take the most likely value from these individual components to get the final reconstruction. Albeit simple, this MPE procedure produces good reconstructions as shown in Figure~\ref{fig:inpainting}.
Note that, for the results in Figure~\ref{fig:inpainting}, we use the same set of integration points to run MPE on each and every image. That means, the results demonstrated here are equivalent regardless of the interpretation of the model: numerical integration on the continuous mixture, or exact inference on a compiled PC.
\clearpage

\clearpage
\section{Monte Carlo (MC) vs Randomised Quasi-Monte Carlo (RQMC)} \label{sec:rqmc}
\begin{figure}[h!]
    \centering
    \includegraphics{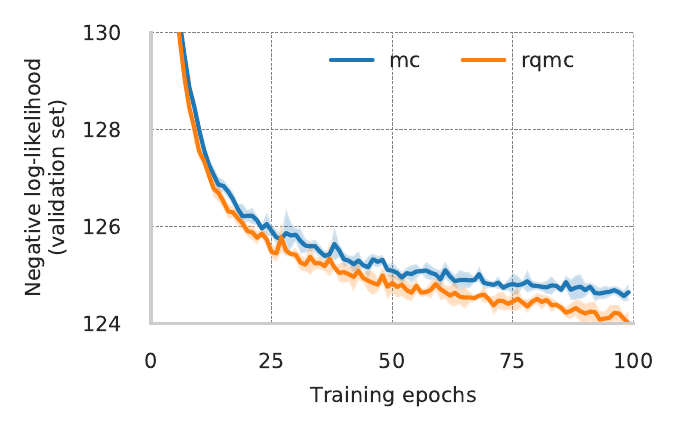}
    \caption{Log-likelihood on validation set against number of training epochs for $\cmf$ trained on Binary MNIST with MC (blue) and RQMC (orange). We use $2^{10}$ integration points in both cases and plot one standard deviation confidence bands computed with 5 different random seeds.}
    \label{fig:mc_rqmc}
\end{figure}

In practice, randomised quasi-Monte Carlo (RQMC) can be seen as a variance reduction method \cite{l2016randomized}, which often translates to faster convergence of learning algorithms \cite{buchholz2018quasi}.
We also have observed RQMC to improve convergence in our use cases, and thus have used RQMC in all of our experiments. To illustrate this, we compare the training of $\cmf$ on Binary MNIST using MC and RQMC. In Figure~\ref{fig:mc_rqmc}, it is clear that RQMC facilitates learning and improves the overall solution by a small but non-negligible amount.

\end{document}